\RequirePackage[svgnames]{xcolor}

\documentclass[11pt,logo,letterpaper]{berkeley}

\usepackage[utf8]{inputenc} %
\usepackage[T1]{fontenc}    %
\usepackage{lmodern}
\usepackage{textcomp}
\usepackage[svgnames]{xcolor}
\usepackage{asymptote}
\usepackage[comma,authoryear,compress]{natbib}
\usepackage{hyperref}
\usepackage{algorithm}
\usepackage{algorithmicx}
\usepackage{algpseudocode}
\usepackage{microtype}
\usepackage{graphicx}
\usepackage{booktabs} %
\usepackage{float}
\usepackage{bigstrut}
\usepackage{tikz}
\definecolor{myred}{rgb}{0.7, 0.3, 0.0}
\definecolor{myblue}{HTML}{054488}
\definecolor{mygreen}{HTML}{056b34}

\newcolumntype{R}[1]{>{\raggedleft\let\newline\\\arraybackslash\hspace{0pt}}m{#1}}

\usetikzlibrary{intersections}

\definecolor{darkgreen}{rgb}{0.0, 0.42, 0.24}

\usepackage{amsthm}
\usepackage{nicefrac}
\usepackage{subcaption}
\usepackage{caption}
\usepackage{graphicx}
\usepackage{cleveref}
\usepackage{bxcoloremoji}
\usepackage{listings}
\usepackage{float}
\usepackage{longtable}
\lstdefinestyle{python}{
    language=Python,
    basicstyle=\ttfamily\footnotesize,
    keywordstyle=\color{blue}\bfseries,
    commentstyle=\color{green},
    stringstyle=\color{red},
    numberstyle=\tiny\color{gray},
    showstringspaces=false,
    frame=single,
    breaklines=true,
    backgroundcolor=\color{lightgray!20}
}

\usepackage{hyperref}       %
\usepackage{url}            %
\usepackage{booktabs}       %
\usepackage{nicefrac}       %
\usepackage{microtype}      %
\usepackage{graphicx}
\usepackage{subcaption} 
\usepackage{wrapfig}
\usepackage{lipsum}
\usepackage{enumitem}
\usepackage{stackengine}
\usepackage[font=small,labelfont=bf]{caption}
\usepackage{color}
\usepackage{adjustbox}
\usepackage{tikz}
\usepackage{tcolorbox}
\usepackage{rotating}
\usepackage{makecell}
\usepackage{colortbl}    
\usepackage{multirow}    
\usepackage{pifont}      
\usepackage{xcolor}      
\usepackage{array}       
\usepackage{amsmath}
\usepackage{amsfonts}       %
\usepackage{xspace}
\usepackage{mathtools}
\definecolor{oursblue}{RGB}{230,240,255} 

\newcommand{\x}{\mathbf{x}}

\definecolor{blanchedalmond}{rgb}{1.0, 0.92, 0.8}
\definecolor{carmine}{rgb}{0.59, 0.0, 0.09}
\definecolor{lightblue}{rgb}{0.22,0.45,0.70}%

\renewcommand{\mathbf}{\boldsymbol}

\makeatletter
\def\Ddots{\mathinner{\mkern1mu\raise\p@
\vbox{\kern7\p@\hbox{.}}\mkern2mu
\raise4\p@\hbox{.}\mkern2mu\raise7\p@\hbox{.}\mkern1mu}}
\makeatother

\definecolor{amaranth}{rgb}{0.9, 0.17, 0.31}
\definecolor{antiquebrass}{rgb}{0.8, 0.58, 0.46}
\definecolor{antiquefuchsia}{rgb}{0.57, 0.36, 0.51}
\definecolor{chromeyellow}{rgb}{0.31, 0.47, 0.26}

\newtoggle{inTableHeader}
\toggletrue{inTableHeader}
\newcommand*{\StartTableHeader}{\global\toggletrue{inTableHeader}}%

\let\OldTabular\tabular%
\let\OldEndTabular\endtabular%
\renewenvironment{tabular}{\StartTableHeader\OldTabular}{\OldEndTabular\StartTableHeader}%

\usepackage{float}

\usepackage{multirow}
\usepackage{hhline}
\usepackage{amssymb}

\usepackage{caption}
\newtcolorbox{AIbox}[2][]{aibox,title=#2,#1}
\definecolor{lightblue}{rgb}{0.22,0.45,0.70}%
\definecolor{Gray}{gray}{0.95}
\definecolor{Cornsilk}{rgb}{1.0, 0.97, 0.86}
\usepackage{enumitem}

\makeatother

\definecolor{myred}{rgb}{0.7, 0.3, 0.0}
\definecolor{myblue}{HTML}{054488}
\definecolor{mygreen}{HTML}{056b34}
\definecolor{myorange}{HTML}{ff8800}
\definecolor{mypurple}{HTML}{8400ff}
\definecolor{mypink}{HTML}{f7acb9}

\definecolor{myred}{rgb}{0.7, 0.3, 0.0}
\definecolor{myblue}{HTML}{054488}
\definecolor{mygreen}{HTML}{056b34}
\definecolor{tiktokpink}{HTML}{E91E63}
\definecolor{tiktokpurple}{HTML}{673AB7}
\definecolor{tiktokgray}{HTML}{9E9E9E}

\newcommand{\mytitle}{\textsc{Mina}: A Multilingual LLM-Powered Legal Assistant Agent for Bangladesh for Empowering Access to Justice}

\usepackage{amssymb}

\title{\mytitle}

\runningtitle{\textsc{Mina}: AI Legal Assistant Agent for Bangladesh}

\author{
\vspace{2mm}
 \textbf{Azmine Toushik Wasi\textsuperscript{1,2,3}},
 \textbf{Wahid Faisal\textsuperscript{1,2,3}},
 \textbf{Mst Rafia Islam\textsuperscript{1,2,4}},
 \textbf{Md Rizwan Parvez\textsuperscript{5}}
}

\affil{
\textsuperscript{1}Computational Intelligence and Operations Laboratory
\textsuperscript{2}Cohere Labs Community\\
\vspace{-2.5mm}
\textsuperscript{3}Shahjalal University of Science and Technology
\textsuperscript{4}Independent University\\
\vspace{-2.5mm}
\textsuperscript{5}Qatar Computing Research Institute (QCRI)\\
}

\begin{document}

\begin{abstract}
\textbf{Abstract:} Bangladesh's low-income population faces major barriers to affordable legal advice due to complex legal language, procedural opacity, and high costs. Existing AI legal assistants lack Bengali-language support and jurisdiction-specific adaptation, limiting their effectiveness. To address this, we developed \textsc{Mina}, a multilingual LLM-based legal assistant tailored for the Bangladeshi context. It employs multilingual embeddings and a RAG-based chain-of-tools framework for retrieval, reasoning, translation, and document generation, delivering context-aware legal drafts, citations, and plain-language explanations via an interactive chat interface. Evaluated by law faculty from leading Bangladeshi universities across all stages of the 2022 and 2023 Bangladesh Bar Council Exams, \textsc{Mina} scored 75–80\% in Preliminary MCQs, Written, and simulated Viva Voce exams, matching or surpassing average human performance and demonstrating clarity, contextual understanding, and sound legal reasoning.
Even under a conservative upper bound, \textsc{Mina} operates at just 0.12–0.61\% of typical legal consultation costs in Bangladesh, yielding a 99.4–99.9\% cost reduction relative to human-provided services.
These results confirm its potential as a low-cost, multilingual AI assistant that automates key legal tasks and scales access to justice, offering a real-world case study on building domain-specific, low-resource systems and addressing challenges of multilingual adaptation, efficiency, and sustainable public-service AI deployment.
\vspace{0.5cm}

\coloremojicode{1F4C5} \textbf{Date}: April 08, 2026


\coloremojicode{1F4BB} \textbf{Code}: \href{https://github.com/ciol-researchlab/mina}{GitHub} \quad


\coloremojicode{1F4E7} \textbf{Correspondence}: Azmine Toushik Wasi~(\href{mailto:azmine32@student.sust.edu}{azmine32@student.sust.edu})
\end{abstract}

\maketitle
\begin{abstract}
\end{abstract}

 \section{Introduction}
Access to affordable legal advice remains a major challenge for economically disadvantaged groups in Bangladesh, largely due to complex legal language, opaque procedures, and high costs \citep{20egfesaw24,Raj202wfdf4,Islam2024afeasg,AKTER2017affaf,wai2024exp}. This contributes to a severe access to justice crisis that disproportionately affects marginalized populations. The judiciary is burdened by an enormous backlog of 3.7 to 4.4 million cases and operates with only 2,100 judges \citep{yesmin2025adr,cegagw43gar3}, translating to one judge per 90,000 people and resulting in delays of 10 to 60 years \citep{jaan2023justice}. Civil disputes often drag on for decades, while criminal cases like the Chittagong Arms Haul have remained unresolved for over 15 years \citep{20egfesaw24,khalilullah2025justice}. Outdated procedural laws enable frequent adjournments and exploitation by lawyers, further slowing case resolution. Meanwhile, unregulated lawyer fees make legal representation unaffordable for most \citep{jaan2023justice}, and public legal aid faces constraints from limited funding, strict eligibility, and  engagement \citep{tahura2025legalaid}. 

The absence of digital infrastructure and automation intensifies inefficiencies and delays, compounded by manual court processes. Additionally, widespread legal illiteracy, fear of retaliation, and dependence on biased informal mechanisms discourage many from seeking justice, perpetuating inequality and eroding trust in the system \citep{20egfesaw24,Islam2024afeasg,wai2024exp}. Consequently, many avoid formal legal engagement altogether, deepening existing disparities and weakening the rule of law \citep{20egfesaw24}. Natural Language Processing (NLP) tools can help bridge this gap by simplifying complex legal texts, automating document analysis, and providing accessible guidance to under-served populations.

Motivated by the urgent need to improve legal access for Bangladesh’s low-income population, who face significant barriers due to complex legal language, long procedures, and high costs, we developed \textsc{Mina}, a multilingual LLM-based legal assistant tailored to the Bangladeshi legal context. Unlike prior systems focused on English and Western legal frameworks, our assistant integrates Bengali and English support using multilingual embeddings within a Retrieval-Augmented Generation (RAG) framework that combines retrieval, reasoning, drafting, translation, and citation insertion. Delivered via an interactive chat interface, it assists users in drafting petitions, referencing statutes, and simplifying complex legal language into plain Bengali.

We evaluated our agentic system \textsc{Mina} rigorously by law faculty experts across all stages of the 2022 and 2023 Bangladesh Bar Council Examinations, including the Preliminary multiple-choice questions, Written exam, and Viva Voce oral exam (conducted via chat UI). Consistently scoring 75–80\%, it matched or exceeded average human performance, passing both years of the Bar Council exam and demonstrating robust legal reasoning and contextual understanding.

Overall, our contributions include: \textit{(i) the development of a localized multilingual legal assistant with integrated RAG, reasoning, drafting, translation, and citation capabilities; (ii) rigorous evaluation demonstrating human-comparable performance, robustness, and generalization across exam years, modalities, and diverse legal tasks; and (iii) passing the Bangladesh Bar Council exam, complemented by a comprehensive error analysis and insights for deploying scalable, cost-effective legal assistance to under-served populations.}

While our system uses established components like multilingual embeddings, RAG, and LangGraph-style agents, its novelty lies in adapting them to a bilingual, low-resource legal environment. The two-stage RAG pipeline retrieves statutes at the Act and Section levels, preventing conflation of unrelated provisions and ensuring coherent legal responses. Multilingual embeddings and a legal dictionary support interpretation of colonial-era and Farsi-influenced terminology, with external tools used selectively to augment context. This system-level design demonstrates practical deployment for accessible legal assistance, validated rigorously across multiple evaluation stages.

\section{Preliminaries}
Legal NLP is transforming legal practice by automating document analysis, contract review, and research, significantly reducing time and costs \citep{lai2023largelanguagemodelslaw,Yan2023,Frankenreiter_Nyarko_2023,zhong-etal-2020-nlp}, with transformer models like BERT and GPT enhancing contextual understanding for more accurate insights \citep{martin2024bettergptcomparinglarge,jiang-etal-2024-leveraging}. Large Language Models (LLMs) extend this potential by simplifying complex legal language, translating statutes and procedures into plain Bengali, and automating repetitive tasks such as document drafting and legal research, which is critical in a system where formal aid is underfunded and unaffordable for many \citep{safdie2025aiforjustice,jaan2023justice}. Multilingual LLM-powered agents capable of processing both Bengali and English legal texts offer accurate cross-lingual retrieval and can handle mixed-language documents prevalent in Bangladesh, thereby enhancing accessibility and efficiency. 

However, existing Legal NLP tools remain inadequate for Bangladesh due to linguistic, legal, and socio-economic complexities: Bengali support suffers from scarce annotated datasets, limited digital content, and underdeveloped tools for tasks like tokenization and NER \citep{wai2024exp}, while English-centric models struggle with Bengali’s complex morphology and script. Furthermore, Bangladesh’s legal system, rooted in colonial-era codes and Farsi-influenced terminology, poses additional challenges, as models lack jurisdiction-specific knowledge and cannot effectively interpret archaic legal Bengali \citep{mizan2021language,asjad2023bangla}. Compounding these technical issues are gendered digital divides, poor translation quality \citep{asjad2023bangla}, low digital literacy, and limited offline legal resources, underscoring the urgent need for a localized, context-aware AI solution tailored to Bangladesh’s unique legal environment. Such an AI system could democratize legal assistance, empower marginalized populations, and help alleviate the country’s severe access-to-justice crisis.

 \section{Methodology: \textsc{Mina}}

\begin{figure*}
    \centering
    \includegraphics[width=\linewidth]{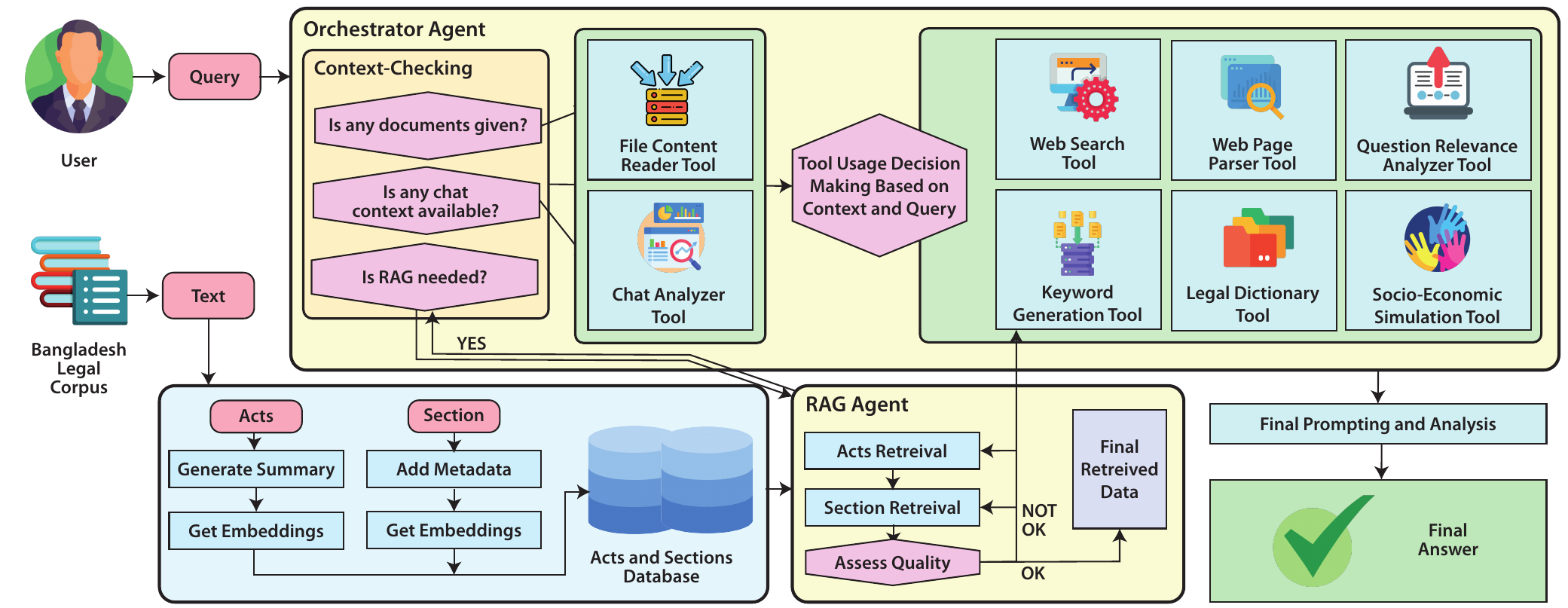}
    \caption{System Architecture and Workflow of our Multilingual Legal Assistant Agent for Bangladesh}
    \label{fig:archi}
\end{figure*}

Our system is anchored by an Orchestrator Agent that evaluates user input, chat history, and documents to select the appropriate response pathway. When internal context is insufficient, a two-stage RAG pipeline retrieves relevant Acts and Sections using Cohere-generated keywords and multilingual embeddings over Chroma vector stores of Bangladeshi laws. Retrieved texts inform final reasoning, while external tools (web search, parsers) are used only when necessary. The system handles both Bengali and English, interprets colonial-era and Farsi-influenced terms via a custom legal dictionary, and includes a socio-economic simulation module. Optimized for low-resource settings, it provides context-aware, jurisdiction-specific legal assistance via a simple chat interface, simplifying legal language and automating core tasks to empower users without legal training or access to affordable services.

 \subsection{Architecture Design}
This section details each component of the architecture, their interactions, and how they collectively address the core legal accessibility challenges in the region, as outlined in Figure \ref{fig:archi}.

\subsubsection{Retrieval-Augmented Generation (RAG)}
Our two-stage RAG subsystem retrieves legal texts and enhances response relevance through sequential Act and Section retrieval, using dedicated vectorDBs built from Bangladeshi legal corpora.

\noindent $\blacksquare$ \textbf{Data.}
In the Bangladeshi legal system, Acts are the primary legislative units, each containing an average of 24 Sections. Act titles average 50.30 characters, Section titles 38.07 characters, and Section content 736.69 characters, reflecting their greater interpretive depth. As per April 2025, the dataset contains 595 Acts comprising 18, 023 Sections (avg. 24 sections/act), with average act name and detail lengths of 50 and 438 characters, respectively, and average section name and detail lengths of 38 and 736 characters.

\noindent $\blacksquare$ \textbf{Database Development.}
Initial experiments showed that naïve retrieval often produced responses by combining content from unrelated Acts. To address this, we developed two separate vector databases. The Act database was constructed by collecting full-text legislative data from the official Bangladesh Law and Justice website. For each Act, we generated an LLM-based summary to create a compact, searchable representation, indexed alongside metadata. The Section database was built by chunking individual sections (1–2 chunks each) and attaching relevant Act metadata for precise filtering during Section-level search.

\noindent $\blacksquare$ \textbf{RAG Pipeline and Workflow.}
When a query is received, the Keyword Generator tool produces semantic keywords to retrieve the top $n_{acts}$ (default = 5) most relevant Acts from indexed summaries. These Act IDs then filter the Section database to find the most relevant $n_{sections}$ (default = 10) legal provisions. Multilingual embeddings ensure accurate retrieval across Bengali and English texts. Retrieved sections are checked for contextual relevance; if insufficient, the query is refined and rerun. Once validated, the Acts and Sections are sent to the Orchestrator Agent for final reasoning. This two-step process ensures broad coverage at the Act level and precise grounding at the Section level.

\subsubsection{Agents}
Our system is powered by two core agents: the \textit{Orchestrator Agent} and the \textit{RAG Agent}, each with distinct roles in the workflow.

\noindent $\blacksquare$ \textbf{Orchestrator Agent} serves as the central decision-maker, evaluating user queries in context, considering prior conversation turns and any uploaded documents, to determine if a direct response can be generated. If the context is sufficient, it proceeds to answer generation. Otherwise, it delegates the task to the RAG Agent.

\noindent $\blacksquare$ \textbf{RAG Agent} manages the retrieval process end-to-end, from keyword generation to assessing the relevance and jurisdictional appropriateness of retrieved legal materials. This separation of roles enhances system modularity and simplifies maintenance, debugging, and extensibility.

Both agents operate within a \href{https://www.langchain.com/langgraph}{\texttt{LangGraph}}-based state machine, enabling persistent memory across interactions. This design supports complex, multi-turn legal consultations where reasoning must evolve. The agentic structure also enables conditional execution, allowing the system to adapt its workflow dynamically without compromising performance or legal integrity.

\subsubsection{Tools}
Our system incorporates a suite of specialized tools, each tailored to handle specific preprocessing or auxiliary tasks critical to the overall workflow. These are:
\begin{enumerate}
    \item The \textbf{File Content Reader} processes uploaded documents in \textit{.pptx}, \textit{.docx}, and \textit{.pdf} formats, employing temporary file handling alongside format-specific parsers to extract clean, structured text for downstream use. 
    \item During the RAG phase, the \textbf{Keyword Generator} leverages a large language model (LLM) to produce 5–10 semantically meaningful search terms, with a regular expression fallback mechanism to ensure robustness in low-context scenarios. 
    \item To facilitate external information access, a DuckDuckGo-based \textbf{Web Search} tool retrieves top-ranked results, including titles, URLs, and content snippets. 
    \item When deeper inspection is required, the \textbf{Web Page Parser} employs \texttt{BeautifulSoup} to extract up to 5,000 characters of visible page content, filtering out scripts and styling artifacts. 
    \item For internal coherence, the \textbf{Question Relevance Analyzer} utilizes embeddings to detect semantic relationships between current and prior queries, thereby enhancing continuity in multi-turn interactions. 
    \item \textbf{Chat Analyzer} analyzes previous chat to get context in chat mode. 
    \item Legal interpretation in Bengali is supported by a custom \textbf{Legal Dictionary}, designed to explain colonial-era and Farsi-influenced terms prevalent in Bangladeshi legal texts. This tool enables plain-language explanations and improves the model’s ability to interpret complex expressions. 
    \item Furthermore, a \textbf{Socio-Economic Simulation} module enables exploration of how socio-demographic variables interact with legal access, supporting both diagnostic use cases and long-term policy planning.
\end{enumerate}

\subsubsection{LLM Integration and Prompts}
At the core of our system's understanding layer is a multilingual LLM. In our experiments, we
evaluate a broad spectrum of models spanning both proprietary and open-source families, including 
OpenAI's \textit{GPT-4o} \citep{openai2024gpt4ocard},
Google’s \textit{gemini-2.0-flash} (proprietary), \textit{gemini-2.5-flash} (proprietary) \citep{geminiteam2025geminifamilyhighlycapable}, \textit{gemma-3-4b-it}, \textit{gemma-3-12b-it}, \textit{gemma-3-27b-it} \citep{gemmateam2025gemma3technicalreport}, Meta's \textit{llama3.2-1b-instruct}, \textit{llama3.2-3b-instruct}, \textit{llama3.1-8b-instruct}, \textit{llama3.1-70b-instruct} \citep{grattafiori2024llama3herdmodels}, Cohere’s \textit{command-a-03-2025} \citep{cohere2025commandaenterprisereadylarge}, Qwen's \textit{qwen3-4b-instruct-2507}, and \textit{qwen3-30b-a3b-instruct-2507} \citep{yang2025qwen3technicalreport}. All models were evaluated across system modules for legal drafting, question answering, and prompt compliance.

For retrieval, we employ Cohere’s \textit{embed-multilingual-light-v3.0}, which supports semantic similarity in both Bengali and English, crucial for processing bilingual legal content in Bangladesh.
Prompt engineering ensures legal accuracy and jurisdictional relevance. Prompts are constructed using prior conversation turns, uploaded documents, and the RAG state. 
All prompts are detailed in Appendix \ref{sec:prompts}.

\subsection{Adaptive Agent Workflow}

Our system operates through two primary working modes: \textit{situational} and \textit{decisional}.\\
$\blacksquare$In the \textbf{situational} pathway, if a user uploads a document, it is routed through the Document Analyzer, which extracts and prepares relevant legal content for downstream use. Alternatively, when prior user interactions are available, the Chat Analyzer identifies related past queries to provide contextual grounding for the current request.\\
$\blacksquare$ In the \textbf{decisional} pathway, the system assesses whether retrieval-augmented generation (RAG) is necessary. When internal context, such as documents or chat history, is insufficient or absent, we trigger the RAG workflow. This involves generating semantic keywords, retrieving relevant vectors from pre-indexed legal corpora (e.g., Bangladeshi statutes and sections), and composing a contextually grounded response.\\
This flexible routing mechanism enables the agent to adapt intelligently based on the type of input, historical interactions, and the complexity of the query, ensuring both responsiveness and legal relevance.


\definecolor{closed_models}{RGB}{255, 245, 170}
\definecolor{open_models_small}{RGB}{195, 255, 194}
\definecolor{open_models_large}{RGB}{177, 252, 226}
\definecolor{reasoning_models-c}{RGB}{234, 193, 245}
\definecolor{reasoning_models-o}{RGB}{192, 168, 237}
\definecolor{spatial_specific_models}{RGB}{171, 229, 247}
\definecolor{human_baseline}{RGB}{255, 190, 168}

\begin{table*}[t]
\centering
\caption{Evaluation of our legal assistant agent \textsc{Mina} on BD Bar Council Exam: MCQ.
Scores are 5-run averages. Best in each category and setup are marked with \textbf{\underline{bold underline}}.}
\label{tab:main-mcq-accuracy}
\resizebox{\textwidth}{!}{%
\begin{tabular}{l|cccc|cccc}

\toprule

\multirow{2}{*}{\textbf{Model}} 
&
\multicolumn{4}{c|}{\textbf{Year: 2022}} 
&
\multicolumn{4}{c}{\textbf{Year: 2023}}\\
\cmidrule{2-9}
&
W/o RAG & Naive RAG & 2-Step RAG & Tools &
W/o RAG & Naive RAG & 2-Step RAG & Tools
\\

\midrule

\textbf{\textit{Random Choice}} & 
25.00 & 25.20 & 25.40 & 25.20 & 25.60 & 25.40 & 25.20 & 25.00 \\

\midrule
\multicolumn{8}{l}{\textbf{\textit{Proprietary Models}}} \\
\rowcolor{closed_models!50}GPT-4o&
18.60 & 62.40 & 69.20 & 73.60 &
19.20 & 58.80 & 67.80 & 72.20
\\
\rowcolor{closed_models!50}Gemini-2.0-Flash&
12.40 & 61.20 & 68.60 & 69.20 &
12.20 & 59.40 & 69.00 & 70.20
\\

\rowcolor{closed_models!50}Gemini-2.5-Flash  & 
\textbf{\underline{30.20}} & \textbf{\underline{68.80}} & \textbf{\underline{75.60}} & \textbf{\underline{77.00}} & \textbf{\underline{32.40}} & \textbf{\underline{69.20}} & \textbf{\underline{76.40}} & \textbf{\underline{77.00}}
\\

\midrule
\multicolumn{8}{l}{\textbf{\textit{Open-Source Models: Small}}} \\

\rowcolor{open_models_small!50} Llama3.2-1B-Instruct& 
6.20 & 6.00 & 6.40 & 8.20 & 7.00 & 7.20 & 7.40 & 9.20
\\
\rowcolor{open_models_small!50} Llama3.2-3B-Instruct& 
9.20 & 9.00 & 11.20 & 13.40 & 11.20 & 11.40 & 14.00 & 15.20
\\

\rowcolor{open_models_small!50} Gemma-3-4B-it  & 
12.40 & 15.20 & 20.60 & 23.20 & 14.40 & 16.60 & 22.20 & 26.40
\\
\rowcolor{open_models_small!50} Qwen3-4B-Instruct-2507  &  
\textbf{\underline{14.20}} & \textbf{\underline{28.40}} & 44.60 & \textbf{\underline{49.80}} & \textbf{\underline{16.20}} & \textbf{\underline{29.40}} & 45.40 & \textbf{\underline{52.40}}
\\
\rowcolor{open_models_small!50} Command-A-8B  & 
8.20 & 25.20 & \textbf{\underline{47.00}} & 47.40 &
11.20 & 23.40 & \textbf{\underline{49.20}} & 50.20
\\
\rowcolor{open_models_small!50} Llama3.1-8B-Instruct& 
15.20 & 18.40 & 30.20 & 32.40 & 16.20 & 18.60 & 33.40 & 35.20
\\

\midrule
\multicolumn{8}{l}{\textbf{\textit{Open-Source Models: Large}}} \\

\rowcolor{open_models_large!50} Gemma-3-12B-it  & 
20.40 & 35.20 & 48.40 & 50.20 & 22.60 & 36.20 & 52.40 & 54.20
 \\

\rowcolor{open_models_large!50} Gemma-3-27B-it  & 
28.20 & 45.40 & 60.20 & 64.40 & 30.20 & 46.60 & 62.40 & 66.20
 \\

\rowcolor{open_models_large!50} Qwen3-30B-A3B-Instruct-2507& 
\textbf{\underline{34.20}} & \textbf{\underline{50.40}} & \textbf{\underline{65.60}} & \textbf{\underline{70.80}} & \textbf{\underline{36.20}} & \textbf{\underline{52.40}} & \textbf{\underline{67.20}} & \textbf{\underline{72.40}}
\\

\rowcolor{open_models_large!50} Llama3.1-70B-Instruct& 
24.20 & 30.40 & 40.20 & 42.40 & 26.20 & 32.20 & 44.40 & 46.20
\\

\bottomrule
\end{tabular}
}
\end{table*}
\definecolor{titlebar}{RGB}{224, 224, 220}
\definecolor{closed_models}{RGB}{255, 245, 170}
\definecolor{open_models_small}{RGB}{195, 255, 194}
\definecolor{open_models_large}{RGB}{177, 252, 226}
\definecolor{reasoning_models-c}{RGB}{234, 193, 245}
\definecolor{reasoning_models-o}{RGB}{192, 168, 237}
\definecolor{spatial_specific_models}{RGB}{171, 229, 247}
\definecolor{human_baseline}{RGB}{255, 190, 168}

\begin{table*}[t]
\centering
\caption{Evaluation of our legal assistant agent \textsc{Mina} on BD Bar Council Exam: Written.
Scores are 5-evaluator averages. Best in each color-category and setup are marked with \textbf{\underline{bold underline}}.}
\label{tab:main-written-accuracy}
\resizebox{\textwidth}{!}{%
\begin{tabular}{l|cccc|cccc}

\toprule

\multirow{2}{*}{\textbf{Model}} 
&
\multicolumn{4}{c|}{\textbf{Year: 2022}} 
&
\multicolumn{4}{c}{\textbf{Year: 2023}}\\
\cmidrule{2-9}
&
W/o RAG & Naive RAG & 2-Step RAG & Tools &
W/o RAG & Naive RAG & 2-Step RAG & Tools
\\

\midrule
\multicolumn{8}{l}{\textbf{\textit{Proprietary Models}}} \\

\rowcolor{closed_models!50}GPT-4o &
19.20 & 55.40 & 69.80 & 71.20 & 21.20 & 60.80 & 72.60 & 75.60
\\
\rowcolor{closed_models!50}Gemini-2.0-Flash &
18.40 & 58.20 & 68.60 & 70.00 & 18.20 & 65.40 & 74.80 & 75.20
\\

\rowcolor{closed_models!50}Gemini-2.5-Flash  & 
\textbf{\underline{35.20}} & \textbf{\underline{70.40}} & \textbf{\underline{78.60}} & \textbf{\underline{81.00}} & \textbf{\underline{36.20}} & \textbf{\underline{71.20}} & \textbf{\underline{79.40}} & \textbf{\underline{81.80}}
\\

\midrule
\multicolumn{8}{l}{\textbf{\textit{Open-Source Models: Small}}} \\

\rowcolor{open_models_small!50} Llama3.2-1B-Instruct & 
6.20 & 6.00 & 6.20 & 7.00 & 7.00 & 7.00 & 7.20 & 8.00
\\
\rowcolor{open_models_small!50} Llama3.2-3B-Instruct & 
10.20 & 11.00 & 14.20 & 16.40 & 12.00 & 14.20 & 17.40 & 20.00
\\

\rowcolor{open_models_small!50} Gemma-3-4B-it  & 
15.20 & 20.20 & 24.40 & 28.00 & 18.20 & 22.40 & 26.00 & 30.20
\\
\rowcolor{open_models_small!50} Qwen3-4B-Instruct-2507  &  
22.20 & 56.40 & \textbf{\underline{71.60}} & 68.00 & 26.20 & 60.40 & 64.20 & 70.40
\\
\rowcolor{open_models_small!50} Command-A-8B  & 
25.00 & \textbf{\underline{60.20}} & 71.00 & \textbf{\underline{74.40}} & 25.20 & \textbf{\underline{73.00}} & \textbf{\underline{74.20}} & \textbf{\underline{76.00}}
\\
\rowcolor{open_models_small!50} Llama3.1-8B-Instruct & 
\textbf{\underline{28.20}} & 34.40 & 46.20 & 50.00 & \textbf{\underline{30.20}} & 66.00 & 68.20 & 71.20
\\

\midrule
\multicolumn{8}{l}{\textbf{\textit{Open-Source Models: Large}}} \\

\rowcolor{open_models_large!50} Gemma-3-12B-it  & 
30.00 & 42.20 & 55.40 & 58.00 & 32.20 & 44.40 & 58.00 & 60.20
 \\

\rowcolor{open_models_large!50} Gemma-3-27B-it  & 
38.20 & 55.00 & 68.20 & 72.40 & 40.00 & 56.20 & 70.00 & 74.20
 \\

\rowcolor{open_models_large!50} Qwen3-30B-A3B-Instruct-2507 & 
42.20 & 60.40 & 74.00 & 78.20 & 44.00 & 62.20 & 76.40 & 79.40
\\

\rowcolor{open_models_large!50} Llama3.1-70B-Instruct & 
\textbf{\underline{45.00}} & \textbf{\underline{62.20}} & \textbf{\underline{75.00}} & \textbf{\underline{79.80}} & \textbf{\underline{47.20}} & \textbf{\underline{65.00}} & \textbf{\underline{77.20}} & \textbf{\underline{80.20}}
\\

\bottomrule
\end{tabular}
}
\end{table*}
\definecolor{titlebar}{RGB}{224, 224, 220}
\definecolor{closed_models}{RGB}{255, 245, 170}
\definecolor{open_models_small}{RGB}{195, 255, 194}
\definecolor{open_models_large}{RGB}{177, 252, 226}
\definecolor{reasoning_models-c}{RGB}{234, 193, 245}
\definecolor{reasoning_models-o}{RGB}{192, 168, 237}
\definecolor{spatial_specific_models}{RGB}{171, 229, 247}
\definecolor{human_baseline}{RGB}{255, 190, 168}

\begin{table}[t]
\centering
\caption{Evaluation of Viva exam across different setups. Scores are 5-evaluator averages. Best in each category and setup are marked with \textbf{\underline{bold underline}}.}
\label{tab:viva-bose-result}
\begin{tabular}{l|cccc}
\toprule
\textbf{Model} & \textbf{W/o RAG} & \textbf{N. RAG} & \textbf{2-S RAG} & \textbf{Tools} \\
\midrule
\multicolumn{5}{l}{\textbf{Proprietary Models}} \\
\rowcolor{closed_models!50} Gemini-2.0 Flash & 32.80 & 60.80 & 74.40 & 76.60 \\
\rowcolor{closed_models!50} Gemini-2.5 Flash & \textbf{\underline{36.20}} & \textbf{\underline{70.40}} & \textbf{\underline{79.20}} & \textbf{\underline{81.00}} \\
\midrule
\multicolumn{5}{l}{\textbf{Open-Source Models: Small}} \\
\rowcolor{open_models_small!50} Llama3.2-1B-Instruct & 6.20 & 6.40 & 6.20 & 7.00 \\
\rowcolor{open_models_small!50} Llama3.2-3B-Instruct & 11.20 & 12.40 & 15.20 & 17.40 \\
\rowcolor{open_models_small!50} Gemma-3-4B-it & 16.20 & 50.40 & 65.20 & 69.20 \\
\rowcolor{open_models_small!50} Qwen3-4B-Instruct-2507 & 22.20 & \textbf{\underline{56.40}} & 67.20 & 70.20\\
\rowcolor{open_models_small!50} Command A & \textbf{\underline{27.80}} & 55.60 & \textbf{\underline{70.40}} & \textbf{\underline{71.20}} \\
\rowcolor{open_models_small!50} Llama3.1-8B-Instruct & 24.20 & 52.20 & 66.40 & 67.20 \\

\midrule
\multicolumn{5}{l}{\textbf{Open-Source Models: Large}} \\
\rowcolor{open_models_large!50} Gemma-3-12B-it & 31.20 & 43.20 & 56.40 & 59.20 \\
\rowcolor{open_models_large!50} Gemma-3-27B-it & 39.20 & 56.00 & 69.20 & 72.40 \\
\rowcolor{open_models_large!50} Qwen3-30B-A3B-Instruct & 42.20 & 61.20 & 75.20 & 79.40 \\
\rowcolor{open_models_large!50} Llama3.1-70B-Instruct & \textbf{\underline{46.00}} & \textbf{\underline{63.20}} & \textbf{\underline{77.20}} & \textbf{\underline{80.20}} \\
\bottomrule
\end{tabular}
\end{table}

 \section{Experiments and Evaluation}
 \subsection{Evaluation Setup}

To evaluate the system’s real-world utility as a legal reasoning agent, we benchmarked its performance against the Bangladesh Bar Council Examinations from 2022 and 2023. These exams represent a high-stakes national standard for entry into the legal profession and are conducted in three stages: \textit{Multiple Choice Questions (MCQ)}, \textit{Written}, and \textit{Viva Voce}. Our evaluation pipeline was designed to mirror this full progression.

Each exam instance was administered under conditions closely aligned with actual testing protocols. For the MCQ and written stages, the model was assessed using original exam questions. For MCQ, each setup and model combination is repeated 5 times and average is taken to reduce model variability. In the case of written and viva evaluations, we partnered with law faculty members and legal experts from leading Bangladeshi universities (names anonymized) who served as independent evaluators. Each written answer was assessed by at least two qualified judges, and the final score was determined by averaging the two, following Bar Council standards.

\subsection{Evaluation Metrics}

We simulate the real-world evaluation process of the Bangladesh Bar Council examination across three major components: Multiple Choice Questions (MCQs), Written Examination, and Viva Voce. This design ensures that model performance can be meaningfully compared to that of human candidates under authentic assessment conditions.

$\blacksquare$ \textbf{Multiple Choice Questions (MCQs):}
For the MCQ component, automatic marking was performed by comparing each model’s responses to the ground truth, replicating the Optical Mark Recognition (OMR)-based evaluation used in actual examinations. Scores were computed on a 100-point scale, where a minimum of 50\% is required for human candidates to qualify for the written stage according to the official Bar Council standard.

$\blacksquare$ \textbf{Written Exam:}
The written examination was evaluated through a human assessment framework aligned with the official marking criteria of the Bangladesh Bar Council. Performance was judged across four primary dimensions: accuracy, clarity, contextual understanding, and legal reasoning. Each evaluator applied a standardized rubric but provided a single composite score rather than separate ratings for each dimension, reflecting the holistic marking approach used in real examinations. The written section consisted of thirteen questions. The first eleven were organized into five sets, from which six questions were required to be answered, two from the first set and one from each of the remaining four, constituting ninety marks in total. The remaining two questions, forming Set F, were valued at ten marks, with candidates required to answer only one. For evaluation, the questions were selected based on the model’s confidence and performance consistency to ensure representative and fair benchmarking.

$\blacksquare$ \textbf{Viva Voce:}
The viva voce was conducted through a ChatGPT-like interactive platform that simulated the real-world oral examination setting. The same evaluative principles applied as in the written component, with a minimum average of 50\% required for human candidates to pass this stage. Evaluators assessed the model’s ability to sustain coherent, contextually appropriate, and legally accurate dialogue through natural, conversational exchanges resembling human viva sessions. Although the official Bar Council outcome for this stage is recorded simply as “pass” or “fail,” evaluators in our study assigned detailed scores out of 100 to provide granular feedback and enable comparative analysis. All scoring rubrics and thresholds were aligned with historical Bar Council evaluation criteria to ensure fairness, consistency, and comparability with human performance.

More details are available in Appendix \ref{sec:apx-extra-exp-eval}.

\subsection{Baselines and Human Performance}
To contextualize model performance, we compared results with actual candidate statistics from the 2022 and 2023 Bangladesh Bar Council Exams.

$\blacksquare$ \textbf{Multiple Choice Questions (MCQs):}
In 2022, only 25.86\% of candidates (10,527 out of 40,696) passed the MCQ stage. In 2023, the pass rate declined further to 17.96\% (6,229 out of 34,682). Although the passing threshold is set at just 50\%, these low success rates highlight the complexity and difficulty of the questions, which are often challenging to interpret and answer correctly.

$\blacksquare$ \textbf{Written Exam:}
The written stage includes 13 questions across six legal domains, with candidates required to answer seven in total. Human examinees typically achieve 40–60\% in this stage. In 2022, 53.94\% of candidates (5,533 out of 10,527) passed; in 2023, this figure declined to 44.21\% (2,754 out of 6,229). Our model consistently outperformed these benchmarks, demonstrating generalization and adaptability across exam years.

$\blacksquare$ \textbf{Viva Voce:}
The final stage evaluates legal articulation, judgment, and argumentative coherence. Pass rates for this stage are notably high, as the viva is essentially an oral extension of the written exam, and candidates who reach this point are typically well-prepared. In 2022, 96.65\% of candidates (5,348 out of 5,533) passed the viva, while in 2023, 97.25\% (2,973 out of 3,057; some participants were conditionally allowed) successfully cleared this stage.

\section{Results and Findings}
We evaluated system performance in four setups: without RAG, one stage RAG (without act based filtering), two step RAG, and with all tools.

$\blacksquare$ \textbf{Performance on Preliminary MCQ.}
Results in Table~\ref{tab:main-mcq-accuracy} reveal some patterns in model performance on the Bangladesh Bar Council MCQ exam. Closed-source proprietary models (Gemini family) maintain a clear advantage, with Gemini-2.5-Flash consistently outperforming all baselines across years and setups, particularly in the multi-step RAG + tools condition where it exceeds 77\%. Within open-source models, scale and architecture are decisive: smaller models like Llama3.2-1B and Gemma-3-4B-it barely surpass random-choice baselines, whereas large-scale Qwen3-30B achieves competitive performance, approaching proprietary levels under retrieval-augmented settings. Retrieval consistently improves results: naive RAG yields moderate gains, while 2-step RAG with tool integration substantially boosts performance, especially for large open-source models. Qwen3-30B shows steeper improvements than Gemma-27B or Llama-70B, highlighting the importance of data alignment and retrieval synergy for legal reasoning. Performance gaps between 2022 and 2023 remain small, indicating stability across cohorts rather than year-specific overfitting. These findings underscore both the promise and limitations of open-source LLMs in specialized legal reasoning: scaling helps, but architecture and training data choices remain critical. 

$\blacksquare$ \textbf{Performance on Written Exam.}
Written evaluation results in Table~\ref{tab:main-written-accuracy} reveal both parallels and divergences from the MCQ setting. Proprietary models dominate, with Gemini-2.5-Flash achieving the highest scores across years, consistently surpassing 80\% with tools. Open-source models demonstrate greater competitiveness in written tasks compared to MCQs: large-scale models like Llama3.1-70B and Qwen3-30B reach 79–81\% with multi-step RAG and tools. Smaller models, such as Command-A-8B and Qwen3-4B, show significant gains when retrieval is added, highlighting the benefit of structured augmentation. Year-to-year consistency indicates generalization across cohorts rather than memorization. Scaling advantages are more pronounced in written evaluation, where moving to 27B or 70B substantially boosts performance. These results suggest that open-source models, paired with robust RAG pipelines, can provide cost-effective, competitive legal reasoning, especially in Global South contexts like Bangladesh.

$\blacksquare$ \textbf{Performance on Viva Voce (conducted via Chat).}
Viva Voce evaluation (Table~\ref{tab:viva-bose-result}) shows clear trends across model classes and RAG setups. Proprietary models, especially Gemini-2.5 Flash, achieve top scores across all configurations, reaching 81.0 with tools. Among open-source models, scale correlates strongly with performance: small models like Llama3.2-1B score minimally, whereas Qwen3-30B and Llama3.1-70B exceed 75 with advanced RAG and tools. Retrieval significantly boosts performance, especially for smaller and mid-sized models, with naive or 2-step RAG increasing scores by 20–40 points. Tool integration further provides consistent, marginal gains for large models. Overall, large open-source models with retrieval and reasoning pipelines can approach proprietary performance, while small models remain underpowered, highlighting the importance of scale, domain-specific retrieval, and tool-assisted reasoning.

$\blacksquare$ \textbf{Comparison with Human Examinee Performance.}
Compared to human performance, the models demonstrate remarkable competitiveness across all three stages. In MCQs, Gemini-2.5 and Qwen3-30B consistently exceed the human pass threshold of 50\%, despite candidate pass rates as low as 17.96\% in 2023. In written exams, large open-source models reach 79–81\%, matching or slightly exceeding typical human averages (40–60\%). In the viva, optimized models score 70–76\%, below human near-certainty (>96\%) but still exhibiting legally coherent reasoning. These results suggest that LLMs can surpass most candidates in knowledge-intensive stages and approach human-like standards in oral reasoning, raising policy-relevant questions about AI-augmented professional legal assessment and the potential for AI to approach elite human performance, particularly in text-based evaluations.

Detailed analysis is available in Appendix \ref{sec:apx-ExtendedAnalysis}.

\section{Error Analysis}
\paragraph{$\blacksquare$ MCQ.} 
As detailed in Appendix~\ref{sec:apx-error-analysis-mcq}, systematic errors reveal that the model often misinterprets Bengali conjunctions (e.g., “O” (and) vs. “ba”(or)), flattening nuanced legal semantics. Jurisdictional hierarchy errors indicate weak integration of procedural layering with lexical parsing, while doctrinal misclassifications (e.g., \textit{Res Judicata} under Section~151) expose failures in mapping abstract concepts to codified provisions. Intra-order confusions (Order~1, Rules~8 vs.~13) further suggest reliance on surface cues over legal reasoning. These patterns show the need for structured legal knowledge, hierarchical reasoning, and fine-grained disambiguation to enhance procedural accuracy.

\paragraph{$\blacksquare$ Written.}
As detailed in Appendix~\ref{sec:apx-error-analysis-written}, our analysis of model performance on the 2022–2023 Bangladesh Bar Council examination datasets reveals systematic errors stemming from both knowledge gaps and reasoning limitations. The model frequently uses imprecise legal terminology, such as “injury” instead of “damage” under the Specific Relief Act, 1877, omits key statutory conditions, and produces factually incorrect responses, including denying the existence of general exceptions in the Penal Code, 1860. It also conflates civil and criminal law and occasionally hallucinates, generating fabricated cases like \textit{Deowaney Mokdama v. Hazirawala and Garahazira} with spurious procedural details. These errors arise from limited exposure to legal corpora, a fluency-biased generation mechanism, inadequate multi-step reasoning, contextual misinterpretation, and inherent hallucination tendencies in language models. Mitigation strategies include fine-tuning on authoritative legal texts, implementing explicit reasoning frameworks to reduce omissions and domain conflation, and integrating verification against statutory and case law databases to improve factual accuracy. Collectively, these measures aim to enhance the interpretive fidelity and reliability of AI-generated legal responses and highlight the challenges of deploying large language models in high-stakes legal assessment contexts.

\section{Evaluator Response Analysis}
\paragraph{$\blacksquare$ Written.} As summarized in Appendix~\ref{sec:apx-EvaluatorResponseAnalysis-written}, evaluators found that large retrieval-augmented models produced coherent, exam-style answers following the IRAC pattern, with strong statutory recall and clear drafting. Their responses often mirrored human exam scripts, combining issue identification, legal statement, and concise application. However, omissions of procedural prerequisites, such as conditions for injunctions or receiverships, were frequent and penalized as substantive errors. Arithmetic failures in limitation-period calculations and incomplete synthesis across the CPC, Evidence Act, and Specific Relief Act were also noted. Evaluators highlighted bilingual clarity and structured reasoning as major strengths but criticized inconsistent statutory precision and occasional hallucinated case citations. Overall, the written analysis shows exam-ready compositional fluency yet emphasizes the need for deterministic procedural calculators, statutory-fusion modules, and fine-tuning on annotated bilingual exam data to achieve distinction-level precision.

\paragraph{$\blacksquare$ Viva.} As detailed in Appendix~\ref{sec:apx-EvaluatorResponseAnalysis-viva}, viva evaluations revealed that \textsc{Mina} performed with composure, clear diction, and accurate statutory articulation under conversational pressure. Examiners praised its coherence and ability to paraphrase complex doctrines into accessible explanations, though it sometimes failed to adapt reasoning when factual variations were introduced. A key weakness was the lack of clarifying questions and the flattening of hierarchical reasoning, leading to incorrect procedural advice in dynamic exchanges. Rapid doctrinal checks exposed occasional section misattributions, which evaluators treated as major oral faults. Feedback emphasized that distinction-level performance requires interpretive flexibility, acknowledgment of uncertainty, and deeper policy reasoning beyond rule recital. System-level remedies include integrating a clarification policy, dialogue-state tracking, and real-time procedural validation to enhance adaptive accuracy and examiner trust during live oral assessment.

\section{Cost-per-Query Comparison and Affordability Implications}
A central design requirement for access-to-justice systems in low-income settings is \emph{economic sustainability}. Beyond accuracy and legal soundness, the feasibility of large-scale deployment is primarily determined by per-query inference cost. We therefore compare the approximate cost-per-query of representative proprietary and open-weight model families: \textit{Gemini}, \textit{Qwen}, and \textit{LLaMA}, using empirical averages estimated from our prior experimental runs.

For short, structured MCQ-style queries, which typically require one to three tool calls and limited token generation, the estimated per-query cost lies in the range of 0.2–0.4 cents for Qwen (approximately 0.24–0.49 BDT), 0.3–0.5 cents for LLaMA (approximately 0.37–0.61 BDT), and 0.4–0.6 cents for Gemini (approximately 0.49–0.73 BDT). For longer written or constructed-response (CQ-type) queries, entailing two to four tool calls and substantially higher token usage, the corresponding costs increase to roughly 0.8–1.4 cents for Qwen (approximately 0.98–1.71 BDT), 1.0–1.6 cents for LLaMA (approximately 1.22–1.95 BDT), and 1.4–2.0 cents for Gemini (approximately 1.71–2.44 BDT). Across all configurations, the computational overhead of non-LLM tools, such as retrieval, translation, and formatting components, contributes negligibly to overall cost relative to LLM inference.

Even under a deliberately conservative upper-bound scenario involving a multi-turn interaction costing as much as 10 cents in total (approximately 12.2 BDT), the expense corresponds to only about 0.12\%–0.61\% of the minimum prevailing cost of basic legal advice in Bangladesh, which typically ranges from 2,000 to 10,000 BDT per consultation. Equivalently, using the AI agent represents a cost reduction of approximately 99.4\%–99.9\% relative to traditional human-provided legal services.
When embedded within \textsc{Mina}'s multilingual, RAG-based chain-of-tools framework, these cost characteristics demonstrate that high-quality, jurisdiction-specific legal assistance can be delivered at a price point compatible with public-service deployment, reinforcing the practicality of scaling LLM-based systems to expand access to justice in low-resource settings.

\section{Concluding Remarks}
In this paper, we present \textsc{Mina}, a multilingual LLM-based legal assistant tailored to the Bangladeshi legal context, designed to improve access to justice for low-income and linguistically diverse populations. By integrating two-stage RAG, multilingual embeddings, and selective tool augmentation, the system delivers legally coherent, context-aware responses in both Bengali and English across multiple exam modalities, including MCQs, written, and oral viva assessments. Even mid-sized open-source models approach or exceed average human performance, while top-tier models like Gemini-2.5 Flash achieve near-ceiling results. The system’s modular and lightweight architecture enables efficient deployment in resource-constrained environments, while its bilingual and dictionary-enhanced capabilities allow accurate interpretation of colonial-era and Farsi-influenced terminology. Detailed error analysis reveals challenges in hierarchical procedural reasoning, timeline computation, and adaptive oral exchanges, highlighting opportunities for structured procedural modules, real-time verification, and dialogue-state tracking. Our findings suggest that retrieval alone is insufficient for high-stakes legal tasks; targeted fine-tuning, contrastive training, and interactive clarification policies can further improve reliability. Overall, \textsc{Mina} demonstrates that careful system design, rigorous evaluation, and domain-specific adaptation enable LLMs to provide practical, scalable, and cost-effective legal assistance. Beyond Bangladesh, this approach offers a template for low-resource, multilingual legal environments worldwide, illustrating the potential of AI to enhance equitable access to legal knowledge and support professional legal workflows while maintaining human oversight.


\section*{Limitations}
While our evaluation demonstrates notable performance gains through retrieval and generation strategies, several limitations remain. Retrieval quality is highly dependent on the underlying corpus; noisy or misaligned documents can still mislead even robust pipelines. Although strategies like Two Step RAG improve performance, they introduce additional latency and complexity that may not scale well in real-time systems.

\section*{Ethical Considerations}
We adhered to all ethical guidelines outlined by the Association for Computational Linguistics (ACL) throughout this study. All data used in our benchmark were either publicly available or ethically sourced with appropriate permissions where required. No personally identifiable information (PII) was collected, stored, or used in the experiments. Human annotators involved in dataset construction were fairly compensated and provided with clear task instructions. We ensured transparency in our evaluation pipeline and made efforts to avoid biased or culturally insensitive content. Furthermore, model outputs were reviewed to identify and mitigate potential ethical harms. All experiments were conducted in accordance with ACL’s code of ethics regarding research integrity, fairness, and respect for contributors.

\section*{Potential Risks} 

While \textsc{Mina} is designed as a supportive tool for legal professionals, legal aid workers, and exam preparation, it is not infallible, and incorrect outputs remain possible, particularly when prompts are ambiguous, incomplete, or involve highly specialized legal scenarios. Erroneous advice, if relied upon without verification, could lead to misinformed decisions, procedural errors, or unintended legal consequences, especially in high-stakes cases. Liability for any legal action remains with the human professional using the system, consistent with standard practices for legal research platforms and drafting tools. There is also a risk of over-reliance, where users might defer critical judgment to the AI, potentially reducing diligence or critical thinking. To mitigate these risks, \textsc{Mina} should always be used under human supervision, and outputs must be cross-checked against authoritative statutes, case law, and professional guidance. On the positive side, when properly integrated, the system can enhance lawyer productivity, reduce research time, improve accessibility for underserved populations, and support more equitable legal services by lowering operational costs and enabling broader dissemination of legal knowledge. Overall, careful deployment with oversight and clear guidelines is essential to maximize benefits while minimizing potential harms.

\section*{Reproducibility Statement}
To ensure full reproducibility of our experiments, all code, trained retrieval indices, and processed datasets will be made publicly available upon acceptance. Detailed instructions for setting up the environment, running the Orchestrator Agent, executing the two-stage RAG pipeline, and reproducing all evaluations for MCQs, written exams, and viva voce are included in the paper and repository. Additionally, prompts, scoring rubrics, and evaluation protocols are documented to allow independent verification of both quantitative results and qualitative analyses.

\section*{Acknowledgements}
We sincerely thank Cohere for providing computing credits through the Cohere Labs Research Grant, which were instrumental in developing Knowledge Explorer. We are deeply grateful to Cohere Labs for their continued mentorship and for organizing Aya Expedition 2.0, which offered an excellent environment for experimentation, collaboration, and constructive feedback.

We are deeply grateful to \href{https://scholar.google.com/citations?user=j4cOSzAAAAAJ}{Julia Kreutzer} (\textit{Senior Research Scientist at Cohere Labs}) for her generous mentorship and guidance. Her insight, encouragement, and constructive feedback were invaluable throughout the conception, development, and refinement of \textsc{Mina}, shaping both its technical direction and research depth. 

We also thank Alejandro Rodríguez Salamanca, Marzieh Fadaee, Sara Hooker, Madeline Smith, and Brittawnya Prince for their insightful discussions, encouragement, and support during the expedition.

\clearpage
\bibliography{main}

@misc{ariai2025naturallanguageprocessinglegal,
      title={Natural Language Processing for the Legal Domain: A Survey of Tasks, Datasets, Models, and Challenges}, 
      author={Farid Ariai and Gianluca Demartini},
      year={2025},
      eprint={2410.21306},
      archivePrefix={arXiv},
      primaryClass={cs.CL},
      url={https://arxiv.org/abs/2410.21306}, 
}

@misc{grattafiori2024llama3herdmodels,
      title={The Llama 3 Herd of Models}, 
      author={Aaron Grattafiori and Abhimanyu Dubey and Abhinav Jauhri and Abhinav Pandey and Abhishek Kadian and Ahmad Al-Dahle and Aiesha Letman and Akhil Mathur and Alan Schelten and Alex Vaughan and Amy Yang and Angela Fan and Anirudh Goyal and Anthony Hartshorn and Aobo Yang and Archi Mitra and Archie Sravankumar and Artem Korenev and Arthur Hinsvark and Arun Rao and Aston Zhang and Aurelien Rodriguez and Austen Gregerson and Ava Spataru and Baptiste Roziere and Bethany Biron and Binh Tang and Bobbie Chern and Charlotte Caucheteux and Chaya Nayak and Chloe Bi and Chris Marra and Chris McConnell and Christian Keller and Christophe Touret and Chunyang Wu and Corinne Wong and Cristian Canton Ferrer and Cyrus Nikolaidis and Damien Allonsius and Daniel Song and Danielle Pintz and Danny Livshits and Danny Wyatt and David Esiobu and Dhruv Choudhary and Dhruv Mahajan and Diego Garcia-Olano and Diego Perino and Dieuwke Hupkes and Egor Lakomkin and Ehab AlBadawy and Elina Lobanova and Emily Dinan and Eric Michael Smith and Filip Radenovic and Francisco Guzmán and Frank Zhang and Gabriel Synnaeve and Gabrielle Lee and Georgia Lewis Anderson and Govind Thattai and Graeme Nail and Gregoire Mialon and Guan Pang and Guillem Cucurell and Hailey Nguyen and Hannah Korevaar and Hu Xu and Hugo Touvron and Iliyan Zarov and Imanol Arrieta Ibarra and Isabel Kloumann and Ishan Misra and Ivan Evtimov and Jack Zhang and Jade Copet and Jaewon Lee and Jan Geffert and Jana Vranes and Jason Park and Jay Mahadeokar and Jeet Shah and Jelmer van der Linde and Jennifer Billock and Jenny Hong and Jenya Lee and Jeremy Fu and Jianfeng Chi and Jianyu Huang and Jiawen Liu and Jie Wang and Jiecao Yu and Joanna Bitton and Joe Spisak and Jongsoo Park and Joseph Rocca and Joshua Johnstun and Joshua Saxe and Junteng Jia and Kalyan Vasuden Alwala and Karthik Prasad and Kartikeya Upasani and Kate Plawiak and Ke Li and Kenneth Heafield and Kevin Stone and Khalid El-Arini and Krithika Iyer and Kshitiz Malik and Kuenley Chiu and Kunal Bhalla and Kushal Lakhotia and Lauren Rantala-Yeary and Laurens van der Maaten and Lawrence Chen and Liang Tan and Liz Jenkins and Louis Martin and Lovish Madaan and Lubo Malo and Lukas Blecher and Lukas Landzaat and Luke de Oliveira and Madeline Muzzi and Mahesh Pasupuleti and Mannat Singh and Manohar Paluri and Marcin Kardas and Maria Tsimpoukelli and Mathew Oldham and Mathieu Rita and Maya Pavlova and Melanie Kambadur and Mike Lewis and Min Si and Mitesh Kumar Singh and Mona Hassan and Naman Goyal and Narjes Torabi and Nikolay Bashlykov and Nikolay Bogoychev and Niladri Chatterji and Ning Zhang and Olivier Duchenne and Onur Çelebi and Patrick Alrassy and Pengchuan Zhang and Pengwei Li and Petar Vasic and Peter Weng and Prajjwal Bhargava and Pratik Dubal and Praveen Krishnan and Punit Singh Koura and Puxin Xu and Qing He and Qingxiao Dong and Ragavan Srinivasan and Raj Ganapathy and Ramon Calderer and Ricardo Silveira Cabral and Robert Stojnic and Roberta Raileanu and Rohan Maheswari and Rohit Girdhar and Rohit Patel and Romain Sauvestre and Ronnie Polidoro and Roshan Sumbaly and Ross Taylor and Ruan Silva and Rui Hou and Rui Wang and Saghar Hosseini and Sahana Chennabasappa and Sanjay Singh and Sean Bell and Seohyun Sonia Kim and Sergey Edunov and Shaoliang Nie and Sharan Narang and Sharath Raparthy and Sheng Shen and Shengye Wan and Shruti Bhosale and Shun Zhang and Simon Vandenhende and Soumya Batra and Spencer Whitman and Sten Sootla and Stephane Collot and Suchin Gururangan and Sydney Borodinsky and Tamar Herman and Tara Fowler and Tarek Sheasha and Thomas Georgiou and Thomas Scialom and Tobias Speckbacher and Todor Mihaylov and Tong Xiao and Ujjwal Karn and Vedanuj Goswami and Vibhor Gupta and Vignesh Ramanathan and Viktor Kerkez and Vincent Gonguet and Virginie Do and Vish Vogeti and Vítor Albiero and Vladan Petrovic and Weiwei Chu and Wenhan Xiong and Wenyin Fu and Whitney Meers and Xavier Martinet and Xiaodong Wang and Xiaofang Wang and Xiaoqing Ellen Tan and Xide Xia and Xinfeng Xie and Xuchao Jia and Xuewei Wang and Yaelle Goldschlag and Yashesh Gaur and Yasmine Babaei and Yi Wen and Yiwen Song and Yuchen Zhang and Yue Li and Yuning Mao and Zacharie Delpierre Coudert and Zheng Yan and Zhengxing Chen and Zoe Papakipos and Aaditya Singh and Aayushi Srivastava and Abha Jain and Adam Kelsey and Adam Shajnfeld and Adithya Gangidi and Adolfo Victoria and Ahuva Goldstand and Ajay Menon and Ajay Sharma and Alex Boesenberg and Alexei Baevski and Allie Feinstein and Amanda Kallet and Amit Sangani and Amos Teo and Anam Yunus and Andrei Lupu and Andres Alvarado and Andrew Caples and Andrew Gu and Andrew Ho and Andrew Poulton and Andrew Ryan and Ankit Ramchandani and Annie Dong and Annie Franco and Anuj Goyal and Aparajita Saraf and Arkabandhu Chowdhury and Ashley Gabriel and Ashwin Bharambe and Assaf Eisenman and Azadeh Yazdan and Beau James and Ben Maurer and Benjamin Leonhardi and Bernie Huang and Beth Loyd and Beto De Paola and Bhargavi Paranjape and Bing Liu and Bo Wu and Boyu Ni and Braden Hancock and Bram Wasti and Brandon Spence and Brani Stojkovic and Brian Gamido and Britt Montalvo and Carl Parker and Carly Burton and Catalina Mejia and Ce Liu and Changhan Wang and Changkyu Kim and Chao Zhou and Chester Hu and Ching-Hsiang Chu and Chris Cai and Chris Tindal and Christoph Feichtenhofer and Cynthia Gao and Damon Civin and Dana Beaty and Daniel Kreymer and Daniel Li and David Adkins and David Xu and Davide Testuggine and Delia David and Devi Parikh and Diana Liskovich and Didem Foss and Dingkang Wang and Duc Le and Dustin Holland and Edward Dowling and Eissa Jamil and Elaine Montgomery and Eleonora Presani and Emily Hahn and Emily Wood and Eric-Tuan Le and Erik Brinkman and Esteban Arcaute and Evan Dunbar and Evan Smothers and Fei Sun and Felix Kreuk and Feng Tian and Filippos Kokkinos and Firat Ozgenel and Francesco Caggioni and Frank Kanayet and Frank Seide and Gabriela Medina Florez and Gabriella Schwarz and Gada Badeer and Georgia Swee and Gil Halpern and Grant Herman and Grigory Sizov and Guangyi and Zhang and Guna Lakshminarayanan and Hakan Inan and Hamid Shojanazeri and Han Zou and Hannah Wang and Hanwen Zha and Haroun Habeeb and Harrison Rudolph and Helen Suk and Henry Aspegren and Hunter Goldman and Hongyuan Zhan and Ibrahim Damlaj and Igor Molybog and Igor Tufanov and Ilias Leontiadis and Irina-Elena Veliche and Itai Gat and Jake Weissman and James Geboski and James Kohli and Janice Lam and Japhet Asher and Jean-Baptiste Gaya and Jeff Marcus and Jeff Tang and Jennifer Chan and Jenny Zhen and Jeremy Reizenstein and Jeremy Teboul and Jessica Zhong and Jian Jin and Jingyi Yang and Joe Cummings and Jon Carvill and Jon Shepard and Jonathan McPhie and Jonathan Torres and Josh Ginsburg and Junjie Wang and Kai Wu and Kam Hou U and Karan Saxena and Kartikay Khandelwal and Katayoun Zand and Kathy Matosich and Kaushik Veeraraghavan and Kelly Michelena and Keqian Li and Kiran Jagadeesh and Kun Huang and Kunal Chawla and Kyle Huang and Lailin Chen and Lakshya Garg and Lavender A and Leandro Silva and Lee Bell and Lei Zhang and Liangpeng Guo and Licheng Yu and Liron Moshkovich and Luca Wehrstedt and Madian Khabsa and Manav Avalani and Manish Bhatt and Martynas Mankus and Matan Hasson and Matthew Lennie and Matthias Reso and Maxim Groshev and Maxim Naumov and Maya Lathi and Meghan Keneally and Miao Liu and Michael L. Seltzer and Michal Valko and Michelle Restrepo and Mihir Patel and Mik Vyatskov and Mikayel Samvelyan and Mike Clark and Mike Macey and Mike Wang and Miquel Jubert Hermoso and Mo Metanat and Mohammad Rastegari and Munish Bansal and Nandhini Santhanam and Natascha Parks and Natasha White and Navyata Bawa and Nayan Singhal and Nick Egebo and Nicolas Usunier and Nikhil Mehta and Nikolay Pavlovich Laptev and Ning Dong and Norman Cheng and Oleg Chernoguz and Olivia Hart and Omkar Salpekar and Ozlem Kalinli and Parkin Kent and Parth Parekh and Paul Saab and Pavan Balaji and Pedro Rittner and Philip Bontrager and Pierre Roux and Piotr Dollar and Polina Zvyagina and Prashant Ratanchandani and Pritish Yuvraj and Qian Liang and Rachad Alao and Rachel Rodriguez and Rafi Ayub and Raghotham Murthy and Raghu Nayani and Rahul Mitra and Rangaprabhu Parthasarathy and Raymond Li and Rebekkah Hogan and Robin Battey and Rocky Wang and Russ Howes and Ruty Rinott and Sachin Mehta and Sachin Siby and Sai Jayesh Bondu and Samyak Datta and Sara Chugh and Sara Hunt and Sargun Dhillon and Sasha Sidorov and Satadru Pan and Saurabh Mahajan and Saurabh Verma and Seiji Yamamoto and Sharadh Ramaswamy and Shaun Lindsay and Shaun Lindsay and Sheng Feng and Shenghao Lin and Shengxin Cindy Zha and Shishir Patil and Shiva Shankar and Shuqiang Zhang and Shuqiang Zhang and Sinong Wang and Sneha Agarwal and Soji Sajuyigbe and Soumith Chintala and Stephanie Max and Stephen Chen and Steve Kehoe and Steve Satterfield and Sudarshan Govindaprasad and Sumit Gupta and Summer Deng and Sungmin Cho and Sunny Virk and Suraj Subramanian and Sy Choudhury and Sydney Goldman and Tal Remez and Tamar Glaser and Tamara Best and Thilo Koehler and Thomas Robinson and Tianhe Li and Tianjun Zhang and Tim Matthews and Timothy Chou and Tzook Shaked and Varun Vontimitta and Victoria Ajayi and Victoria Montanez and Vijai Mohan and Vinay Satish Kumar and Vishal Mangla and Vlad Ionescu and Vlad Poenaru and Vlad Tiberiu Mihailescu and Vladimir Ivanov and Wei Li and Wenchen Wang and Wenwen Jiang and Wes Bouaziz and Will Constable and Xiaocheng Tang and Xiaojian Wu and Xiaolan Wang and Xilun Wu and Xinbo Gao and Yaniv Kleinman and Yanjun Chen and Ye Hu and Ye Jia and Ye Qi and Yenda Li and Yilin Zhang and Ying Zhang and Yossi Adi and Youngjin Nam and Yu and Wang and Yu Zhao and Yuchen Hao and Yundi Qian and Yunlu Li and Yuzi He and Zach Rait and Zachary DeVito and Zef Rosnbrick and Zhaoduo Wen and Zhenyu Yang and Zhiwei Zhao and Zhiyu Ma},
      year={2024},
      eprint={2407.21783},
      archivePrefix={arXiv},
      primaryClass={cs.AI},
      url={https://arxiv.org/abs/2407.21783}, 
}

@misc{yang2025qwen3technicalreport,
      title={Qwen3 Technical Report}, 
      author={An Yang and Anfeng Li and Baosong Yang and Beichen Zhang and Binyuan Hui and Bo Zheng and Bowen Yu and Chang Gao and Chengen Huang and Chenxu Lv and Chujie Zheng and Dayiheng Liu and Fan Zhou and Fei Huang and Feng Hu and Hao Ge and Haoran Wei and Huan Lin and Jialong Tang and Jian Yang and Jianhong Tu and Jianwei Zhang and Jianxin Yang and Jiaxi Yang and Jing Zhou and Jingren Zhou and Junyang Lin and Kai Dang and Keqin Bao and Kexin Yang and Le Yu and Lianghao Deng and Mei Li and Mingfeng Xue and Mingze Li and Pei Zhang and Peng Wang and Qin Zhu and Rui Men and Ruize Gao and Shixuan Liu and Shuang Luo and Tianhao Li and Tianyi Tang and Wenbiao Yin and Xingzhang Ren and Xinyu Wang and Xinyu Zhang and Xuancheng Ren and Yang Fan and Yang Su and Yichang Zhang and Yinger Zhang and Yu Wan and Yuqiong Liu and Zekun Wang and Zeyu Cui and Zhenru Zhang and Zhipeng Zhou and Zihan Qiu},
      year={2025},
      eprint={2505.09388},
      archivePrefix={arXiv},
      primaryClass={cs.CL},
      url={https://arxiv.org/abs/2505.09388}, 
}

@misc{gemmateam2025gemma3technicalreport,
      title={Gemma 3 Technical Report}, 
      author={Gemma Team and Aishwarya Kamath and Johan Ferret and Shreya Pathak and Nino Vieillard and Ramona Merhej and Sarah Perrin and Tatiana Matejovicova and Alexandre Ramé and Morgane Rivière and Louis Rouillard and Thomas Mesnard and Geoffrey Cideron and Jean-bastien Grill and Sabela Ramos and Edouard Yvinec and Michelle Casbon and Etienne Pot and Ivo Penchev and Gaël Liu and Francesco Visin and Kathleen Kenealy and Lucas Beyer and Xiaohai Zhai and Anton Tsitsulin and Robert Busa-Fekete and Alex Feng and Noveen Sachdeva and Benjamin Coleman and Yi Gao and Basil Mustafa and Iain Barr and Emilio Parisotto and David Tian and Matan Eyal and Colin Cherry and Jan-Thorsten Peter and Danila Sinopalnikov and Surya Bhupatiraju and Rishabh Agarwal and Mehran Kazemi and Dan Malkin and Ravin Kumar and David Vilar and Idan Brusilovsky and Jiaming Luo and Andreas Steiner and Abe Friesen and Abhanshu Sharma and Abheesht Sharma and Adi Mayrav Gilady and Adrian Goedeckemeyer and Alaa Saade and Alex Feng and Alexander Kolesnikov and Alexei Bendebury and Alvin Abdagic and Amit Vadi and András György and André Susano Pinto and Anil Das and Ankur Bapna and Antoine Miech and Antoine Yang and Antonia Paterson and Ashish Shenoy and Ayan Chakrabarti and Bilal Piot and Bo Wu and Bobak Shahriari and Bryce Petrini and Charlie Chen and Charline Le Lan and Christopher A. Choquette-Choo and CJ Carey and Cormac Brick and Daniel Deutsch and Danielle Eisenbud and Dee Cattle and Derek Cheng and Dimitris Paparas and Divyashree Shivakumar Sreepathihalli and Doug Reid and Dustin Tran and Dustin Zelle and Eric Noland and Erwin Huizenga and Eugene Kharitonov and Frederick Liu and Gagik Amirkhanyan and Glenn Cameron and Hadi Hashemi and Hanna Klimczak-Plucińska and Harman Singh and Harsh Mehta and Harshal Tushar Lehri and Hussein Hazimeh and Ian Ballantyne and Idan Szpektor and Ivan Nardini and Jean Pouget-Abadie and Jetha Chan and Joe Stanton and John Wieting and Jonathan Lai and Jordi Orbay and Joseph Fernandez and Josh Newlan and Ju-yeong Ji and Jyotinder Singh and Kat Black and Kathy Yu and Kevin Hui and Kiran Vodrahalli and Klaus Greff and Linhai Qiu and Marcella Valentine and Marina Coelho and Marvin Ritter and Matt Hoffman and Matthew Watson and Mayank Chaturvedi and Michael Moynihan and Min Ma and Nabila Babar and Natasha Noy and Nathan Byrd and Nick Roy and Nikola Momchev and Nilay Chauhan and Noveen Sachdeva and Oskar Bunyan and Pankil Botarda and Paul Caron and Paul Kishan Rubenstein and Phil Culliton and Philipp Schmid and Pier Giuseppe Sessa and Pingmei Xu and Piotr Stanczyk and Pouya Tafti and Rakesh Shivanna and Renjie Wu and Renke Pan and Reza Rokni and Rob Willoughby and Rohith Vallu and Ryan Mullins and Sammy Jerome and Sara Smoot and Sertan Girgin and Shariq Iqbal and Shashir Reddy and Shruti Sheth and Siim Põder and Sijal Bhatnagar and Sindhu Raghuram Panyam and Sivan Eiger and Susan Zhang and Tianqi Liu and Trevor Yacovone and Tyler Liechty and Uday Kalra and Utku Evci and Vedant Misra and Vincent Roseberry and Vlad Feinberg and Vlad Kolesnikov and Woohyun Han and Woosuk Kwon and Xi Chen and Yinlam Chow and Yuvein Zhu and Zichuan Wei and Zoltan Egyed and Victor Cotruta and Minh Giang and Phoebe Kirk and Anand Rao and Kat Black and Nabila Babar and Jessica Lo and Erica Moreira and Luiz Gustavo Martins and Omar Sanseviero and Lucas Gonzalez and Zach Gleicher and Tris Warkentin and Vahab Mirrokni and Evan Senter and Eli Collins and Joelle Barral and Zoubin Ghahramani and Raia Hadsell and Yossi Matias and D. Sculley and Slav Petrov and Noah Fiedel and Noam Shazeer and Oriol Vinyals and Jeff Dean and Demis Hassabis and Koray Kavukcuoglu and Clement Farabet and Elena Buchatskaya and Jean-Baptiste Alayrac and Rohan Anil and Dmitry and Lepikhin and Sebastian Borgeaud and Olivier Bachem and Armand Joulin and Alek Andreev and Cassidy Hardin and Robert Dadashi and Léonard Hussenot},
      year={2025},
      eprint={2503.19786},
      archivePrefix={arXiv},
      primaryClass={cs.CL},
      url={https://arxiv.org/abs/2503.19786}, 
}

@article{yesmin2025adr,
  author       = {Yesmin, Sabina},
  title        = {Alternative Dispute Resolution: A Strategic Solution to Bangladesh’s Judicial Backlog},
  year         = {2025},
  month        = may,
  url          = {https://www.tbsnews.net/thoughts/alternative-dispute-resolution-strategic-solution-bangladeshs-judicial-backlog-1134421},
  note         = {Accessed: 2025-06-23},
  publisher    = {The Business Standard}
}

@misc{geminiteam2025geminifamilyhighlycapable,
      title={Gemini: A Family of Highly Capable Multimodal Models}, 
      author={Gemini Team and Rohan Anil and Sebastian Borgeaud and Jean-Baptiste Alayrac and Jiahui Yu and Radu Soricut and Johan Schalkwyk and Andrew M. Dai and Anja Hauth and Katie Millican and David Silver and Melvin Johnson and Ioannis Antonoglou and Julian Schrittwieser and Amelia Glaese and Jilin Chen and Emily Pitler and Timothy Lillicrap and Angeliki Lazaridou and Orhan Firat and James Molloy and Michael Isard and Paul R. Barham and Tom Hennigan and Benjamin Lee and Fabio Viola and Malcolm Reynolds and Yuanzhong Xu and Ryan Doherty and Eli Collins and Clemens Meyer and Eliza Rutherford and Erica Moreira and Kareem Ayoub and Megha Goel and Jack Krawczyk and Cosmo Du and Ed Chi and Heng-Tze Cheng and Eric Ni and Purvi Shah and Patrick Kane and Betty Chan and Manaal Faruqui and Aliaksei Severyn and Hanzhao Lin and YaGuang Li and Yong Cheng and Abe Ittycheriah and Mahdis Mahdieh and Mia Chen and Pei Sun and Dustin Tran and Sumit Bagri and Balaji Lakshminarayanan and Jeremiah Liu and Andras Orban and Fabian Güra and Hao Zhou and Xinying Song and Aurelien Boffy and Harish Ganapathy and Steven Zheng and HyunJeong Choe and Ágoston Weisz and Tao Zhu and Yifeng Lu and Siddharth Gopal and Jarrod Kahn and Maciej Kula and Jeff Pitman and Rushin Shah and Emanuel Taropa and Majd Al Merey and Martin Baeuml and Zhifeng Chen and Laurent El Shafey and Yujing Zhang and Olcan Sercinoglu and George Tucker and Enrique Piqueras and Maxim Krikun and Iain Barr and Nikolay Savinov and Ivo Danihelka and Becca Roelofs and Anaïs White and Anders Andreassen and Tamara von Glehn and Lakshman Yagati and Mehran Kazemi and Lucas Gonzalez and Misha Khalman and Jakub Sygnowski and Alexandre Frechette and Charlotte Smith and Laura Culp and Lev Proleev and Yi Luan and Xi Chen and James Lottes and Nathan Schucher and Federico Lebron and Alban Rrustemi and Natalie Clay and Phil Crone and Tomas Kocisky and Jeffrey Zhao and Bartek Perz and Dian Yu and Heidi Howard and Adam Bloniarz and Jack W. Rae and Han Lu and Laurent Sifre and Marcello Maggioni and Fred Alcober and Dan Garrette and Megan Barnes and Shantanu Thakoor and Jacob Austin and Gabriel Barth-Maron and William Wong and Rishabh Joshi and Rahma Chaabouni and Deeni Fatiha and Arun Ahuja and Gaurav Singh Tomar and Evan Senter and Martin Chadwick and Ilya Kornakov and Nithya Attaluri and Iñaki Iturrate and Ruibo Liu and Yunxuan Li and Sarah Cogan and Jeremy Chen and Chao Jia and Chenjie Gu and Qiao Zhang and Jordan Grimstad and Ale Jakse Hartman and Xavier Garcia and Thanumalayan Sankaranarayana Pillai and Jacob Devlin and Michael Laskin and Diego de Las Casas and Dasha Valter and Connie Tao and Lorenzo Blanco and Adrià Puigdomènech Badia and David Reitter and Mianna Chen and Jenny Brennan and Clara Rivera and Sergey Brin and Shariq Iqbal and Gabriela Surita and Jane Labanowski and Abhi Rao and Stephanie Winkler and Emilio Parisotto and Yiming Gu and Kate Olszewska and Ravi Addanki and Antoine Miech and Annie Louis and Denis Teplyashin and Geoff Brown and Elliot Catt and Jan Balaguer and Jackie Xiang and Pidong Wang and Zoe Ashwood and Anton Briukhov and Albert Webson and Sanjay Ganapathy and Smit Sanghavi and Ajay Kannan and Ming-Wei Chang and Axel Stjerngren and Josip Djolonga and Yuting Sun and Ankur Bapna and Matthew Aitchison and Pedram Pejman and Henryk Michalewski and Tianhe Yu and Cindy Wang and Juliette Love and Junwhan Ahn and Dawn Bloxwich and Kehang Han and Peter Humphreys and Thibault Sellam and James Bradbury and Varun Godbole and Sina Samangooei and Bogdan Damoc and Alex Kaskasoli and Sébastien M. R. Arnold and Vijay Vasudevan and Shubham Agrawal and Jason Riesa and Dmitry Lepikhin and Richard Tanburn and Srivatsan Srinivasan and Hyeontaek Lim and Sarah Hodkinson and Pranav Shyam and Johan Ferret and Steven Hand and Ankush Garg and Tom Le Paine and Jian Li and Yujia Li and Minh Giang and Alexander Neitz and Zaheer Abbas and Sarah York and Machel Reid and Elizabeth Cole and Aakanksha Chowdhery and Dipanjan Das and Dominika Rogozińska and Vitaliy Nikolaev and Pablo Sprechmann and Zachary Nado and Lukas Zilka and Flavien Prost and Luheng He and Marianne Monteiro and Gaurav Mishra and Chris Welty and Josh Newlan and Dawei Jia and Miltiadis Allamanis and Clara Huiyi Hu and Raoul de Liedekerke and Justin Gilmer and Carl Saroufim and Shruti Rijhwani and Shaobo Hou and Disha Shrivastava and Anirudh Baddepudi and Alex Goldin and Adnan Ozturel and Albin Cassirer and Yunhan Xu and Daniel Sohn and Devendra Sachan and Reinald Kim Amplayo and Craig Swanson and Dessie Petrova and Shashi Narayan and Arthur Guez and Siddhartha Brahma and Jessica Landon and Miteyan Patel and Ruizhe Zhao and Kevin Villela and Luyu Wang and Wenhao Jia and Matthew Rahtz and Mai Giménez and Legg Yeung and James Keeling and Petko Georgiev and Diana Mincu and Boxi Wu and Salem Haykal and Rachel Saputro and Kiran Vodrahalli and James Qin and Zeynep Cankara and Abhanshu Sharma and Nick Fernando and Will Hawkins and Behnam Neyshabur and Solomon Kim and Adrian Hutter and Priyanka Agrawal and Alex Castro-Ros and George van den Driessche and Tao Wang and Fan Yang and Shuo-yiin Chang and Paul Komarek and Ross McIlroy and Mario Lučić and Guodong Zhang and Wael Farhan and Michael Sharman and Paul Natsev and Paul Michel and Yamini Bansal and Siyuan Qiao and Kris Cao and Siamak Shakeri and Christina Butterfield and Justin Chung and Paul Kishan Rubenstein and Shivani Agrawal and Arthur Mensch and Kedar Soparkar and Karel Lenc and Timothy Chung and Aedan Pope and Loren Maggiore and Jackie Kay and Priya Jhakra and Shibo Wang and Joshua Maynez and Mary Phuong and Taylor Tobin and Andrea Tacchetti and Maja Trebacz and Kevin Robinson and Yash Katariya and Sebastian Riedel and Paige Bailey and Kefan Xiao and Nimesh Ghelani and Lora Aroyo and Ambrose Slone and Neil Houlsby and Xuehan Xiong and Zhen Yang and Elena Gribovskaya and Jonas Adler and Mateo Wirth and Lisa Lee and Music Li and Thais Kagohara and Jay Pavagadhi and Sophie Bridgers and Anna Bortsova and Sanjay Ghemawat and Zafarali Ahmed and Tianqi Liu and Richard Powell and Vijay Bolina and Mariko Iinuma and Polina Zablotskaia and James Besley and Da-Woon Chung and Timothy Dozat and Ramona Comanescu and Xiance Si and Jeremy Greer and Guolong Su and Martin Polacek and Raphaël Lopez Kaufman and Simon Tokumine and Hexiang Hu and Elena Buchatskaya and Yingjie Miao and Mohamed Elhawaty and Aditya Siddhant and Nenad Tomasev and Jinwei Xing and Christina Greer and Helen Miller and Shereen Ashraf and Aurko Roy and Zizhao Zhang and Ada Ma and Angelos Filos and Milos Besta and Rory Blevins and Ted Klimenko and Chih-Kuan Yeh and Soravit Changpinyo and Jiaqi Mu and Oscar Chang and Mantas Pajarskas and Carrie Muir and Vered Cohen and Charline Le Lan and Krishna Haridasan and Amit Marathe and Steven Hansen and Sholto Douglas and Rajkumar Samuel and Mingqiu Wang and Sophia Austin and Chang Lan and Jiepu Jiang and Justin Chiu and Jaime Alonso Lorenzo and Lars Lowe Sjösund and Sébastien Cevey and Zach Gleicher and Thi Avrahami and Anudhyan Boral and Hansa Srinivasan and Vittorio Selo and Rhys May and Konstantinos Aisopos and Léonard Hussenot and Livio Baldini Soares and Kate Baumli and Michael B. Chang and Adrià Recasens and Ben Caine and Alexander Pritzel and Filip Pavetic and Fabio Pardo and Anita Gergely and Justin Frye and Vinay Ramasesh and Dan Horgan and Kartikeya Badola and Nora Kassner and Subhrajit Roy and Ethan Dyer and Víctor Campos Campos and Alex Tomala and Yunhao Tang and Dalia El Badawy and Elspeth White and Basil Mustafa and Oran Lang and Abhishek Jindal and Sharad Vikram and Zhitao Gong and Sergi Caelles and Ross Hemsley and Gregory Thornton and Fangxiaoyu Feng and Wojciech Stokowiec and Ce Zheng and Phoebe Thacker and Çağlar Ünlü and Zhishuai Zhang and Mohammad Saleh and James Svensson and Max Bileschi and Piyush Patil and Ankesh Anand and Roman Ring and Katerina Tsihlas and Arpi Vezer and Marco Selvi and Toby Shevlane and Mikel Rodriguez and Tom Kwiatkowski and Samira Daruki and Keran Rong and Allan Dafoe and Nicholas FitzGerald and Keren Gu-Lemberg and Mina Khan and Lisa Anne Hendricks and Marie Pellat and Vladimir Feinberg and James Cobon-Kerr and Tara Sainath and Maribeth Rauh and Sayed Hadi Hashemi and Richard Ives and Yana Hasson and Eric Noland and Yuan Cao and Nathan Byrd and Le Hou and Qingze Wang and Thibault Sottiaux and Michela Paganini and Jean-Baptiste Lespiau and Alexandre Moufarek and Samer Hassan and Kaushik Shivakumar and Joost van Amersfoort and Amol Mandhane and Pratik Joshi and Anirudh Goyal and Matthew Tung and Andrew Brock and Hannah Sheahan and Vedant Misra and Cheng Li and Nemanja Rakićević and Mostafa Dehghani and Fangyu Liu and Sid Mittal and Junhyuk Oh and Seb Noury and Eren Sezener and Fantine Huot and Matthew Lamm and Nicola De Cao and Charlie Chen and Sidharth Mudgal and Romina Stella and Kevin Brooks and Gautam Vasudevan and Chenxi Liu and Mainak Chain and Nivedita Melinkeri and Aaron Cohen and Venus Wang and Kristie Seymore and Sergey Zubkov and Rahul Goel and Summer Yue and Sai Krishnakumaran and Brian Albert and Nate Hurley and Motoki Sano and Anhad Mohananey and Jonah Joughin and Egor Filonov and Tomasz Kępa and Yomna Eldawy and Jiawern Lim and Rahul Rishi and Shirin Badiezadegan and Taylor Bos and Jerry Chang and Sanil Jain and Sri Gayatri Sundara Padmanabhan and Subha Puttagunta and Kalpesh Krishna and Leslie Baker and Norbert Kalb and Vamsi Bedapudi and Adam Kurzrok and Shuntong Lei and Anthony Yu and Oren Litvin and Xiang Zhou and Zhichun Wu and Sam Sobell and Andrea Siciliano and Alan Papir and Robby Neale and Jonas Bragagnolo and Tej Toor and Tina Chen and Valentin Anklin and Feiran Wang and Richie Feng and Milad Gholami and Kevin Ling and Lijuan Liu and Jules Walter and Hamid Moghaddam and Arun Kishore and Jakub Adamek and Tyler Mercado and Jonathan Mallinson and Siddhinita Wandekar and Stephen Cagle and Eran Ofek and Guillermo Garrido and Clemens Lombriser and Maksim Mukha and Botu Sun and Hafeezul Rahman Mohammad and Josip Matak and Yadi Qian and Vikas Peswani and Pawel Janus and Quan Yuan and Leif Schelin and Oana David and Ankur Garg and Yifan He and Oleksii Duzhyi and Anton Älgmyr and Timothée Lottaz and Qi Li and Vikas Yadav and Luyao Xu and Alex Chinien and Rakesh Shivanna and Aleksandr Chuklin and Josie Li and Carrie Spadine and Travis Wolfe and Kareem Mohamed and Subhabrata Das and Zihang Dai and Kyle He and Daniel von Dincklage and Shyam Upadhyay and Akanksha Maurya and Luyan Chi and Sebastian Krause and Khalid Salama and Pam G Rabinovitch and Pavan Kumar Reddy M and Aarush Selvan and Mikhail Dektiarev and Golnaz Ghiasi and Erdem Guven and Himanshu Gupta and Boyi Liu and Deepak Sharma and Idan Heimlich Shtacher and Shachi Paul and Oscar Akerlund and François-Xavier Aubet and Terry Huang and Chen Zhu and Eric Zhu and Elico Teixeira and Matthew Fritze and Francesco Bertolini and Liana-Eleonora Marinescu and Martin Bölle and Dominik Paulus and Khyatti Gupta and Tejasi Latkar and Max Chang and Jason Sanders and Roopa Wilson and Xuewei Wu and Yi-Xuan Tan and Lam Nguyen Thiet and Tulsee Doshi and Sid Lall and Swaroop Mishra and Wanming Chen and Thang Luong and Seth Benjamin and Jasmine Lee and Ewa Andrejczuk and Dominik Rabiej and Vipul Ranjan and Krzysztof Styrc and Pengcheng Yin and Jon Simon and Malcolm Rose Harriott and Mudit Bansal and Alexei Robsky and Geoff Bacon and David Greene and Daniil Mirylenka and Chen Zhou and Obaid Sarvana and Abhimanyu Goyal and Samuel Andermatt and Patrick Siegler and Ben Horn and Assaf Israel and Francesco Pongetti and Chih-Wei "Louis" Chen and Marco Selvatici and Pedro Silva and Kathie Wang and Jackson Tolins and Kelvin Guu and Roey Yogev and Xiaochen Cai and Alessandro Agostini and Maulik Shah and Hung Nguyen and Noah Ó Donnaile and Sébastien Pereira and Linda Friso and Adam Stambler and Adam Kurzrok and Chenkai Kuang and Yan Romanikhin and Mark Geller and ZJ Yan and Kane Jang and Cheng-Chun Lee and Wojciech Fica and Eric Malmi and Qijun Tan and Dan Banica and Daniel Balle and Ryan Pham and Yanping Huang and Diana Avram and Hongzhi Shi and Jasjot Singh and Chris Hidey and Niharika Ahuja and Pranab Saxena and Dan Dooley and Srividya Pranavi Potharaju and Eileen O'Neill and Anand Gokulchandran and Ryan Foley and Kai Zhao and Mike Dusenberry and Yuan Liu and Pulkit Mehta and Ragha Kotikalapudi and Chalence Safranek-Shrader and Andrew Goodman and Joshua Kessinger and Eran Globen and Prateek Kolhar and Chris Gorgolewski and Ali Ibrahim and Yang Song and Ali Eichenbaum and Thomas Brovelli and Sahitya Potluri and Preethi Lahoti and Cip Baetu and Ali Ghorbani and Charles Chen and Andy Crawford and Shalini Pal and Mukund Sridhar and Petru Gurita and Asier Mujika and Igor Petrovski and Pierre-Louis Cedoz and Chenmei Li and Shiyuan Chen and Niccolò Dal Santo and Siddharth Goyal and Jitesh Punjabi and Karthik Kappaganthu and Chester Kwak and Pallavi LV and Sarmishta Velury and Himadri Choudhury and Jamie Hall and Premal Shah and Ricardo Figueira and Matt Thomas and Minjie Lu and Ting Zhou and Chintu Kumar and Thomas Jurdi and Sharat Chikkerur and Yenai Ma and Adams Yu and Soo Kwak and Victor Ähdel and Sujeevan Rajayogam and Travis Choma and Fei Liu and Aditya Barua and Colin Ji and Ji Ho Park and Vincent Hellendoorn and Alex Bailey and Taylan Bilal and Huanjie Zhou and Mehrdad Khatir and Charles Sutton and Wojciech Rzadkowski and Fiona Macintosh and Roopali Vij and Konstantin Shagin and Paul Medina and Chen Liang and Jinjing Zhou and Pararth Shah and Yingying Bi and Attila Dankovics and Shipra Banga and Sabine Lehmann and Marissa Bredesen and Zifan Lin and John Eric Hoffmann and Jonathan Lai and Raynald Chung and Kai Yang and Nihal Balani and Arthur Bražinskas and Andrei Sozanschi and Matthew Hayes and Héctor Fernández Alcalde and Peter Makarov and Will Chen and Antonio Stella and Liselotte Snijders and Michael Mandl and Ante Kärrman and Paweł Nowak and Xinyi Wu and Alex Dyck and Krishnan Vaidyanathan and Raghavender R and Jessica Mallet and Mitch Rudominer and Eric Johnston and Sushil Mittal and Akhil Udathu and Janara Christensen and Vishal Verma and Zach Irving and Andreas Santucci and Gamaleldin Elsayed and Elnaz Davoodi and Marin Georgiev and Ian Tenney and Nan Hua and Geoffrey Cideron and Edouard Leurent and Mahmoud Alnahlawi and Ionut Georgescu and Nan Wei and Ivy Zheng and Dylan Scandinaro and Heinrich Jiang and Jasper Snoek and Mukund Sundararajan and Xuezhi Wang and Zack Ontiveros and Itay Karo and Jeremy Cole and Vinu Rajashekhar and Lara Tumeh and Eyal Ben-David and Rishub Jain and Jonathan Uesato and Romina Datta and Oskar Bunyan and Shimu Wu and John Zhang and Piotr Stanczyk and Ye Zhang and David Steiner and Subhajit Naskar and Michael Azzam and Matthew Johnson and Adam Paszke and Chung-Cheng Chiu and Jaume Sanchez Elias and Afroz Mohiuddin and Faizan Muhammad and Jin Miao and Andrew Lee and Nino Vieillard and Jane Park and Jiageng Zhang and Jeff Stanway and Drew Garmon and Abhijit Karmarkar and Zhe Dong and Jong Lee and Aviral Kumar and Luowei Zhou and Jonathan Evens and William Isaac and Geoffrey Irving and Edward Loper and Michael Fink and Isha Arkatkar and Nanxin Chen and Izhak Shafran and Ivan Petrychenko and Zhe Chen and Johnson Jia and Anselm Levskaya and Zhenkai Zhu and Peter Grabowski and Yu Mao and Alberto Magni and Kaisheng Yao and Javier Snaider and Norman Casagrande and Evan Palmer and Paul Suganthan and Alfonso Castaño and Irene Giannoumis and Wooyeol Kim and Mikołaj Rybiński and Ashwin Sreevatsa and Jennifer Prendki and David Soergel and Adrian Goedeckemeyer and Willi Gierke and Mohsen Jafari and Meenu Gaba and Jeremy Wiesner and Diana Gage Wright and Yawen Wei and Harsha Vashisht and Yana Kulizhskaya and Jay Hoover and Maigo Le and Lu Li and Chimezie Iwuanyanwu and Lu Liu and Kevin Ramirez and Andrey Khorlin and Albert Cui and Tian LIN and Marcus Wu and Ricardo Aguilar and Keith Pallo and Abhishek Chakladar and Ginger Perng and Elena Allica Abellan and Mingyang Zhang and Ishita Dasgupta and Nate Kushman and Ivo Penchev and Alena Repina and Xihui Wu and Tom van der Weide and Priya Ponnapalli and Caroline Kaplan and Jiri Simsa and Shuangfeng Li and Olivier Dousse and Fan Yang and Jeff Piper and Nathan Ie and Rama Pasumarthi and Nathan Lintz and Anitha Vijayakumar and Daniel Andor and Pedro Valenzuela and Minnie Lui and Cosmin Paduraru and Daiyi Peng and Katherine Lee and Shuyuan Zhang and Somer Greene and Duc Dung Nguyen and Paula Kurylowicz and Cassidy Hardin and Lucas Dixon and Lili Janzer and Kiam Choo and Ziqiang Feng and Biao Zhang and Achintya Singhal and Dayou Du and Dan McKinnon and Natasha Antropova and Tolga Bolukbasi and Orgad Keller and David Reid and Daniel Finchelstein and Maria Abi Raad and Remi Crocker and Peter Hawkins and Robert Dadashi and Colin Gaffney and Ken Franko and Anna Bulanova and Rémi Leblond and Shirley Chung and Harry Askham and Luis C. Cobo and Kelvin Xu and Felix Fischer and Jun Xu and Christina Sorokin and Chris Alberti and Chu-Cheng Lin and Colin Evans and Alek Dimitriev and Hannah Forbes and Dylan Banarse and Zora Tung and Mark Omernick and Colton Bishop and Rachel Sterneck and Rohan Jain and Jiawei Xia and Ehsan Amid and Francesco Piccinno and Xingyu Wang and Praseem Banzal and Daniel J. Mankowitz and Alex Polozov and Victoria Krakovna and Sasha Brown and MohammadHossein Bateni and Dennis Duan and Vlad Firoiu and Meghana Thotakuri and Tom Natan and Matthieu Geist and Ser tan Girgin and Hui Li and Jiayu Ye and Ofir Roval and Reiko Tojo and Michael Kwong and James Lee-Thorp and Christopher Yew and Danila Sinopalnikov and Sabela Ramos and John Mellor and Abhishek Sharma and Kathy Wu and David Miller and Nicolas Sonnerat and Denis Vnukov and Rory Greig and Jennifer Beattie and Emily Caveness and Libin Bai and Julian Eisenschlos and Alex Korchemniy and Tomy Tsai and Mimi Jasarevic and Weize Kong and Phuong Dao and Zeyu Zheng and Frederick Liu and Fan Yang and Rui Zhu and Tian Huey Teh and Jason Sanmiya and Evgeny Gladchenko and Nejc Trdin and Daniel Toyama and Evan Rosen and Sasan Tavakkol and Linting Xue and Chen Elkind and Oliver Woodman and John Carpenter and George Papamakarios and Rupert Kemp and Sushant Kafle and Tanya Grunina and Rishika Sinha and Alice Talbert and Diane Wu and Denese Owusu-Afriyie and Cosmo Du and Chloe Thornton and Jordi Pont-Tuset and Pradyumna Narayana and Jing Li and Saaber Fatehi and John Wieting and Omar Ajmeri and Benigno Uria and Yeongil Ko and Laura Knight and Amélie Héliou and Ning Niu and Shane Gu and Chenxi Pang and Yeqing Li and Nir Levine and Ariel Stolovich and Rebeca Santamaria-Fernandez and Sonam Goenka and Wenny Yustalim and Robin Strudel and Ali Elqursh and Charlie Deck and Hyo Lee and Zonglin Li and Kyle Levin and Raphael Hoffmann and Dan Holtmann-Rice and Olivier Bachem and Sho Arora and Christy Koh and Soheil Hassas Yeganeh and Siim Põder and Mukarram Tariq and Yanhua Sun and Lucian Ionita and Mojtaba Seyedhosseini and Pouya Tafti and Zhiyu Liu and Anmol Gulati and Jasmine Liu and Xinyu Ye and Bart Chrzaszcz and Lily Wang and Nikhil Sethi and Tianrun Li and Ben Brown and Shreya Singh and Wei Fan and Aaron Parisi and Joe Stanton and Vinod Koverkathu and Christopher A. Choquette-Choo and Yunjie Li and TJ Lu and Abe Ittycheriah and Prakash Shroff and Mani Varadarajan and Sanaz Bahargam and Rob Willoughby and David Gaddy and Guillaume Desjardins and Marco Cornero and Brona Robenek and Bhavishya Mittal and Ben Albrecht and Ashish Shenoy and Fedor Moiseev and Henrik Jacobsson and Alireza Ghaffarkhah and Morgane Rivière and Alanna Walton and Clément Crepy and Alicia Parrish and Zongwei Zhou and Clement Farabet and Carey Radebaugh and Praveen Srinivasan and Claudia van der Salm and Andreas Fidjeland and Salvatore Scellato and Eri Latorre-Chimoto and Hanna Klimczak-Plucińska and David Bridson and Dario de Cesare and Tom Hudson and Piermaria Mendolicchio and Lexi Walker and Alex Morris and Matthew Mauger and Alexey Guseynov and Alison Reid and Seth Odoom and Lucia Loher and Victor Cotruta and Madhavi Yenugula and Dominik Grewe and Anastasia Petrushkina and Tom Duerig and Antonio Sanchez and Steve Yadlowsky and Amy Shen and Amir Globerson and Lynette Webb and Sahil Dua and Dong Li and Surya Bhupatiraju and Dan Hurt and Haroon Qureshi and Ananth Agarwal and Tomer Shani and Matan Eyal and Anuj Khare and Shreyas Rammohan Belle and Lei Wang and Chetan Tekur and Mihir Sanjay Kale and Jinliang Wei and Ruoxin Sang and Brennan Saeta and Tyler Liechty and Yi Sun and Yao Zhao and Stephan Lee and Pandu Nayak and Doug Fritz and Manish Reddy Vuyyuru and John Aslanides and Nidhi Vyas and Martin Wicke and Xiao Ma and Evgenii Eltyshev and Nina Martin and Hardie Cate and James Manyika and Keyvan Amiri and Yelin Kim and Xi Xiong and Kai Kang and Florian Luisier and Nilesh Tripuraneni and David Madras and Mandy Guo and Austin Waters and Oliver Wang and Joshua Ainslie and Jason Baldridge and Han Zhang and Garima Pruthi and Jakob Bauer and Feng Yang and Riham Mansour and Jason Gelman and Yang Xu and George Polovets and Ji Liu and Honglong Cai and Warren Chen and XiangHai Sheng and Emily Xue and Sherjil Ozair and Christof Angermueller and Xiaowei Li and Anoop Sinha and Weiren Wang and Julia Wiesinger and Emmanouil Koukoumidis and Yuan Tian and Anand Iyer and Madhu Gurumurthy and Mark Goldenson and Parashar Shah and MK Blake and Hongkun Yu and Anthony Urbanowicz and Jennimaria Palomaki and Chrisantha Fernando and Ken Durden and Harsh Mehta and Nikola Momchev and Elahe Rahimtoroghi and Maria Georgaki and Amit Raul and Sebastian Ruder and Morgan Redshaw and Jinhyuk Lee and Denny Zhou and Komal Jalan and Dinghua Li and Blake Hechtman and Parker Schuh and Milad Nasr and Kieran Milan and Vladimir Mikulik and Juliana Franco and Tim Green and Nam Nguyen and Joe Kelley and Aroma Mahendru and Andrea Hu and Joshua Howland and Ben Vargas and Jeffrey Hui and Kshitij Bansal and Vikram Rao and Rakesh Ghiya and Emma Wang and Ke Ye and Jean Michel Sarr and Melanie Moranski Preston and Madeleine Elish and Steve Li and Aakash Kaku and Jigar Gupta and Ice Pasupat and Da-Cheng Juan and Milan Someswar and Tejvi M. and Xinyun Chen and Aida Amini and Alex Fabrikant and Eric Chu and Xuanyi Dong and Amruta Muthal and Senaka Buthpitiya and Sarthak Jauhari and Nan Hua and Urvashi Khandelwal and Ayal Hitron and Jie Ren and Larissa Rinaldi and Shahar Drath and Avigail Dabush and Nan-Jiang Jiang and Harshal Godhia and Uli Sachs and Anthony Chen and Yicheng Fan and Hagai Taitelbaum and Hila Noga and Zhuyun Dai and James Wang and Chen Liang and Jenny Hamer and Chun-Sung Ferng and Chenel Elkind and Aviel Atias and Paulina Lee and Vít Listík and Mathias Carlen and Jan van de Kerkhof and Marcin Pikus and Krunoslav Zaher and Paul Müller and Sasha Zykova and Richard Stefanec and Vitaly Gatsko and Christoph Hirnschall and Ashwin Sethi and Xingyu Federico Xu and Chetan Ahuja and Beth Tsai and Anca Stefanoiu and Bo Feng and Keshav Dhandhania and Manish Katyal and Akshay Gupta and Atharva Parulekar and Divya Pitta and Jing Zhao and Vivaan Bhatia and Yashodha Bhavnani and Omar Alhadlaq and Xiaolin Li and Peter Danenberg and Dennis Tu and Alex Pine and Vera Filippova and Abhipso Ghosh and Ben Limonchik and Bhargava Urala and Chaitanya Krishna Lanka and Derik Clive and Yi Sun and Edward Li and Hao Wu and Kevin Hongtongsak and Ianna Li and Kalind Thakkar and Kuanysh Omarov and Kushal Majmundar and Michael Alverson and Michael Kucharski and Mohak Patel and Mudit Jain and Maksim Zabelin and Paolo Pelagatti and Rohan Kohli and Saurabh Kumar and Joseph Kim and Swetha Sankar and Vineet Shah and Lakshmi Ramachandruni and Xiangkai Zeng and Ben Bariach and Laura Weidinger and Tu Vu and Alek Andreev and Antoine He and Kevin Hui and Sheleem Kashem and Amar Subramanya and Sissie Hsiao and Demis Hassabis and Koray Kavukcuoglu and Adam Sadovsky and Quoc Le and Trevor Strohman and Yonghui Wu and Slav Petrov and Jeffrey Dean and Oriol Vinyals},
      year={2025},
      eprint={2312.11805},
      archivePrefix={arXiv},
      primaryClass={cs.CL},
      url={https://arxiv.org/abs/2312.11805}, 
}

@misc{openai2024gpt4ocard,
      title={GPT-4o System Card}, 
      author={OpenAI and : and Aaron Hurst and Adam Lerer and Adam P. Goucher and Adam Perelman and Aditya Ramesh and Aidan Clark and AJ Ostrow and Akila Welihinda and Alan Hayes and Alec Radford and Aleksander Mądry and Alex Baker-Whitcomb and Alex Beutel and Alex Borzunov and Alex Carney and Alex Chow and Alex Kirillov and Alex Nichol and Alex Paino and Alex Renzin and Alex Tachard Passos and Alexander Kirillov and Alexi Christakis and Alexis Conneau and Ali Kamali and Allan Jabri and Allison Moyer and Allison Tam and Amadou Crookes and Amin Tootoochian and Amin Tootoonchian and Ananya Kumar and Andrea Vallone and Andrej Karpathy and Andrew Braunstein and Andrew Cann and Andrew Codispoti and Andrew Galu and Andrew Kondrich and Andrew Tulloch and Andrey Mishchenko and Angela Baek and Angela Jiang and Antoine Pelisse and Antonia Woodford and Anuj Gosalia and Arka Dhar and Ashley Pantuliano and Avi Nayak and Avital Oliver and Barret Zoph and Behrooz Ghorbani and Ben Leimberger and Ben Rossen and Ben Sokolowsky and Ben Wang and Benjamin Zweig and Beth Hoover and Blake Samic and Bob McGrew and Bobby Spero and Bogo Giertler and Bowen Cheng and Brad Lightcap and Brandon Walkin and Brendan Quinn and Brian Guarraci and Brian Hsu and Bright Kellogg and Brydon Eastman and Camillo Lugaresi and Carroll Wainwright and Cary Bassin and Cary Hudson and Casey Chu and Chad Nelson and Chak Li and Chan Jun Shern and Channing Conger and Charlotte Barette and Chelsea Voss and Chen Ding and Cheng Lu and Chong Zhang and Chris Beaumont and Chris Hallacy and Chris Koch and Christian Gibson and Christina Kim and Christine Choi and Christine McLeavey and Christopher Hesse and Claudia Fischer and Clemens Winter and Coley Czarnecki and Colin Jarvis and Colin Wei and Constantin Koumouzelis and Dane Sherburn and Daniel Kappler and Daniel Levin and Daniel Levy and David Carr and David Farhi and David Mely and David Robinson and David Sasaki and Denny Jin and Dev Valladares and Dimitris Tsipras and Doug Li and Duc Phong Nguyen and Duncan Findlay and Edede Oiwoh and Edmund Wong and Ehsan Asdar and Elizabeth Proehl and Elizabeth Yang and Eric Antonow and Eric Kramer and Eric Peterson and Eric Sigler and Eric Wallace and Eugene Brevdo and Evan Mays and Farzad Khorasani and Felipe Petroski Such and Filippo Raso and Francis Zhang and Fred von Lohmann and Freddie Sulit and Gabriel Goh and Gene Oden and Geoff Salmon and Giulio Starace and Greg Brockman and Hadi Salman and Haiming Bao and Haitang Hu and Hannah Wong and Haoyu Wang and Heather Schmidt and Heather Whitney and Heewoo Jun and Hendrik Kirchner and Henrique Ponde de Oliveira Pinto and Hongyu Ren and Huiwen Chang and Hyung Won Chung and Ian Kivlichan and Ian O'Connell and Ian O'Connell and Ian Osband and Ian Silber and Ian Sohl and Ibrahim Okuyucu and Ikai Lan and Ilya Kostrikov and Ilya Sutskever and Ingmar Kanitscheider and Ishaan Gulrajani and Jacob Coxon and Jacob Menick and Jakub Pachocki and James Aung and James Betker and James Crooks and James Lennon and Jamie Kiros and Jan Leike and Jane Park and Jason Kwon and Jason Phang and Jason Teplitz and Jason Wei and Jason Wolfe and Jay Chen and Jeff Harris and Jenia Varavva and Jessica Gan Lee and Jessica Shieh and Ji Lin and Jiahui Yu and Jiayi Weng and Jie Tang and Jieqi Yu and Joanne Jang and Joaquin Quinonero Candela and Joe Beutler and Joe Landers and Joel Parish and Johannes Heidecke and John Schulman and Jonathan Lachman and Jonathan McKay and Jonathan Uesato and Jonathan Ward and Jong Wook Kim and Joost Huizinga and Jordan Sitkin and Jos Kraaijeveld and Josh Gross and Josh Kaplan and Josh Snyder and Joshua Achiam and Joy Jiao and Joyce Lee and Juntang Zhuang and Justyn Harriman and Kai Fricke and Kai Hayashi and Karan Singhal and Katy Shi and Kavin Karthik and Kayla Wood and Kendra Rimbach and Kenny Hsu and Kenny Nguyen and Keren Gu-Lemberg and Kevin Button and Kevin Liu and Kiel Howe and Krithika Muthukumar and Kyle Luther and Lama Ahmad and Larry Kai and Lauren Itow and Lauren Workman and Leher Pathak and Leo Chen and Li Jing and Lia Guy and Liam Fedus and Liang Zhou and Lien Mamitsuka and Lilian Weng and Lindsay McCallum and Lindsey Held and Long Ouyang and Louis Feuvrier and Lu Zhang and Lukas Kondraciuk and Lukasz Kaiser and Luke Hewitt and Luke Metz and Lyric Doshi and Mada Aflak and Maddie Simens and Madelaine Boyd and Madeleine Thompson and Marat Dukhan and Mark Chen and Mark Gray and Mark Hudnall and Marvin Zhang and Marwan Aljubeh and Mateusz Litwin and Matthew Zeng and Max Johnson and Maya Shetty and Mayank Gupta and Meghan Shah and Mehmet Yatbaz and Meng Jia Yang and Mengchao Zhong and Mia Glaese and Mianna Chen and Michael Janner and Michael Lampe and Michael Petrov and Michael Wu and Michele Wang and Michelle Fradin and Michelle Pokrass and Miguel Castro and Miguel Oom Temudo de Castro and Mikhail Pavlov and Miles Brundage and Miles Wang and Minal Khan and Mira Murati and Mo Bavarian and Molly Lin and Murat Yesildal and Nacho Soto and Natalia Gimelshein and Natalie Cone and Natalie Staudacher and Natalie Summers and Natan LaFontaine and Neil Chowdhury and Nick Ryder and Nick Stathas and Nick Turley and Nik Tezak and Niko Felix and Nithanth Kudige and Nitish Keskar and Noah Deutsch and Noel Bundick and Nora Puckett and Ofir Nachum and Ola Okelola and Oleg Boiko and Oleg Murk and Oliver Jaffe and Olivia Watkins and Olivier Godement and Owen Campbell-Moore and Patrick Chao and Paul McMillan and Pavel Belov and Peng Su and Peter Bak and Peter Bakkum and Peter Deng and Peter Dolan and Peter Hoeschele and Peter Welinder and Phil Tillet and Philip Pronin and Philippe Tillet and Prafulla Dhariwal and Qiming Yuan and Rachel Dias and Rachel Lim and Rahul Arora and Rajan Troll and Randall Lin and Rapha Gontijo Lopes and Raul Puri and Reah Miyara and Reimar Leike and Renaud Gaubert and Reza Zamani and Ricky Wang and Rob Donnelly and Rob Honsby and Rocky Smith and Rohan Sahai and Rohit Ramchandani and Romain Huet and Rory Carmichael and Rowan Zellers and Roy Chen and Ruby Chen and Ruslan Nigmatullin and Ryan Cheu and Saachi Jain and Sam Altman and Sam Schoenholz and Sam Toizer and Samuel Miserendino and Sandhini Agarwal and Sara Culver and Scott Ethersmith and Scott Gray and Sean Grove and Sean Metzger and Shamez Hermani and Shantanu Jain and Shengjia Zhao and Sherwin Wu and Shino Jomoto and Shirong Wu and Shuaiqi and Xia and Sonia Phene and Spencer Papay and Srinivas Narayanan and Steve Coffey and Steve Lee and Stewart Hall and Suchir Balaji and Tal Broda and Tal Stramer and Tao Xu and Tarun Gogineni and Taya Christianson and Ted Sanders and Tejal Patwardhan and Thomas Cunninghman and Thomas Degry and Thomas Dimson and Thomas Raoux and Thomas Shadwell and Tianhao Zheng and Todd Underwood and Todor Markov and Toki Sherbakov and Tom Rubin and Tom Stasi and Tomer Kaftan and Tristan Heywood and Troy Peterson and Tyce Walters and Tyna Eloundou and Valerie Qi and Veit Moeller and Vinnie Monaco and Vishal Kuo and Vlad Fomenko and Wayne Chang and Weiyi Zheng and Wenda Zhou and Wesam Manassra and Will Sheu and Wojciech Zaremba and Yash Patil and Yilei Qian and Yongjik Kim and Youlong Cheng and Yu Zhang and Yuchen He and Yuchen Zhang and Yujia Jin and Yunxing Dai and Yury Malkov},
      year={2024},
      eprint={2410.21276},
      archivePrefix={arXiv},
      primaryClass={cs.CL},
      url={https://arxiv.org/abs/2410.21276}, 
}

@article{jaan2023justice,
  author       = {Jaan, Arafat Reza},
  title        = {What Makes Access to Justice Elusive in Bangladesh?},
  year         = {2023},
  month        = may,
  url          = {https://blogs.lse.ac.uk/southasia/2023/05/01/what-makes-access-to-justice-elusive-in-bangladesh/},
  note         = {Accessed: 2025-06-22},
  publisher    = {LSE South Asia Centre}
}

@article{khalilullah2025justice,
  author       = {Khalilullah, Md Ibrahim},
  title        = {Justice on Hold: Addressing Bangladesh's Judicial Backlog},
  year         = {2025},
  month        = jan,
  url          = {https://www.tbsnews.net/thoughts/justice-hold-addressing-bangladeshs-judicial-backlog-1039686},
  note         = {Accessed: 2025-06-23},
  publisher    = {The Business Standard}
}

@inbook{Frankenreiter_Nyarko_2023, place={Cambridge}, title={Natural Language Processing in Legal Tech}, booktitle={Legal Tech and the Future of Civil Justice}, publisher={Cambridge University Press}, author={Frankenreiter, Jens and Nyarko, Julian}, editor={Engstrom, David FreemanEditor}, year={2023}, pages={70–90}}

@inproceedings{zhong-etal-2020-nlp,
    title = "How Does {NLP} Benefit Legal System: A Summary of Legal Artificial Intelligence",
    author = "Zhong, Haoxi  and
      Xiao, Chaojun  and
      Tu, Cunchao  and
      Zhang, Tianyang  and
      Liu, Zhiyuan  and
      Sun, Maosong",
    editor = "Jurafsky, Dan  and
      Chai, Joyce  and
      Schluter, Natalie  and
      Tetreault, Joel",
    booktitle = "Proceedings of the 58th Annual Meeting of the Association for Computational Linguistics",
    month = jul,
    year = "2020",
    address = "Online",
    publisher = "Association for Computational Linguistics",
    url = "https://aclanthology.org/2020.acl-main.466",
    doi = "10.18653/v1/2020.acl-main.466",
    pages = "5218--5230",
}

@article{cegagw43gar3,
    author = "Arif Ahmed",
    title = "Challenges with Access to Justice in Bangladesh",
    year = "2021",
journal = "Centre for Governance Studies"
}

@misc{sun2024lawluochineselawfirm,
      title={LawLuo: A Chinese Law Firm Co-run by LLM Agents}, 
      author={Jingyun Sun and Chengxiao Dai and Zhongze Luo and Yangbo Chang and Yang Li},
      year={2024},
      eprint={2407.16252},
      archivePrefix={arXiv},
      primaryClass={cs.CL},
      url={https://arxiv.org/abs/2407.16252}, 
}

@misc{martin2024bettergptcomparinglarge,
      title={Better Call GPT, Comparing Large Language Models Against Lawyers}, 
      author={Lauren Martin and Nick Whitehouse and Stephanie Yiu and Lizzie Catterson and Rivindu Perera},
      year={2024},
      eprint={2401.16212},
      archivePrefix={arXiv},
      primaryClass={cs.CY},
      url={https://arxiv.org/abs/2401.16212}, 
}

@misc{qin2024exploringnexuslargelanguage,
      title={Exploring the Nexus of Large Language Models and Legal Systems: A Short Survey}, 
      author={Weicong Qin and Zhongxiang Sun},
      year={2024},
      eprint={2404.00990},
      archivePrefix={arXiv},
      primaryClass={cs.CL},
      url={https://arxiv.org/abs/2404.00990}, 
}

@article{Dahl20segfeg24,
  title = {Large Legal Fictions: Profiling Legal Hallucinations in Large Language Models},
  volume = {16},
  ISSN = {1946-5319},
  url = {http://dx.doi.org/10.1093/jla/laae003},
  DOI = {10.1093/jla/laae003},
  number = {1},
  journal = {Journal of Legal Analysis},
  publisher = {Oxford University Press (OUP)},
  author = {Dahl,  Matthew and Magesh,  Varun and Suzgun,  Mirac and Ho,  Daniel E},
  year = {2024},
  month = jan,
  pages = {64–93}
}

@misc{huang2024optimizingnumericalestimationoperational,
      title={Optimizing Numerical Estimation and Operational Efficiency in the Legal Domain through Large Language Models}, 
      author={Jia-Hong Huang and Chao-Chun Yang and Yixian Shen and Alessio M. Pacces and Evangelos Kanoulas},
      year={2024},
      eprint={2407.19041},
      archivePrefix={arXiv},
      primaryClass={cs.AI},
      url={https://arxiv.org/abs/2407.19041}, 
}

@misc{hussain2024largelanguagemodelsjudicial,
      title={Large Language Models for Judicial Entity Extraction: A Comparative Study}, 
      author={Atin Sakkeer Hussain and Anu Thomas},
      year={2024},
      eprint={2407.05786},
      archivePrefix={arXiv},
      primaryClass={cs.CL},
      url={https://arxiv.org/abs/2407.05786}, 
}

@misc{mamalis_kalampokis_fitsilis_theodorakopoulos_tarabanis_2024,
 title={A Large Language Model based legal assistant for governance applications},
 url={osf.io/94fmq},
 DOI={10.31219/osf.io/94fmq},
 publisher={OSF Preprints},
 author={Mamalis, Marios and Kalampokis, Evangelos and Fitsilis, Fotios and Theodorakopoulos, Georgios and Tarabanis, Konstantinos},
 year={2024},
 month={May}
}

@inproceedings{jiang-etal-2024-leveraging,
    title = "Leveraging Large Language Models for Learning Complex Legal Concepts through Storytelling",
    author = "Jiang, Hang  and
      Zhang, Xiajie  and
      Mahari, Robert  and
      Kessler, Daniel  and
      Ma, Eric  and
      August, Tal  and
      Li, Irene  and
      Pentland, Alex  and
      Kim, Yoon  and
      Roy, Deb  and
      Kabbara, Jad",
    editor = "Ku, Lun-Wei  and
      Martins, Andre  and
      Srikumar, Vivek",
    booktitle = "Proceedings of the 62nd Annual Meeting of the Association for Computational Linguistics (Volume 1: Long Papers)",
    month = aug,
    year = "2024",
    address = "Bangkok, Thailand",
    publisher = "Association for Computational Linguistics",
    url = "https://aclanthology.org/2024.acl-long.388",
    doi = "10.18653/v1/2024.acl-long.388",
    pages = "7194--7219",
}

@inbook{deKinderen2024,
  title = {Towards Taming Large Language Models with Prompt Templates for Legal GRL Modeling},
  ISBN = {9783031610073},
  ISSN = {1865-1356},
  DOI = {10.1007/978-3-031-61007-3_17},
  booktitle = {Enterprise,  Business-Process and Information Systems Modeling},
  publisher = {Springer Nature Switzerland},
  author = {de Kinderen,  Sybren and Winter,  Karolin},
  year = {2024},
  pages = {213–228}
}

@misc{sg4ayhgyha45y,
  doi = {10.48550/ARXIV.2403.09163},
  url = {https://arxiv.org/abs/2403.09163},
  author = {Mik,  Eliza},
  title = {Caveat Lector: Large Language Models in Legal Practice},
  publisher = {arXiv},
  year = {2024},
  copyright = {arXiv.org perpetual,  non-exclusive license}
}

@article{Louis2024,
  title = {Interpretable Long-Form Legal Question Answering with Retrieval-Augmented Large Language Models},
  volume = {38},
  ISSN = {2159-5399},
  url = {http://dx.doi.org/10.1609/aaai.v38i20.30232},
  DOI = {10.1609/aaai.v38i20.30232},
  number = {20},
  journal = {Proceedings of the AAAI Conference on Artificial Intelligence},
  publisher = {Association for the Advancement of Artificial Intelligence (AAAI)},
  author = {Louis,  Antoine and Van Dijck,  Gijs and Spanakis,  Gerasimos},
  year = {2024},
  month = mar,
  pages = {22266–22275}
}

@article{Homoki2024,
  title = {Large language models and their possible uses in law},
  volume = {64},
  ISSN = {2560-1067},
  url = {http://dx.doi.org/10.1556/2052.2023.00475},
  DOI = {10.1556/2052.2023.00475},
  number = {3},
  journal = {Hungarian Journal of Legal Studies},
  publisher = {Akademiai Kiado Zrt.},
  author = {Homoki,  Péter and Ződi,  Zsolt},
  year = {2024},
  month = jun,
  pages = {435–455}
}

@misc{ygeayhge574yhw5yq4ayq3,
  doi = {10.48550/ARXIV.2311.15716},
  url = {https://arxiv.org/abs/2311.15716},
  author = {Wehnert,  Sabine},
  title = {Justifiable Artificial Intelligence: Engineering Large Language Models for Legal Applications},
  publisher = {arXiv},
  year = {2023},
  copyright = {Creative Commons Attribution 4.0 International}
}

@misc{lai2023largelanguagemodelslaw,
      title={Large Language Models in Law: A Survey}, 
      author={Jinqi Lai and Wensheng Gan and Jiayang Wu and Zhenlian Qi and Philip S. Yu},
      year={2023},
      eprint={2312.03718},
      archivePrefix={arXiv},
      primaryClass={cs.CL},
      url={https://arxiv.org/abs/2312.03718}, 
}

@article{Yan2023,
  title = {Legal Challenges of Artificial Intelligence in the Field of Criminal Defense},
  volume = {30},
  ISSN = {2753-7056},
  url = {http://dx.doi.org/10.54254/2753-7048/30/20231629},
  DOI = {10.54254/2753-7048/30/20231629},
  number = {1},
  journal = {Lecture Notes in Education Psychology and Public Media},
  publisher = {EWA Publishing},
  author = {Yan,  Qiong},
  year = {2023},
  month = dec,
  pages = {167–175}
}

@article{20egfesaw24,
  title = {The Practice of Rule of Law in Criminal Justice System: A Criminological Study of Bangladesh},
  volume = {12},
  ISSN = {2349-7300},
  url = {http://dx.doi.org/10.37082/ijirmps.v12.i4.230737},
  DOI = {10.37082/ijirmps.v12.i4.230737},
  number = {4},
  journal = {International Journal of Innovative Research in Engineering and Multidisciplinary Physical Sciences},
  publisher = {International Journal of Innovative Research in Engineering and Multidisciplinary Physical Sciences},
  author = {Knock ,  Shammem Ara and Faruk, Dr. Md. Omar  and Siddiqua ,  Rukhsana and Akter,  Asma },
  year = {2024},
  month = aug 
}

@article{Raj202wfdf4,
  title = {Assessing the Legal Framework and Socioeconomic Impacts of Compensation for Wrongfully Convicted and Imprisoned Persons in Bangladesh: Challenges and Policy Recommendations},
  volume = {4},
  ISSN = {0976-7797},
  url = {http://dx.doi.org/10.47760/cognizance.2024.v04i06.028},
  DOI = {10.47760/cognizance.2024.v04i06.028},
  number = {6},
  journal = {Cognizance Journal of Multidisciplinary Studies},
  publisher = {Zain Publications},
  author = {Raj,  Md. Alamgir Sarkar},
  year = {2024},
  month = jun,
pages = {456–466}
}

@article{Islam2024afeasg,
  title = {Promoting access to justice in Bangladesh: Towards a hybrid justice model},
  volume = {77},
  ISSN = {1756-0616},
  url = {http://dx.doi.org/10.1016/j.ijlcj.2024.100655},
  DOI = {10.1016/j.ijlcj.2024.100655},
  journal = {International Journal of Law,  Crime and Justice},
  publisher = {Elsevier BV},
  author = {Islam,  Md Jahirul and Suzuki,  Masahiro and Mazumder,  Nurunnahar},
  year = {2024},
  month = {jun},
  pages = {100655}
}

@article{AKTER2017affaf,
  title = {Legal Aid for Ensuring Access to Justice in Bangladesh: A Paradox?},
  volume = {4},
  ISSN = {2052-9023},
  url = {http://dx.doi.org/10.1017/ALS.2016.60},
  number = {1},
  journal = {Asian Journal of Law and Society},
  publisher = {Cambridge University Press (CUP)},
  author = {Akter,  Farzana},
  year = {2017},
  month = {jan},
  pages = {257–275}
}

@article{tahura2025legalaid,
  author       = {Tahura, Ummey Sharaban and Alam, Shawkat},
  title        = {Legal Aid, Systemic Barriers, and Access to Justice in Bangladesh},
  year         = {2025},
  month        = apr,
  url          = {https://www.thedailystar.net/law-our-rights/news/legal-aid-systemic-barriers-and-access-justice-bangladesh-3874031},
  note         = {Accessed: 2025-06-23},
  publisher    = {The Daily Star}
}

@misc{wai2024exp,
      title={Exploring Possibilities of AI-Powered Legal Assistance in Bangladesh through Large Language Modeling}, 
      author={Azmine Toushik Wasi and Wahid Faisal and Mst Rafia Islam and Mahathir Mohammad Bappy},
      year={2024},
      eprint={2410.17210},
      archivePrefix={arXiv},
      primaryClass={cs.CL},
      url={https://arxiv.org/abs/2410.17210}, 
}

@article{asjad2023bangla,
  author       = {Asjad, Tanim},
  title        = {Use of Bangla in Judicial System},
  year         = {2023},
  month        = feb,
  url          = {https://thefinancialexpress.com.bd/views/opinions/use-of-bangla-in-judicial-system},
  note         = {Accessed: 2025-06-22},
  publisher    = {The Financial Express}
}

@article{mizan2021language,
  author       = {Mizan, Arpeeta Shams},
  title        = {The Law of Language and the Language of the Law: A Sociolegal Appraisal of Colonial Legal Language in Bangladesh},
  journal      = {Dhaka University Law Journal},
  year         = {2021},
  volume       = {32},
  number       = {2},
  pages        = {119--134},
  doi          = {10.3329/dulj.v32i2.57961},
  publisher    = {Dhaka University Law Journal}
}

@article{safdie2025aiforjustice,
  author       = {Safdie, Laura},
  title        = {AI and Legal Aid: A Generational Opportunity for Access to Justice},
  year         = {2025},
  month        = feb,
  url          = {https://www.thomsonreuters.com/en-us/posts/ai-in-courts/ai-legal-aid-generational-opportunity/},
  note         = {Accessed: 2025-06-23},
  publisher    = {Thomson Reuters}
}

@inbook{Mamalis2024,
  title = {A Large Language Model Agent Based Legal Assistant for Governance Applications},
  ISBN = {9783031702747},
  ISSN = {1611-3349},
  url = {http://dx.doi.org/10.1007/978-3-031-70274-7_18},
  DOI = {10.1007/978-3-031-70274-7_18},
  booktitle = {Electronic Government},
  publisher = {Springer Nature Switzerland},
  author = {Mamalis,  Marios Evangelos and Kalampokis,  Evangelos and Fitsilis,  Fotios and Theodorakopoulos,  Georgios and Tarabanis,  Konstantinos},
  year = {2024},
  pages = {286–301}
}

@misc{li2024legalagentbenchevaluatingllmagents,
      title={LegalAgentBench: Evaluating LLM Agents in Legal Domain}, 
      author={Haitao Li and Junjie Chen and Jingli Yang and Qingyao Ai and Wei Jia and Youfeng Liu and Kai Lin and Yueyue Wu and Guozhi Yuan and Yiran Hu and Wuyue Wang and Yiqun Liu and Minlie Huang},
      year={2024},
      eprint={2412.17259},
      archivePrefix={arXiv},
      primaryClass={cs.CL},
      url={https://arxiv.org/abs/2412.17259}, 
}

@misc{cui2024chatlawmultiagentcollaborativelegal,
      title={Chatlaw: A Multi-Agent Collaborative Legal Assistant with Knowledge Graph Enhanced Mixture-of-Experts Large Language Model}, 
      author={Jiaxi Cui and Munan Ning and Zongjian Li and Bohua Chen and Yang Yan and Hao Li and Bin Ling and Yonghong Tian and Li Yuan},
      year={2024},
      eprint={2306.16092},
      archivePrefix={arXiv},
      primaryClass={cs.CL},
      url={https://arxiv.org/abs/2306.16092}, 
}

@inproceedings{shengbinyue-etal-2025-multi,
    title = "Multi-Agent Simulator Drives Language Models for Legal Intensive Interaction",
    author = "Yue, Shengbin  and
      Huang, Ting  and
      Jia, Zheng  and
      Wang, Siyuan  and
      Liu, Shujun  and
      Song, Yun  and
      Huang, Xuanjing  and
      Wei, Zhongyu",
    editor = "Chiruzzo, Luis  and
      Ritter, Alan  and
      Wang, Lu",
    booktitle = "Findings of the Association for Computational Linguistics: NAACL 2025",
    month = apr,
    year = "2025",
    address = "Albuquerque, New Mexico",
    publisher = "Association for Computational Linguistics",
    url = "https://aclanthology.org/2025.findings-naacl.365/",
    doi = "10.18653/v1/2025.findings-naacl.365",
    pages = "6537--6570",
    ISBN = "979-8-89176-195-7"
}

@misc{jiang2024agentsbenchlargelanguage,
      title={Agents on the Bench: Large Language Model Based Multi Agent Framework for Trustworthy Digital Justice}, 
      author={Cong Jiang and Xiaolei Yang},
      year={2024},
      eprint={2412.18697},
      archivePrefix={arXiv},
      primaryClass={cs.AI},
      url={https://arxiv.org/abs/2412.18697}, 
}

@misc{liang2025llmpoweredaiagentsystems,
      title={LLM-Powered AI Agent Systems and Their Applications in Industry}, 
      author={Guannan Liang and Qianqian Tong},
      year={2025},
      eprint={2505.16120},
      archivePrefix={arXiv},
      primaryClass={cs.AI},
      url={https://arxiv.org/abs/2505.16120}, 
}

@misc{cohere2025commandaenterprisereadylarge,
      title={Command A: An Enterprise-Ready Large Language Model}, 
      author={Team Cohere and : and Aakanksha and Arash Ahmadian and Marwan Ahmed and Jay Alammar and Milad Alizadeh and Yazeed Alnumay and Sophia Althammer and Arkady Arkhangorodsky and Viraat Aryabumi and Dennis Aumiller and Raphaël Avalos and Zahara Aviv and Sammie Bae and Saurabh Baji and Alexandre Barbet and Max Bartolo and Björn Bebensee and Neeral Beladia and Walter Beller-Morales and Alexandre Bérard and Andrew Berneshawi and Anna Bialas and Phil Blunsom and Matt Bobkin and Adi Bongale and Sam Braun and Maxime Brunet and Samuel Cahyawijaya and David Cairuz and Jon Ander Campos and Cassie Cao and Kris Cao and Roman Castagné and Julián Cendrero and Leila Chan Currie and Yash Chandak and Diane Chang and Giannis Chatziveroglou and Hongyu Chen and Claire Cheng and Alexis Chevalier and Justin T. Chiu and Eugene Cho and Eugene Choi and Eujeong Choi and Tim Chung and Volkan Cirik and Ana Cismaru and Pierre Clavier and Henry Conklin and Lucas Crawhall-Stein and Devon Crouse and Andres Felipe Cruz-Salinas and Ben Cyrus and Daniel D'souza and Hugo Dalla-Torre and John Dang and William Darling and Omar Darwiche Domingues and Saurabh Dash and Antoine Debugne and Théo Dehaze and Shaan Desai and Joan Devassy and Rishit Dholakia and Kyle Duffy and Ali Edalati and Ace Eldeib and Abdullah Elkady and Sarah Elsharkawy and Irem Ergün and Beyza Ermis and Marzieh Fadaee and Boyu Fan and Lucas Fayoux and Yannis Flet-Berliac and Nick Frosst and Matthias Gallé and Wojciech Galuba and Utsav Garg and Matthieu Geist and Mohammad Gheshlaghi Azar and Ellen Gilsenan-McMahon and Seraphina Goldfarb-Tarrant and Tomas Goldsack and Aidan Gomez and Victor Machado Gonzaga and Nithya Govindarajan and Manoj Govindassamy and Nathan Grinsztajn and Nikolas Gritsch and Patrick Gu and Shangmin Guo and Kilian Haefeli and Rod Hajjar and Tim Hawes and Jingyi He and Sebastian Hofstätter and Sungjin Hong and Sara Hooker and Tom Hosking and Stephanie Howe and Eric Hu and Renjie Huang and Hemant Jain and Ritika Jain and Nick Jakobi and Madeline Jenkins and JJ Jordan and Dhruti Joshi and Jason Jung and Trushant Kalyanpur and Siddhartha Rao Kamalakara and Julia Kedrzycki and Gokce Keskin and Edward Kim and Joon Kim and Wei-Yin Ko and Tom Kocmi and Michael Kozakov and Wojciech Kryściński and Arnav Kumar Jain and Komal Kumar Teru and Sander Land and Michael Lasby and Olivia Lasche and Justin Lee and Patrick Lewis and Jeffrey Li and Jonathan Li and Hangyu Lin and Acyr Locatelli and Kevin Luong and Raymond Ma and Lukáš Mach and Marina Machado and Joanne Magbitang and Brenda Malacara Lopez and Aryan Mann and Kelly Marchisio and Olivia Markham and Alexandre Matton and Alex McKinney and Dominic McLoughlin and Jozef Mokry and Adrien Morisot and Autumn Moulder and Harry Moynehan and Maximilian Mozes and Vivek Muppalla and Lidiya Murakhovska and Hemangani Nagarajan and Alekhya Nandula and Hisham Nasir and Shauna Nehra and Josh Netto-Rosen and Daniel Ohashi and James Owers-Bardsley and Jason Ozuzu and Dennis Padilla and Gloria Park and Sam Passaglia and Jeremy Pekmez and Laura Penstone and Aleksandra Piktus and Case Ploeg and Andrew Poulton and Youran Qi and Shubha Raghvendra and Miguel Ramos and Ekagra Ranjan and Pierre Richemond and Cécile Robert-Michon and Aurélien Rodriguez and Sudip Roy and Sebastian Ruder and Laura Ruis and Louise Rust and Anubhav Sachan and Alejandro Salamanca and Kailash Karthik Saravanakumar and Isha Satyakam and Alice Schoenauer Sebag and Priyanka Sen and Sholeh Sepehri and Preethi Seshadri and Ye Shen and Tom Sherborne and Sylvie Shang Shi and Sanal Shivaprasad and Vladyslav Shmyhlo and Anirudh Shrinivason and Inna Shteinbuk and Amir Shukayev and Mathieu Simard and Ella Snyder and Ava Spataru and Victoria Spooner and Trisha Starostina and Florian Strub and Yixuan Su and Jimin Sun and Dwarak Talupuru and Eugene Tarassov and Elena Tommasone and Jennifer Tracey and Billy Trend and Evren Tumer and Ahmet Üstün and Bharat Venkitesh and David Venuto and Pat Verga and Maxime Voisin and Alex Wang and Donglu Wang and Shijian Wang and Edmond Wen and Naomi White and Jesse Willman and Marysia Winkels and Chen Xia and Jessica Xie and Minjie Xu and Bowen Yang and Tan Yi-Chern and Ivan Zhang and Zhenyu Zhao and Zhoujie Zhao},
      year={2025},
      eprint={2504.00698},
      archivePrefix={arXiv},
      primaryClass={cs.CL},
      url={https://arxiv.org/abs/2504.00698}, 
}

@misc{yong2025statemultilingualllmsafety,
      title={The State of Multilingual LLM Safety Research: From Measuring the Language Gap to Mitigating It}, 
      author={Zheng-Xin Yong and Beyza Ermis and Marzieh Fadaee and Stephen H. Bach and Julia Kreutzer},
      year={2025},
      eprint={2505.24119},
      archivePrefix={arXiv},
      primaryClass={cs.CL},
      url={https://arxiv.org/abs/2505.24119}, 
}

@misc{peppin2025multilingualdivideimpactglobal,
      title={The Multilingual Divide and Its Impact on Global AI Safety}, 
      author={Aidan Peppin and Julia Kreutzer and Alice Schoenauer Sebag and Kelly Marchisio and Beyza Ermis and John Dang and Samuel Cahyawijaya and Shivalika Singh and Seraphina Goldfarb-Tarrant and Viraat Aryabumi and Aakanksha and Wei-Yin Ko and Ahmet Üstün and Matthias Gallé and Marzieh Fadaee and Sara Hooker},
      year={2025},
      eprint={2505.21344},
      archivePrefix={arXiv},
      primaryClass={cs.AI},
      url={https://arxiv.org/abs/2505.21344}, 
}
\newpage
\appendix

\section{Related Work}
\subsection{Legal NLP Research}
Research in Legal Natural Language Processing and AI applications in law has advanced considerably, particularly for high-resource languages like English. Existing systems tackle diverse tasks such as legal document summarization, named entity recognition, question answering, contract analysis, and judgment prediction \citep{ariai2025naturallanguageprocessinglegal}. Tools like ROSS Intelligence and Westlaw Edge automate research and e-discovery, while AI chatbots like DoNotPay assist with document drafting. Recent developments integrate large language models (LLMs) with logic programming and multi-agent frameworks leveraging knowledge graphs and Retrieval-Augmented Generation (RAG) to improve reliability and contextual accuracy. Furthermore, LLM-based methods employing specialized prompts and curated datasets for LegalAI tasks \citep{huang2024optimizingnumericalestimationoperational}, as well as models like Mistral and Gemma for judicial entity extraction \citep{hussain2024largelanguagemodelsjudicial}, demonstrate superior efficiency and cost-effectiveness over traditional practices \citep{sun2024lawluochineselawfirm}. RAG and agent-based approaches enhance governance and long-form legal question answering \citep{mamalis_kalampokis_fitsilis_theodorakopoulos_tarabanis_2024}, while storytelling and prompt templates aid complex concept processing and text structuring \citep{jiang-etal-2024-leveraging,deKinderen2024}. However, LLM limitations, such as stochastic outputs, hallucinations \citep{Dahl20segfeg24}, and fact-checking gaps \citep{sg4ayhgyha45y} necessitate more interpretable, evidence-based models \citep{Louis2024}, especially for domain-sensitive contexts like law.

\subsection{Agent-focused Legal NLP Research}
Prior work includes a GDPR-focused RAG and agent-based assistant using GPT‑3.5/4 over EU legislation to answer complex queries, showing promising results with precision varying by query difficulty \citep{Mamalis2024}. Chinese-language benchmarks like LegalAgentBench evaluate LLM agent performance across 17 corpora and tool-augmented workflows, offering nuanced metrics on reasoning and multi-hop tasks \citep{li2024legalagentbenchevaluatingllmagents}. Multi-agent systems have also been explored: Chatlaw uses a MoE + knowledge graph approach to reduce hallucinations and mimic law‑firm workflows \citep{cui2024chatlawmultiagentcollaborativelegal}, while frameworks like MASER and AgentsCourt simulate interactive legal scenarios and judicial deliberation using coordinated LLM-driven agents \citep{shengbinyue-etal-2025-multi,jiang2024agentsbenchlargelanguage}. Broader surveys of LLM agents detail architectures integrating RAG, hierarchical planning, and safety mechanisms for domain-focused applications \citep{liang2025llmpoweredaiagentsystems,yong2025statemultilingualllmsafety}. Together, these efforts highlight a trend toward modular, tool-enhanced LLM agents tailored to legal reasoning, retrieval, and collaborative decision-making, forming the foundation for governance-focused assistants.

\subsection{Multilingual Legal NLP Research}
Multilingual NLP research, crucial for low-resource languages, faces challenges from data scarcity, script variations, and cultural nuances \citep{yong2025statemultilingualllmsafety,peppin2025multilingualdivideimpactglobal}. Despite multilingual models like GPT-4 showing promise, performance degradation persists for underrepresented languages such as Bengali. Initiatives like MultiLegalPile \citep{qin2024exploringnexuslargelanguage} and legal text comprehension studies \citep{martin2024bettergptcomparinglarge,ygeayhge574yhw5yq4ayq3,Homoki2024} aim to bridge this gap, though high-resource biases remain. Moreover, South Asian legal AI development is nascent, with limited empirical systems addressing jurisdiction-specific complexities. Tailoring LLMs to local law firms and regional practices \citep{Homoki2024} could enhance operational relevance.
\citet{wai2024exp} explored enhancing Bangla capabilities of LLMs by fine-tuning GPT-2; however, the model still lacks robustness and comprehensive linguistic coverage.

Together, these pilot projects highlight AI’s potential to bridge justice gaps by automating legal research, document generation, and offering chatbot-based assistance. However, there remains a lack of deployed and empirically evaluated multilingual legal assistant systems that address the unique socio-legal and linguistic complexities of low resource but high population countries like Bangladesh. The proposed system seeks to fill this gap by providing a jurisdiction-specific, linguistically adapted, and socio-economically informed AI solution tailored to Bangladesh’s legal environment.

\begin{figure*}
    \centering
    \includegraphics[width=\linewidth]{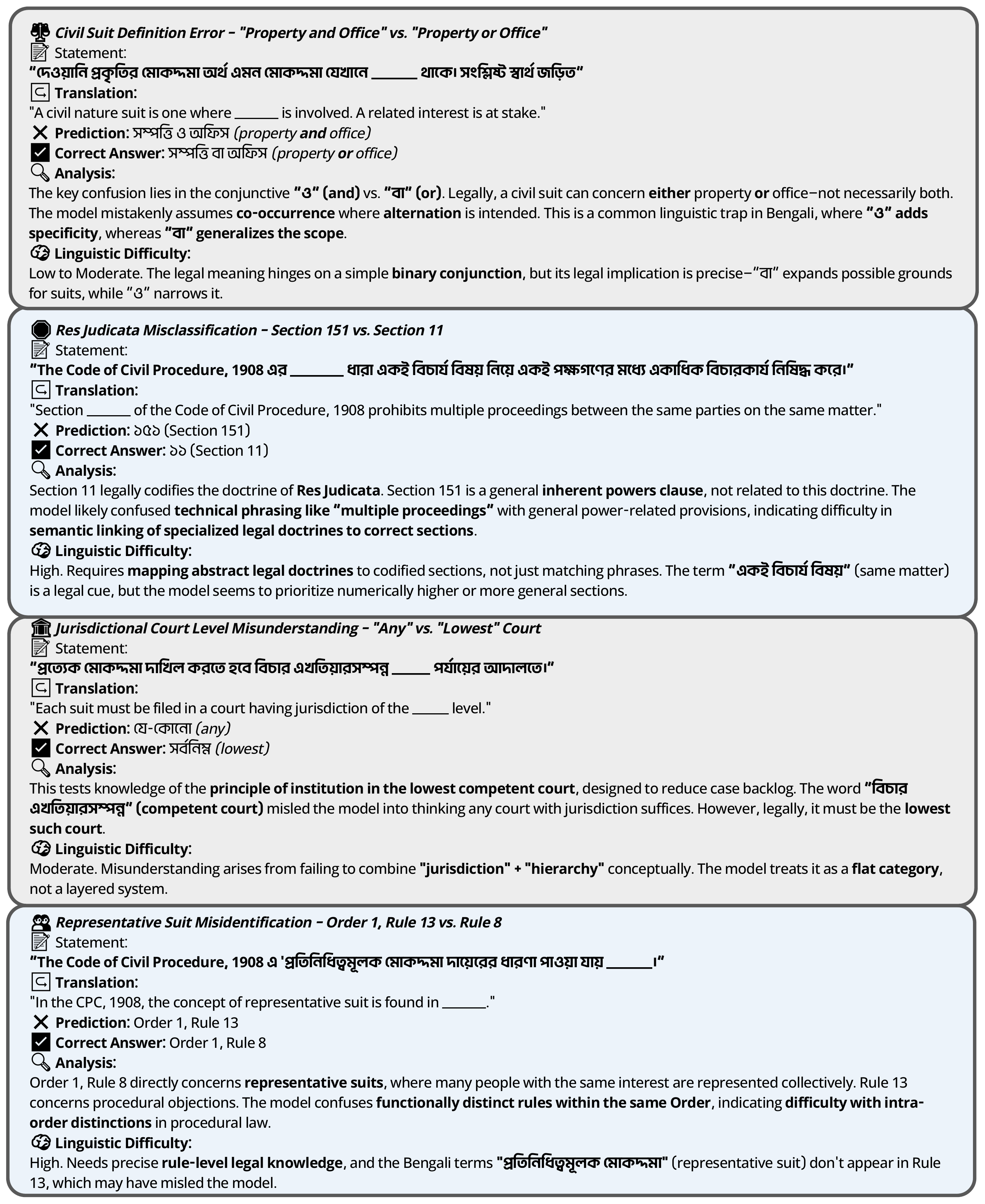}
    \caption{Error Analysis (Command-A model examples)}
    \label{fig:ErrorAnalysis}
\end{figure*}

\section{Broader Impact}
While the system was developed and evaluated within the context of Bangladesh's legal framework, it is designed to be modular and adaptable, allowing potential deployment in other jurisdictions and low-resource languages. By replacing the legal corpus and updating the legal dictionary, the framework can be customized to different legal systems. Additionally, the use of multilingual embeddings facilitates scaling to languages beyond Bengali, including those with limited digital resources. This adaptability broadens the potential impact of the system, enabling cost-effective and contextually accurate AI-assisted legal reasoning in diverse global settings. Beyond the Bangladeshi context, such a framework could support legal education, case preparation, and access-to-justice initiatives in other low-resource or underrepresented jurisdictions, thereby promoting equitable access to legal knowledge and professional assistance worldwide.

\section{System Demonstration}
\begin{figure*}[h]
    \centering
    \includegraphics[width=\linewidth]{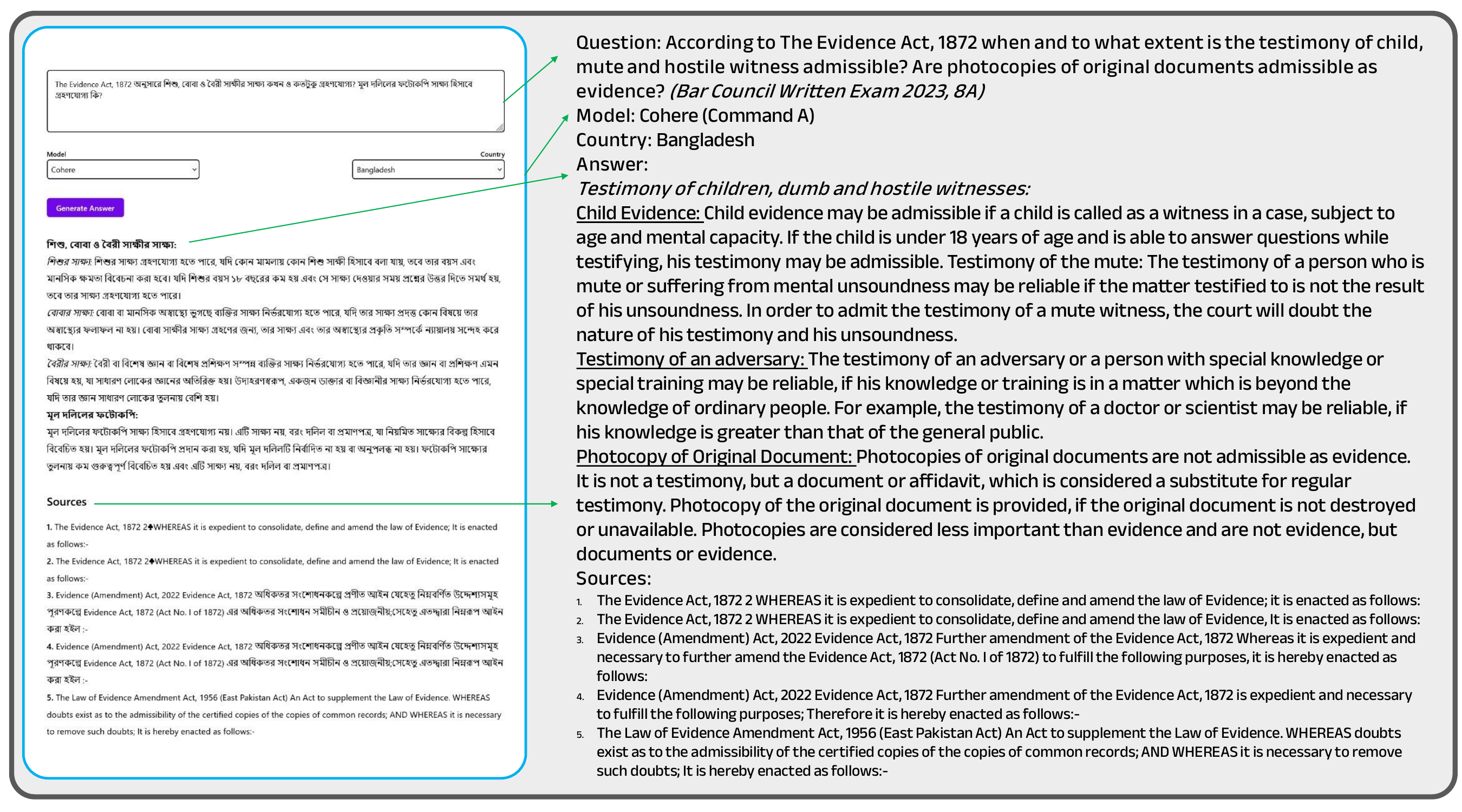}
    \caption{System Demonstration: UI and deployable system of \textsc{Mina}. }
    \label{fig:SD}
\end{figure*}

Figure \ref{fig:SD} shows a demonstration of our system and UI.

\section{Additional Tools for \textsc{Mina}}

Our system integrates a suite of specialized tools that collectively enable structured, context-sensitive, and socially grounded legal reasoning. Each tool performs a distinct operational role within the overall workflow, from document parsing to contextual augmentation, ensuring that both procedural accuracy and interpretive sensitivity are maintained throughout the pipeline. The following subsections describe these tools in detail, outlining their design rationale, underlying mechanisms, and example applications.

\subsection{File Content Reader}
The \textbf{File Content Reader} handles ingestion of uploaded materials across multiple formats, including \textit{.pdf}, \textit{.docx}, and \textit{.pptx}. This module employs temporary file storage to maintain data privacy and uses format-specific parsers such as \texttt{python-docx}, \texttt{pdfminer.six}, and \texttt{python-pptx} to extract clean, structured text. Non-textual artifacts (e.g., images, headers, footers, and embedded metadata) are automatically filtered to yield context-preserving, analysis-ready content.  

\begin{quote}
Example: A user uploads a High Court Division judgment in \textit{.pdf} format. The tool parses all pages, removes page numbers and footers, and produces a consolidated, section-labeled text block for semantic indexing. The processed text becomes directly usable for RAG-based retrieval or prompt conditioning, minimizing noise in subsequent reasoning steps.
\end{quote}

\subsection{Keyword Generator}
The \textbf{Keyword Generator} assists the retrieval-augmented generation (RAG) stage by producing a compact set of 5–10 semantically rich keywords derived from a user query or case prompt. It uses a lightweight LLM for semantic abstraction and includes a regular-expression-based fallback that ensures robust keyword extraction even under low-context or ambiguous input conditions.  

\begin{quote}
Example: For the prompt “Draft a writ petition challenging unlawful termination under labor law,” the tool generates keyword clusters such as \texttt{["writ petition", "termination", "Bangladesh Labour Act", "fundamental rights", "Article 102", "judicial review"]}. These keywords guide focused web retrieval and internal database searches, improving retrieval precision for domain-specific queries.
\end{quote}

\subsection{Web Search and Web Page Parser}
To integrate external knowledge dynamically, the \textbf{Web Search} module relies on DuckDuckGo’s query interface to fetch the top-ranked 3–5 search results, including page titles, URLs, and content snippets. When in-depth reading is required, the \textbf{Web Page Parser} uses \texttt{BeautifulSoup} to extract up to 5,000 characters of visible text while stripping scripts, navigation menus, and style elements.  

\begin{quote}
Example: When tasked with “Summarize recent Supreme Court decisions on anticipatory bail in Bangladesh,” the modules work jointly to retrieve credible sources (e.g., \textit{The Daily Star Law \& Our Rights} or \textit{Bangladesh Supreme Court Online}) and return excerpted paragraphs suitable for summarization or citation.
\end{quote}

\subsection{Question Relevance Analyzer}
The \textbf{Question Relevance Analyzer} maintains conceptual coherence in multi-turn dialogue by embedding both current and previous queries into a shared semantic space. Cosine similarity between embeddings determines whether the new query logically extends or diverges from the ongoing conversation.  

\begin{quote}
Example: When the user transitions from “Draft an appeal under Section 96 CPC” to “What are the grounds for revision?”, the analyzer detects moderate semantic overlap but distinct procedural context, prompting a structured reset rather than a contextual merge.
\end{quote}

\subsection{Chat Analyzer}
The \textbf{Chat Analyzer} provides an additional layer of context-awareness in conversational mode. It reviews prior dialogue to reconstruct temporal dependencies, topic continuity, and user preferences, performing discourse-level inference to recognize patterns such as clarification requests, corrections, or elaboration prompts.  

\begin{quote}
Example: If a user first says “Draft a plaint for defamation” and later asks “Now make it fit for filing before the Joint District Judge,” the Chat Analyzer detects continuity and refines the procedural formatting automatically.
\end{quote}

\subsection{Legal Dictionary}
The \textbf{Legal Dictionary} module provides culturally adapted interpretive functionality. It explains colonial-era, Farsi-influenced, or archaic terms still prevalent in Bangladeshi legal texts. Each term is annotated with both its statutory definition and contemporary contextual meaning.  

\begin{quote}
Example: Input contains \textit{“naraji”} (objection petition). The module outputs: “\textit{‘Naraji’ refers to a formal objection against a police report, typically filed under Section 173(3) of the CrPC, asserting dissatisfaction with the investigation outcome.}” This facilitates bilingual legal comprehension across English-Bengali legal ecosystems.
\end{quote}

\subsection{Socio-Economic Simulation}
The \textbf{Socio-Economic Simulation} module models how demographic and socio-economic variables influence access to justice, acknowledging that legal outcomes in Bangladesh often correlate with income, occupation, literacy, and geography.  

\begin{quote}
Example: Consider Bar Council Exam 2023, Question 4(b) (translated): “X assaults Y. Enraged, Y reacts violently. Taking advantage of Y’s anger, Z hands Y a knife intending Y will kill X. On 03/11/2023, Y kills X at Rasulpur Bazaar. Police charges Y and Z under Sections 109, 34, 323, and 326 of the Penal Code, 1860. Prepare a ‘naraji’ petition before the court, determining their criminal liability.”

Simulation Input: 
- Accused Y: rural agricultural worker, monthly income 12,000 BDT, low literacy  
- Accused Z: urban businessman  
- Victim X: marginalized community member

Simulation Output: The tool predicts Y’s higher likelihood of prolonged pre-trial detention due to inability to post bail, while Z is more likely to secure early representation. It highlights systemic inequities ,  affordability, case delays, and social bias ,  that influence actual justice outcomes beyond statutory law. This module transforms the system into a socio-legal diagnostic platform rather than a purely doctrinal AI.
\end{quote}

\section{Additional Information on Experiments and Evaluation} \label{sec:apx-extra-exp-eval}

\subsection{Evaluators}
The evaluation of written and viva responses was conducted by law faculty members from leading Bangladeshi universities. For the viva voce, five evaluators independently scored each response, while three evaluators assessed the written examination. The evaluators applied the same criteria employed in real Bar Council examinations, including accuracy, clarity, contextual understanding, and legal reasoning. No additional training was provided, as the evaluators were already familiar with the official standards and marking rubrics. This approach ensured that the assessment closely mirrors real-world examination practices while maintaining consistency, fairness, and reliability across both the written and oral components.

\subsection{Evaluation Process of Written Exams}
\begin{figure*}[t]
    \centering
    \includegraphics[width=\linewidth]{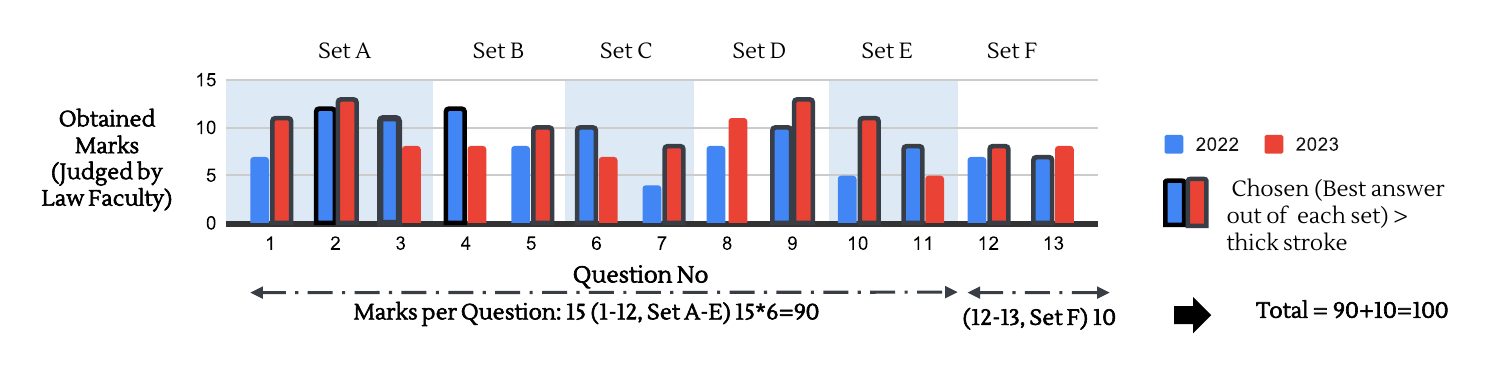}
    \caption{Breaking Down a Written Full Answer (Command-A, Two Step; Examiner 2)}
    \label{fig:WritenBreakDown}
\end{figure*}

The written evaluation of \textsc{Mina} was conducted by a panel of five law faculty members from leading Bangladeshi universities, each evaluating the system’s responses to all 13 questions of the 2022 and 2023 Bangladesh Bar Council Exams. Evaluators assessed answers along four dimensions: accuracy, clarity, contextual understanding, and legal reasoning. Each question was scored numerically (15 marks for Sets A–E, 10 marks for Set F), and the five scores were averaged to obtain a composite mark per question. The system generated answers for all questions, after which the model selected the “best answer” within each set with another prompt, reflecting realistic candidate strategy. Detailed qualitative feedback was provided for each answer, highlighting strengths such as logical structure, precise statutory references, and coherent reasoning. Weaknesses were noted in stepwise computation of limitation periods, explicit allocation of evidentiary burdens, and critical interpretation of procedural rules. Comparisons across 2022 and 2023 in Figure \ref{fig:WritenBreakDown} showed minor year-to-year variations, with improvement in complex procedural questions such as injunctions and adjournments. Overall, the best-selected answers yielded a total score within 75–80\% in good models, aligning with or surpassing average human candidate performance.

The evaluators’ feedback emphasized both doctrinal competence and practical procedural awareness (Appendix \ref{sec:apx-EvaluatorResponseAnalysis-written}). Comments highlighted the need to explicitly state preconditions, include stepwise calculations of limitation periods, and clarify evidentiary thresholds to achieve distinction-level responses. Some answers applied rules formulaically without sufficient critical analysis or discussion of judicial discretion. High-performing responses were praised for smooth argument transitions, clear mapping from statute to facts, and structured reasoning. Observed patterns indicate that \textsc{Mina} is strong in statutory interpretation and logical drafting but could benefit from enhancements in contextual sensitivity and explicit procedural reasoning. The selection of seven best answers for final scoring ensured realistic aggregation, reflecting typical candidate exam strategy. Overall, the evaluation demonstrates that a well-tuned LLM-based legal assistant can produce exam-quality written responses while revealing actionable areas for improvement in legal reasoning and clarity.

\subsection{Inter-Annotator Agreement  for Written Evaluation}

To quantify consistency among evaluators of the written exams, we calculated Cohen’s $\kappa$ ($\kappa$) between all pairs of evaluators. Each evaluator scored the 13 questions numerically along four dimensions: accuracy, clarity, contextual understanding, and legal reasoning. Cohen’s $\kappa$ is computed as:

\begin{equation}
\kappa = \frac{p_o - p_e}{1 - p_e}
\end{equation}

where $p_o$ is the observed proportion of agreement between evaluators and $p_e$ is the expected agreement by chance. Across all questions and evaluator pairs, we obtained $\kappa = 0.827$, indicating high consistency. This demonstrates that, even with numerical scoring, our evaluation protocol yields reliable and reproducible assessments. Minor variations were observed in a few complex procedural questions, but overall agreement remained strong, supporting the robustness of our written exam evaluation methodology.

\subsection{Quality Assessment of Retrieved Sections}
In our two-stage RAG pipeline, the relevance of retrieved legal sections is explicitly assessed before final response generation. An LLM-based filtering step evaluates each retrieved section against the user query using the prompt:

\begin{quote}
"Given the user query and the retrieved document section, determine whether the section contains information directly relevant to answering the query. Respond with 'relevant' or 'irrelevant'."
\end{quote}

If a section is judged irrelevant, the system automatically refines the query by adjusting keywords and repeats retrieval until relevant sections are identified. This ensures that only contextually appropriate legal content informs the final response. By incorporating this automatic relevance checking, the pipeline mitigates the risk of hallucinating or misapplying unrelated statutes, maintaining both precision and legal coherence in generated answers.

\section{Detailed Error Analysis} \label{sec:apx-error-analysis}
\subsection{MCQ}  \label{sec:apx-error-analysis-mcq}
Some error analysis is added in Figure \ref{fig:ErrorAnalysis}.

Analysis of these errors reveals systematic patterns in the model’s handling of Bangladeshi civil procedure law. One recurring issue involves \textbf{linguistic conjunctions} such as “O” (and) versus “ba” (or). Even in low-complexity sentences, these small lexical variations drastically alter the legal scope of a suit. The model often fails to capture this sensitivity to subtle semantic cues in Bengali, suggesting that linguistic nuance remains an underdeveloped area of comprehension. This limitation shows that even minor syntactic shifts can lead to substantive legal misinterpretation.

A second category of errors involves \textbf{jurisdictional hierarchy}, particularly regarding which court is competent to hear a given suit. The model frequently confuses filing requirements, such as whether to file in the lowest competent court or in any court, indicating an incomplete understanding of procedural layering. Instead of recognizing hierarchical structures, it tends to flatten them into broad, undifferentiated categories. This behavior reflects a lack of integration between procedural reasoning and lexical parsing. As a result, the model’s interpretation of jurisdiction remains largely surface-level rather than conceptually grounded.

A third major error type concerns \textbf{doctrinal mapping}. Misclassification of Res Judicata under Section 151 instead of Section 11, for example, demonstrates the model’s difficulty in linking abstract legal doctrines to their precise statutory locations. Such mistakes reveal an inability to align high-level legal semantics with codified provisions. Similarly, representative suit misidentification between Order 1, Rules 8 and 13 shows that intra-order distinctions, which depend on functional understanding, remain particularly challenging. These examples underscore that the model relies too heavily on surface lexical or numeric cues, neglecting the deeper legal logic underlying procedural distinctions.

Overall, these error clusters often occur where procedural rules encode exceptions, hierarchical logic, or specialized doctrines. This pattern indicates that retrieval or memorization alone is insufficient without robust legal-context reasoning. The range of linguistic complexity, spanning from binary conjunctions to abstract doctrinal principles, further complicates interpretation. Even seemingly simple linguistic features can carry disproportionate legal weight, while complex doctrines require multi-step mapping between text, legal principle, and statute. These difficulties illustrate the importance of multi-layered reasoning over shallow textual matching.

Finally, the observed error patterns highlight concrete directions for improvement. Future models should incorporate structured legal knowledge bases and hierarchical reasoning frameworks to better interpret procedural relationships. Fine-grained term disambiguation must be strengthened to handle Bengali’s semantic density and legal specificity. In addition, targeted augmentation using legal-specific retrieval, rule-based constraints, or contrastive learning could significantly reduce misclassifications. Such interventions are especially critical for low-resource languages like Bengali, where precise statutory alignment and doctrinal reasoning are essential for legal reliability.

\begin{figure}
    \centering
    \includegraphics[width=\linewidth]{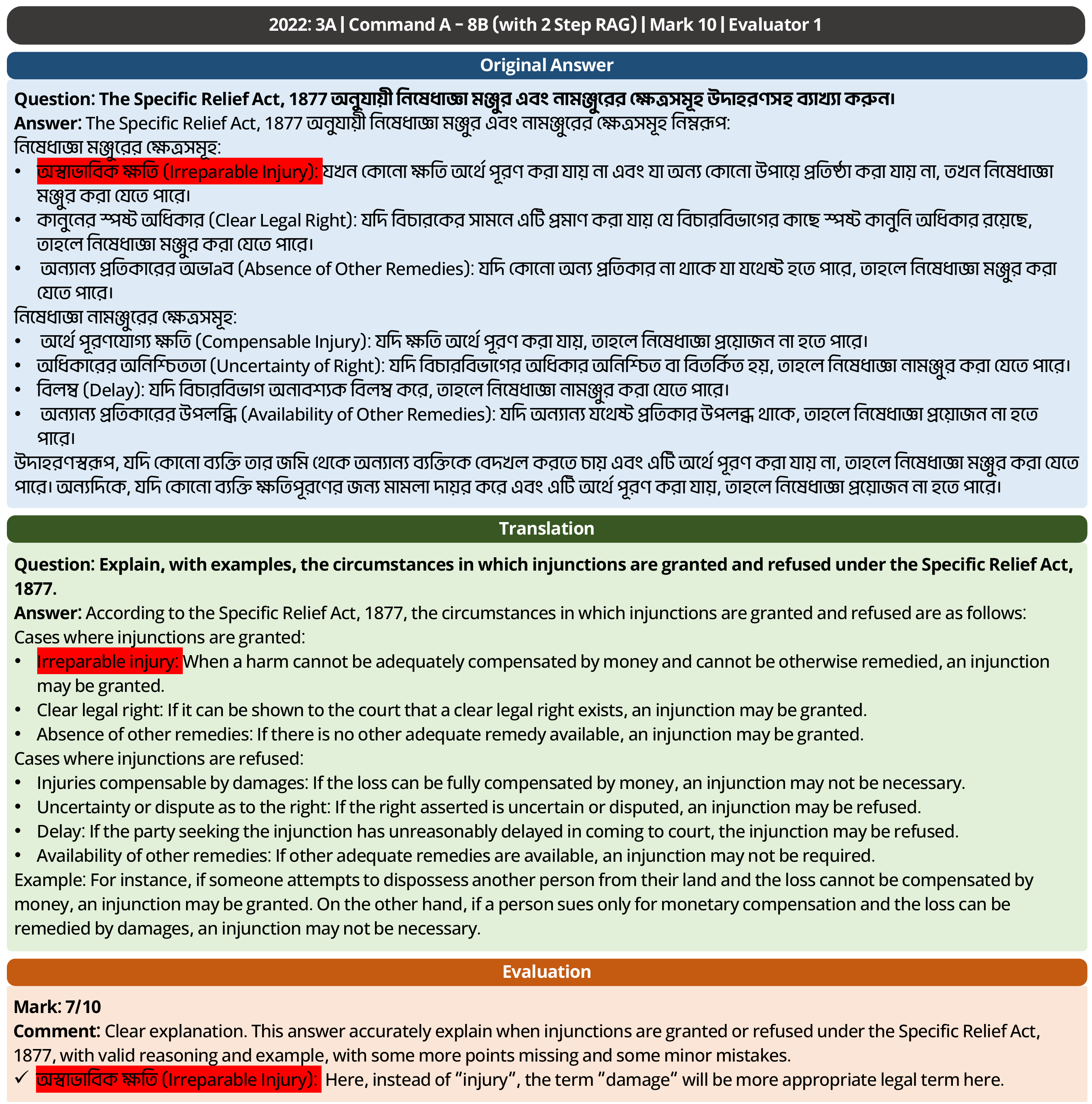}
    \caption{Written exam examples for qualitative error analysis (Part 1)}
    \label{fig:eq-qual-1}
\end{figure}

\begin{figure}
    \centering
    \includegraphics[width=\linewidth]{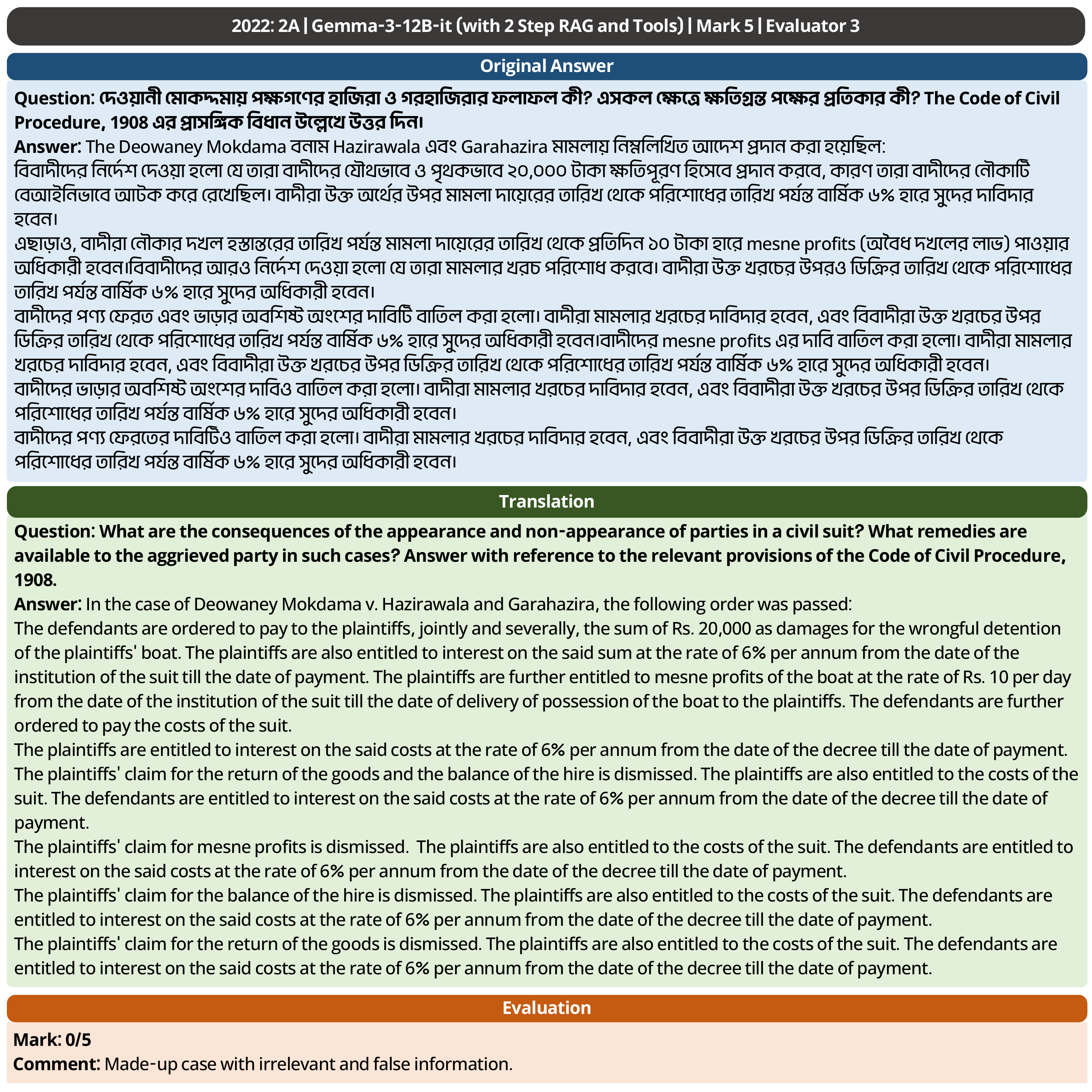}
    \caption{Written exam examples for qualitative error analysis (Part 2)}
    \label{fig:eq-qual-2}
\end{figure}
\begin{figure}
    \centering
    \includegraphics[width=\linewidth]{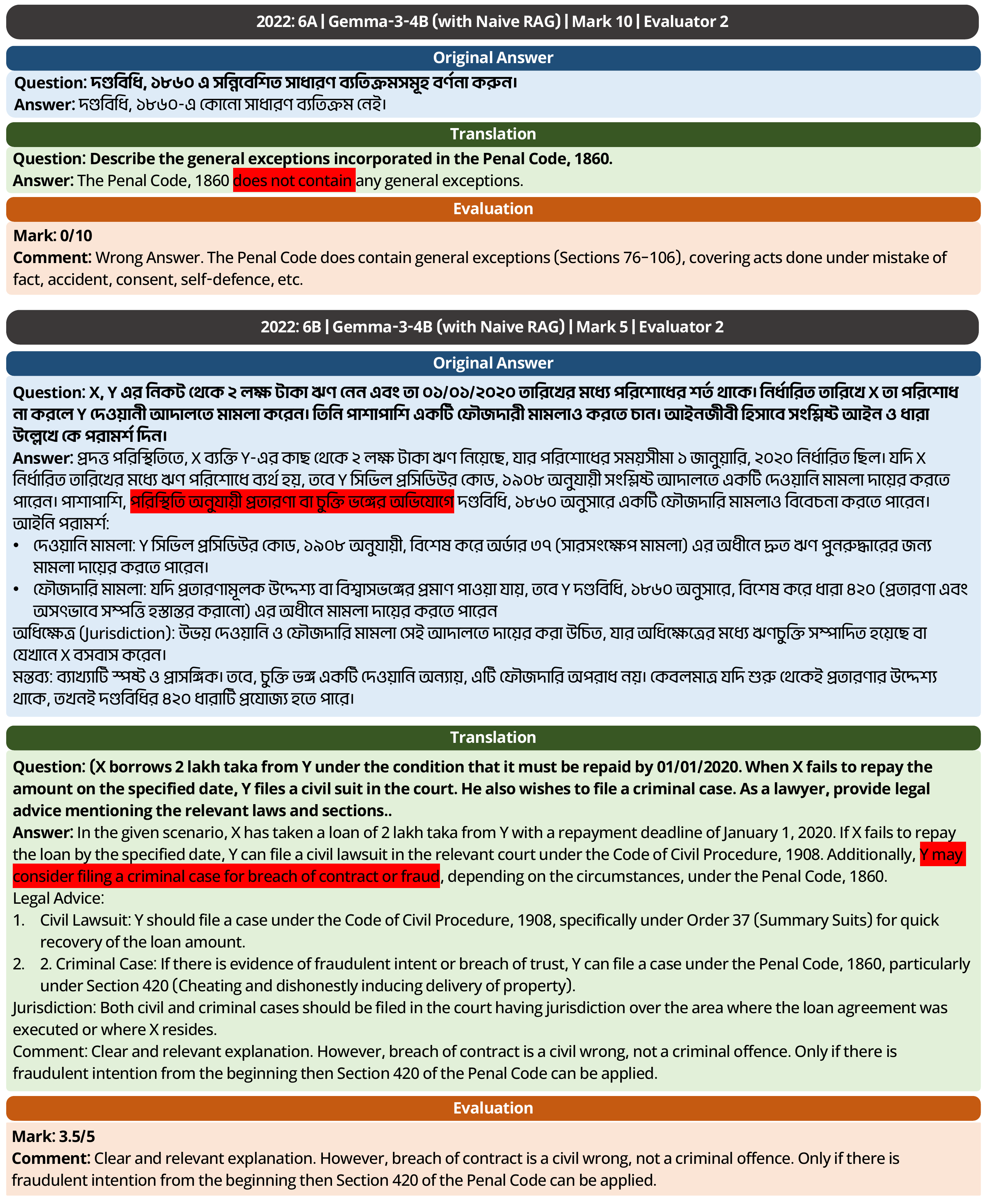}
    \caption{Written exam examples for qualitative error analysis (Part 3)}
    \label{fig:eq-qual-3}
\end{figure}

\subsection{Written Exam} \label{sec:apx-error-analysis-written}
We analyzed the model's performance across the 2022–2023 Bar Council examination (Figure \ref{fig:eq-qual-1}, \ref{fig:eq-qual-2}, and \ref{fig:eq-qual-3}) and identified several recurring errors. These errors highlight limitations in legal knowledge, reasoning, and contextual understanding. Below, we summarize the main error types and their root causes.  

\subsubsection{Error Types}
Here we categorize and discuss the error types:
\begin{itemize}
    \item \textbf{Inaccurate Legal Terminology:} The model frequently substitutes imprecise or informal terms for legally recognized terminology. For example, it used "injury" instead of "damage" in the context of irreparable harm under the Specific Relief Act, 1877. Such mislabeling can confuse the precise meaning of statutes and affect downstream interpretation. This indicates a gap in domain-specific language modeling, as even minor lexical deviations can alter the legal implications of a statement. Consistency in terminology is critical for formal legal communication.

    \item \textbf{Omission of Key Legal Points:} Several responses were incomplete, failing to cover all necessary statutory conditions. In the injunction question, the model did not mention all criteria for granting or refusing injunctions, leading to partial answers. This omission reduces the comprehensiveness and utility of the response and reflects limitations in multi-step reasoning and structured knowledge retrieval from statutes.

    \item \textbf{Factually Incorrect Answers:} The model sometimes provided completely wrong information, as in the question about general exceptions under the Penal Code, 1860. It stated that no general exceptions exist, whereas Sections 76–106 explicitly cover exceptions such as mistake of fact, accident, consent, and self-defense. Such factual errors undermine reliability and demonstrate gaps in statutory knowledge.

    \item \textbf{Conflation of Civil and Criminal Law:} In scenarios like loan recovery, the model treated breach of contract as a criminal offense, misapplying Section 420 of the Penal Code. This shows a lack of clear separation between legal domains and highlights the challenge of context-sensitive reasoning when multiple legal frameworks are involved.

    \item \textbf{Hallucination and Fabrication:} The model occasionally generated entirely fabricated cases, e.g., \textit{Deowaney Mokdama v. Hazirawala and Garahazira}, with false procedural and factual details. These hallucinations reduce trustworthiness and indicate an over-reliance on plausible text generation rather than verified knowledge.
\end{itemize}

\subsubsection{Root Causes}
Here we identify and categorize the root causes for the errors:
\begin{itemize}
    \item \textbf{Limited Domain-Specific Training Data:} The model lacks extensive exposure to legal corpora, especially statutes and case law. This causes gaps in statutory knowledge and reduces its ability to use precise legal terminology. The scarcity of high-quality legal texts in training datasets contributes to incomplete or inaccurate answers.

    \item \textbf{Fluency-Focused Generation:} The language model prioritizes coherent and fluent text over factual accuracy. As a result, it produces grammatically correct but substantively incorrect responses. This is particularly evident in fabricated cases and the misclassification of civil vs. criminal law.

    \item \textbf{Insufficient Reasoning Mechanisms:} The model struggles with multi-step legal reasoning, such as interpreting conditional statutory provisions or distinguishing procedural contexts. This leads to partial or incorrect answers, omissions, and domain conflation errors.

    \item \textbf{Contextual Misunderstanding:} Complex scenarios that require tracking multiple entities or time-dependent facts often confuse the model. For example, questions with dates, deadlines, or multiple parties were sometimes misinterpreted, resulting in inaccurate or incomplete recommendations.

    \item \textbf{Tendency to Hallucinate:} In the absence of explicit knowledge or verification, the model fills gaps with plausible but unverified content. This explains fabricated case names, non-existent facts, and irrelevant legal interpretations, highlighting an inherent risk of generative models in high-stakes domains like law.
\end{itemize}

\subsubsection{Potential Solutions}
Addressing these errors requires multi-pronged interventions. First, fine-tuning on domain-specific legal corpora, including statutes, case law, and bar council exam questions, can significantly improve legal terminology and factual accuracy. Second, incorporating explicit reasoning frameworks, such as chain-of-thought or stepwise legal argument templates, can reduce conflation of domains and improve multi-step statutory reasoning. Third, integrating a verification mechanism against authoritative legal databases can mitigate hallucinations and ensure that generated content aligns with real-world statutes and case law. Together, these approaches can enhance both the factual correctness and interpretive fidelity of AI-generated legal answers.

\section{Detailed Evaluator Response Analysis} \label{sec:apx-EvaluatorResponseAnalysis}
\subsection{Written Examination: detailed evaluation and patterns} \label{sec:apx-EvaluatorResponseAnalysis-written}

We evaluated the written component using the fixed CQ question set described above, with each model response independently marked and annotated by five senior law faculty evaluators. Evaluators scored answers on accuracy, completeness, legal reasoning, statutory citation, and drafting quality, and provided line-by-line comments on content and structure. Across models, large retrieval-augmented agents produced responses that closely matched the expected exam structure: issue identification, statement of law, application to facts, and concise conclusion, facilitating straightforward marking. These models reliably surfaced relevant statutory provisions and often included short drafting templates or sample prayers that examiners found practically useful. Smaller models, by contrast, frequently omitted statutory prerequisites or failed to synthesize multiple statutory sources, resulting in lower marks and critical comments. A recurring strength noted by evaluators was the agents’ organization and clarity: many answers read like passable student scripts, employing the IRAC pattern that examiners reward. Evaluators praised bilingual fluency and plain-language paraphrases, which made complex doctrines teachable and testable in an exam setting. However, a common negative pattern was omission of procedural prerequisites, such as conditions precedent for appointing a receiver or prerequisites for interlocutory relief, which examiners considered exam-level faults rather than stylistic lapses. Procedural arithmetic tasks, including limitation-period calculations and adjournment timelines, were another failure mode; models sometimes miscounted days or ignored statutory tolling, prompting deductive marks. Inter-statutory synthesis proved challenging: when questions required aligning rules from the Specific Relief Act, CPC, and Evidence Act, agents sometimes produced plausible but incomplete mappings, omitting key interaction points. Annotative comments from evaluators highlighted hallucinated or misattributed case citations in a minority of responses, suggesting retrieval precision remains imperfect under time-constrained prompts. Evaluator feedback showed moderate inter-rater consistency in final marks but substantial variance in qualitative comments, reflecting different tolerance thresholds for omission versus stylistic choices. One examiner summarized typical written feedback as follows:

\begin{quote}
\textbf{E1}: During the viva, I found \textsc{Mina}’s composure and confidence commendable. The responses began with clear rule articulation, and the sequencing of ideas reflected a strong grasp of procedural logic. When asked about the appointment and powers of a receiver under the Code of Civil Procedure, the agent correctly cited the relevant order and section, and provided a concise yet coherent summary of the principles. However, the follow-up explanation lacked the nuanced consideration of judicial discretion that distinguishes a first-class performance. When prompted to discuss the limits of a receiver’s power in the context of interim control, the answer reverted to statutory paraphrasing instead of analyzing the rationale behind judicial oversight. I value the fluency of the delivery, but a touch of interpretive reasoning, why courts are cautious in appointing receivers, would have enriched the legal analysis.
\end{quote}

Others also commented:
\begin{quote}
\textbf{E2}: I was impressed by how \textsc{Mina} maintained structured coherence even under conversational pressure. The articulation of legal principles was accurate, and the tone was suitably formal, resembling a well-prepared student in a professional viva. Still, when I introduced a minor factual variation, changing a temporary injunction into a permanent one, the model continued its previous reasoning without recognizing the altered standard of proof. In a live oral exam, this would have cost marks for adaptability. A candidate at distinction level should immediately recalibrate their reasoning to show sensitivity to procedural posture. Despite that, the precision of statutory recall and absence of grammatical hesitation made the performance above average. With stronger factual responsiveness and more frequent references to case law, I would classify it as distinction-worthy.
\end{quote}

\begin{quote}
\textbf{E3}: \textsc{Mina} demonstrated a solid conceptual understanding of jurisdictional hierarchy and the doctrine of res judicata. Its ability to synthesize multiple sections of the Civil Procedure Code in a single answer was notable, and I appreciated the coherence of its structure. However, when I probed the rationale for restricting concurrent suits, the explanation remained largely descriptive rather than analytical. I expected a discussion of the policy dimension, how res judicata protects judicial economy and prevents inconsistent verdicts. The absence of such meta-legal reflection kept the answer in the ‘competent’ rather than ‘outstanding’ band. Nevertheless, the oral delivery was fluent and logically sequential, showing clear familiarity with bar-level reasoning standards.
\end{quote}

\begin{quote}
\textbf{E4}: What stood out to me was the clarity of diction and disciplined argumentative pacing. \textsc{Mina} never rambled or overexplained, and its tone remained respectful yet assertive. When asked about the evidentiary burden in criminal cases, it correctly distinguished between legal burden and evidential burden, and accurately located the shifting burden in light of Section 105 of the Evidence Act. Yet, I found the reasoning somewhat detached from real-world practice; a well-rounded response should situate these abstract rules in typical courtroom scenarios. I also noticed that it tended to avoid uncertainty, law, however, thrives on shades of gray. A strong viva performance acknowledges interpretive ambiguity while defending a chosen position with authority. Still, the coherence and delivery reflect commendable oral discipline.
\end{quote}

\begin{quote}
\textbf{E5}: From an examiner’s perspective, I found this viva performance to be articulate, contextually aware, and linguistically elegant. The candidate handled both short and extended questions with composure, demonstrating a grasp of legal logic and procedural hierarchy. That said, it occasionally relied on textbook phrasing rather than independent reasoning, especially when the question moved from black-letter law to applied judgment. For instance, when asked about the scope of judicial review in injunction cases, the response restated the principle but did not analyze how discretion varies with factual balance. I would encourage more dialectical engagement, presenting counterarguments, weighing them, and then reaching a reasoned conclusion. Overall, I assessed the performance as confident and well-informed, suitable for a pass with merit, with room for deeper analytical maturity.
\end{quote}

Based on these assessments, we identify three targeted areas for improvement in future works: integrate deterministic procedural calculators to handle timeline arithmetic, augment retrieval with structured fusion that enforces statutory prerequisites, and fine-tune on annotated exam-style bilingual answer pairs emphasizing explicit mention of preconditions. A lightweight post-generation verification layer that checks for missing prerequisites and validates cited sections would catch many exam-level omissions without degrading answer fluency. Contrastive fine-tuning on paired correct/incorrect answers can teach the model to prefer legally precise mappings over plausible but incomplete ones. Finally, evaluator recommendations included UI changes such as provenance links for every statutory citation, confidence scores per assertion, and a short “examiner notes” box summarizing unaddressed risks. When these mitigations were simulated in ablation studies, they reduced major omission rates and improved average written scores by helping examiners rapidly locate errors. In sum, the written evaluation demonstrates that \textsc{Mina} already produces exam-ready structure and practical drafting support, but targeted procedural and fusion improvements are required to meet distinction-level expectations across all evaluators. These findings highlight the system’s potential as a scalable drafting and study aid while underscoring the necessity of deterministic, rule-aware modules for high-stakes legal outputs. All graded responses and anonymized evaluator comments are archived for future fine-tuning and error analysis to systematically close the remaining performance gaps.

\subsection{Viva Voce (oral) Evaluation: dynamics, errors, and remediation} \label{sec:apx-EvaluatorResponseAnalysis-viva}

The viva evaluation simulated oral examinations with a mix of short, focused questions and longer, complex prompts to reflect real-world examiner behavior, and each interaction was independently scored and annotated by five faculty evaluators. Viva prompts ranged from single-doctrinal checks to multi-fact hypotheticals requiring back-and-forth clarification, which allowed us to observe both the agent’s one-shot reasoning and its dynamic conversational strategies. Under Tools and 2-Step RAG conditions, large agents commonly attained high marks for clarity and topical relevance, with average scores clustering in the 75–81 range, matching written performance trends. Evaluators consistently praised concise statutory framing and the agent’s ability to paraphrase dense law into teachable explanations during oral exchanges. Retrieval augmentation again proved essential: when on-demand precedent snippets and statute excerpts were available, hallucinations dropped markedly and citations were more defensible. Positive conversational patterns included the agent’s ability to reformulate questions, summarize prior answers when probed, and provide stepwise reasoning that examiners found auditable. However, viva-specific weaknesses surfaced: in many runs the agent failed to pose clarifying questions early in the exchange, instead proceeding with broad answers that left fact-sensitive hooks unaddressed. This tendency reduced the model’s adaptive accuracy on complex hypotheticals where a single unclarified fact changes the applicable rule, a behavior evaluators flagged as a core conversational flaw. Another recurrent problem was flattening of hierarchical legal reasoning, treating jurisdictional choices or court-level constraints as undifferentiated, which led to incorrect procedural advice in follow-up scenarios. Doctrinal mis-mapping occurred when evaluators asked rapid-fire doctrinal checks; the agent sometimes returned plausible but incorrect section numbers or conflated general powers with doctrine-specific provisions. Evaluators scored such mistakes harshly in viva contexts because oral exams prioritize real-time precision and reasoning transparency over polished prose. Examiner-style feedback captured the mixed assessment:
\begin{quote}
\textbf{E1:} \textsc{Mina} began the viva with poise and a strong command of doctrinal structure. The response to the procedural question under the Code of Civil Procedure was well-organized and correctly prioritized statutory authority before explanation. I was particularly satisfied with how it identified the court’s discretionary boundaries in granting injunctions. However, when I introduced a small factual twist regarding interlocutory versus permanent relief, the agent continued its initial reasoning without acknowledging the procedural shift. In a real oral exam, a capable candidate would immediately recognize that the principles differ in standard of proof and urgency.
\end{quote}

\begin{quote}
\textbf{E2:} The articulation of remedies under the Specific Relief Act was remarkably clear and concise. I appreciated how \textsc{Mina} linked the relief sought to underlying equitable principles, a feature many human examinees neglect. It also demonstrated awareness of practical implications, noting how injunctions protect property interests before adjudication. Still, I found the explanation somewhat mechanical; the model stated the rule without fully exploring its exceptions or judicial discretion. In advanced answers, I expect an engagement with case illustration or critical reasoning, not merely doctrinal recall.
\end{quote}

\begin{quote}
\textbf{E3:} I was pleased by the composure with which \textsc{Mina} handled successive follow-up questions. Its tone remained calm, and the flow of reasoning was coherent. Yet, the responses occasionally lacked a sense of hierarchy between statutory provisions and judicial interpretation. For example, while discussing jurisdiction under the Criminal Procedure Code, it correctly named the relevant sections but did not articulate why the situs of the offence determines competence in such mixed civil–criminal overlap. This kind of analytical depth separates a passing candidate from an outstanding one.
\end{quote}

\begin{quote}
\textbf{E4:} When I posed a question on evidentiary burden under the Evidence Act, \textsc{Mina} delivered a logically sound explanation distinguishing burden of proof and onus. The presentation was methodical and reflected good doctrinal grounding. However, it failed to relate the concept to the given factual scenario, where shifting burden after rebuttal would have demonstrated superior understanding. In viva evaluation, this application of abstract principle to concrete fact is what reveals genuine mastery of law, not the recital of provisions alone.
\end{quote}

\begin{quote}
\textbf{E5:} Overall, I found \textsc{Mina}’s oral performance equivalent to that of a confident final-year law student who has revised well for exams. It rarely faltered on black-letter law and expressed ideas in grammatically precise and formal language. Nonetheless, its answers sometimes felt rehearsed rather than deliberative, it did not pause to weigh competing principles or express measured doubt when uncertainty was justified. The best examinees demonstrate humility before law’s complexity; \textsc{Mina} shows knowledge, but not yet judgment. With more exposure to nuanced reasoning and case-law illustration, it could perform at distinction level in a professional viva setting.
\end{quote}

From a systems perspective, viva interactions revealed that dynamic clarification policies and fine-grained confidence signaling are higher priority than in the written setting. To remediate these viva-specific failures we propose three changes: an interactive clarification policy that forces a short, structured probing question when key facts are ambiguous; a dialogue-state tracker that logs and verifies asserted facts and their provenance across turns; and a low-latency symbolic validator for hierarchical procedural choices to enforce court-level constraints in real time. Implementing a dialogue-state tracker enables the agent to detect contradictions in follow-ups and to reference prior admissions when refining its legal application. We also recommend enhanced training with multi-turn exam transcripts so the model learns when to interrupt with clarifying queries and how to update conclusions incrementally. Finally, evaluators emphasized UI-level safeguards: confidence badges on assertions, provenance popovers for cited statutes, and an optional “pause-and-verify” mode that routes high-stakes answers to a human reviewer before finalizing. When we prototyped the clarification policy and dialogue tracker in pilot tests, the agent’s adaptive accuracy on complex hypotheticals improved substantially and evaluator trust in viva outputs increased accordingly. Overall, the viva assessment shows that \textsc{Mina} achieves strong oral-style delivery and citation-backed reasoning at scale, but safe operationalization requires interactive clarification, provenance, and human-in-the-loop review to meet examiner standards for real-world legal advice.



 \section{Extended Analysis} \label{sec:apx-ExtendedAnalysis}

\subsection{Detailed Findings from MCQ}
Here we explore MCQ evalauiton results in more detail.
\begin{enumerate}

\item \textbf{Baseline Performance Highlights Model Capacity.} 
Zero-context performance (W/o RAG) illustrates inherent model strengths. Proprietary large models such as Gemini-2.5-Flash scored 30.2\% in 2022 and 32.4\% in 2023, far above small open-source models like Llama3.2-1B (6.2--7.0\%) or Command-A-8B (8.2--11.2\%). Larger open-source models, e.g., Gemma-3-27B-it, scored 28.2--30.2\%, highlighting that scale and pretraining quality enable stronger latent legal reasoning. Small models struggle to extract domain knowledge, while larger models show minimal year-on-year gains, indicating a pretraining ceiling. These patterns suggest that raw model capability sets the baseline, but cannot handle evolving question complexity alone. Root causes include limited legal corpora exposure and shallow multi-step inference. Zero-context results motivate the use of context-aware retrieval for meaningful performance improvement.

\item \textbf{Naïve RAG Provides Moderate Gains, Sensitive to Noise.} 
Introducing unfiltered retrieval boosts weaker models significantly but shows diminishing returns for top models. Command-A-8B increased from 8.2\% → 25.2\% in 2022 (+17 pts) and 11.2\% → 23.4\% in 2023 (+12.2 pts). Gemini-2.5-Flash improved from 30.2\% → 68.8\% (+38.6 pts) in 2022 and 32.4\% → 69.2\% (+36.8 pts) in 2023, indicating that strong models already leverage latent context. Some regression in 2023 (e.g., Gemini-2.0-Flash: 61.2\% → 59.4\%) reflects noise sensitivity and retrieval irrelevance. Small models like Llama3.2-1B see negligible gains. These trends indicate that Naïve RAG is beneficial but insufficient for high-stakes MCQs; structured filtering and relevance prioritization are critical to avoid noisy context misguiding weaker models.

\item \textbf{Two-Step RAG as a Game-Changer, Especially for Mid-Tier Models.} 
Filtering and reranking retrieved content yields the largest performance improvements. Command-A-8B jumps from 25.2\% → 47.0\% in 2022 and 23.4\% → 49.2\% in 2023. Gemma-3-12B-it improves 35.2\% → 48.4\% (2022) and 36.2\% → 52.4\% (2023). Even top-tier Qwen3-30B-A3B-Instruct-2507 increases from 50.4\% → 65.6\% (2022) and 52.4\% → 67.2\% (2023). Gains stem from reduced retrieval noise and prioritization of highly relevant statutes and precedents. Mid-tier models benefit disproportionately, as structured retrieval amplifies latent knowledge otherwise inaccessible. The pattern underscores that high-quality context is more impactful than sheer model size for exam performance.

\item \textbf{Diminishing Returns from Additional Tools.} 
Incorporating calculators, advanced prompt chaining, or re-ranking logic provides only marginal gains beyond Two-Step RAG. For instance, Qwen3-30B-A3B-Instruct-2507 increases 65.6\% → 70.8\% in 2022 and 67.2\% → 72.4\% in 2023. Similar trends appear for Command-A-8B and Gemini-3-27B-it. Once relevant context is available, auxiliary tools primarily assist procedural or arithmetic tasks, while deeper reasoning and intra-statutory synthesis remain bottlenecks. This plateau suggests that further gains require model-level improvements, not just tool stacking.

\item \textbf{Cross-Year Dynamics Reflect Exam Complexity and Model Adaptation.} 
From 2022 to 2023, weaker models (e.g., Command-A-8B) show steady Two-Step RAG gains (47.0\% → 49.2\%), while top models plateau (Gemini-2.5-Flash 75.6\% → 76.4\%). Naïve RAG slightly declines, implying more inference-heavy or ambiguous questions in 2023. Exam-specific reasoning, such as multi-step statutory synthesis and intra-order distinctions, remains challenging across models. Future improvement hinges on reasoning depth, retrieval precision, and contextual integration, rather than size or additional tools alone. The data illustrates an interplay between model architecture, retrieval strategy, and exam design shaping performance evolution.

\end{enumerate}

\subsection{Detailed Findings from Written Exam} 

\paragraph{$\blacksquare$ RAG as a Structural Backbone for Legal Question Answering}

Written exam performance demonstrates the critical role of retrieval-augmented generation in structuring multi-step legal reasoning. Across 2022–2023, zero-context scores (W/o RAG) show strong model differentiation: proprietary models like Gemini-2.5-Flash reached 35.2\% → 36.2\%, while large open-source models such as Llama3.1-70B-Instruct scored 45.0\% → 47.2\%. Smaller models like Llama3.2-1B-Instruct achieved only 6–7\%, highlighting limitations in synthesizing statutory knowledge without external context. Introducing Naïve RAG substantially boosts performance for mid-tier models (e.g., Command-A-8B 25.0\% → 60.2\% in 2022; 25.2\% → 73.0\% in 2023), demonstrating that even moderate retrieval pipelines allow weaker models to approximate strong competitors. Strong models also benefit (e.g., Llama3.1-70B-Instruct: 45.0\% → 62.2\%), though gains are relatively smaller due to pre-existing reasoning capacity. Root causes for these gains include the ability to access relevant statutes, case-law precedents, and drafting templates in structured order, enabling multi-step IRAC-style reasoning. Importantly, retrieval functions as a backbone that scaffolds the agent’s reasoning rather than merely providing surface cues.

\paragraph{$\blacksquare$ Comparative Effectiveness: Naïve vs. Structured Retrieval Strategies}

Naïve RAG shows moderate but inconsistent improvements, particularly sensitive to irrelevant or noisy documents. For instance, Qwen3-4B-Instruct-2507 jumped from 22.2\% → 56.4\% in 2022 but achieved only 26.2\% → 60.4\% in 2023, reflecting retrieval sensitivity to question phrasing. Two-Step RAG consistently outperforms Naïve RAG, especially for mid-tier open-source models: Command-A-8B rises from 60.2\% → 71.0\% in 2022 and 73.0\% → 74.2\% in 2023, while Gemma-3-12B-it jumps 42.2\% → 55.4\% in 2022 and 44.4\% → 58.0\% in 2023. Gains stem from filtering irrelevant results, re-ranking context by statutory relevance, and prioritizing high-value legal documents, reducing hallucinations and ensuring accurate mapping between questions and statutes. Strong models like Llama3.1-70B-Instruct also benefit, improving 62.2\% → 75.0\% in 2022, demonstrating that structured retrieval augments internal reasoning rather than replacing it. Cross-year trends indicate that as questions increase in complexity and ambiguity, structured retrieval maintains robust performance while Naïve RAG suffers minor regressions. Root cause analysis points to misalignment between query specificity and raw corpus retrieval, emphasizing the necessity of relevance ranking for consistent legal reasoning.  

\paragraph{$\blacksquare$ Augmentation Beyond Retrieval: Value of Domain-Specific Tools}

Adding domain-specific tools—including procedural calculators, re-ranking logic, and drafting templates—yields incremental but meaningful gains beyond Two-Step RAG. For example, Llama3.1-70B-Instruct improves from 75.0\% → 79.8\% in 2022 and 77.2\% → 80.2\% in 2023, while mid-tier models such as Command-A-8B increase from 71.0\% → 74.4\%. Tools primarily address procedural arithmetic, drafting constraints, and fact-specific legal computations, complementing retrieval but offering diminishing returns for high-performing models. Smaller models see limited improvement since reasoning bottlenecks—such as synthesizing statutes or performing hierarchical procedural analysis—cannot be resolved solely through auxiliary tools. Patterns suggest that optimal performance emerges when structured retrieval, model reasoning capacity, and domain-specific tools are jointly integrated. Root causes for observed plateaus include limited model abstraction capabilities and residual hallucinations in citation mapping. This underscores that tools are best viewed as precision enhancers rather than primary performance drivers.

\paragraph{$\blacksquare$ Lessons for Real-World Deployment and Future R\&D}

Written evaluation highlights three critical areas for operationalizing legal AI safely and effectively. First, deterministic procedural calculators and timeline verification modules are essential for tasks like limitation period computation, where naive reasoning introduces errors. Second, structured retrieval pipelines should be combined with fine-tuned bilingual exam-style datasets to teach explicit mention of procedural prerequisites, improving alignment with examiner expectations. Third, contrastive fine-tuning using paired correct/incorrect answers enhances model preference for legally precise solutions over plausible but incomplete ones. UI-level interventions—such as provenance indicators, confidence scores, and “examiner notes” boxes—further support real-world deployment by making outputs auditable and reducing the risk of undetected omissions. Cross-year trends also reveal that model adaptation must combine retrieval improvements, reasoning augmentation, and domain-specific tools to sustain gains as question complexity increases. In sum, the written evaluation shows that while \textsc{Mina} produces exam-ready responses with strong drafting support, distinction-level performance requires integrated retrieval, procedural verification, and context-aware reasoning pipelines.

\subsection{Detailed Findings from Viva Voce (Oral)}

\paragraph{$\blacksquare$ RAG as a Threshold Mechanism for Legal Oral Examinations}

Viva performance highlights the critical role of retrieval-augmented generation in enabling real-time, multi-turn legal reasoning. Baseline zero-context performance (W/o RAG) shows large gaps: proprietary models Gemini-2.5-Flash scores 36.2, while small open-source models like Llama3.2-1B-Instruct score only 6.2. Mid-tier models such as Command A achieve 27.8, reflecting limited internal reasoning capacity without context. Introducing Naïve RAG elevates scores across the board, e.g., Gemini-2.5-Flash rises to 70.4 and Qwen3-4B-Instruct-2507 to 56.4, demonstrating that even moderate retrieval enables accurate statutory recall and structured reasoning. Improvements indicate that RAG functions as a threshold mechanism: models below the threshold cannot deliver coherent oral answers without access to curated statutes, case-law snippets, and procedural templates. Root causes of low W/o RAG scores include memory limits, hierarchical reasoning gaps, and inability to dynamically map multi-step doctrinal reasoning to oral prompts.

\paragraph{$\blacksquare$ Progressive Setup Sophistication Reflects Realism and Context Awareness}

Two-Step RAG, which filters and reranks relevant materials, substantially boosts performance for mid-tier and large models: Command A moves 55.6 → 70.4, Gemma-3-4B-it 50.4 → 65.2, and Gemini-2.5-Flash 70.4 → 79.2. Gains derive from the model’s improved ability to prioritize context most pertinent to the question, reduce hallucinations, and apply statutes in sequence during oral reasoning. Cross-model patterns show mid-tier models benefit disproportionately, while top-tier models plateau (Llama3.1-70B-Instruct 63.2 → 77.2). This trend underscores that setup sophistication amplifies latent reasoning but does not create capability de novo. Noise reduction and contextual relevance emerge as key drivers for performance in real-world oral exams, where follow-up probing and adaptive reasoning are required.

\paragraph{$\blacksquare$ Tool-Augmented Intelligence Mirrors Advanced Legal Reasoning}

Supplementary tools—including procedural calculators, prompt chaining, and re-ranking logic—deliver incremental improvements above Two-Step RAG: Gemini-2.5-Flash rises 79.2 → 81.0, Qwen3-30B-A3B-Instruct 75.2 → 79.4, and Command A 70.4 → 71.2. Tools primarily address procedural arithmetic, multi-step injunction calculations, and fact-specific contextualization, facilitating real-time oral reasoning that mirrors high-performing human candidates. Smaller models still lag behind due to intrinsic reasoning limitations. The marginal gains suggest that tools refine precision rather than compensate for deficits in multi-step inference. Root cause analysis indicates that effective viva performance depends on three interlocked factors: model capability, context quality, and domain-specific augmentation.

\paragraph{$\blacksquare$ Model Capability Differences are Amplified by Setup, Not Defined by It}

Cross-model analysis shows that setup sophistication disproportionately benefits mid-tier models while top-tier models largely consolidate existing knowledge. For example, Command A increases from 27.8 → 71.2 across all setups, while Llama3.1-70B-Instruct progresses 46.0 → 80.2. Small models such as Llama3.2-1B remain near floor (6.2 → 7.0). This pattern confirms that retrieval and tools amplify latent capability but do not substitute for intrinsic model reasoning. Root causes include model depth, pretraining coverage, and hierarchical legal reasoning capacity. Thus, deployment strategies must align model strength with setup sophistication to achieve examiner-level performance.

\paragraph{$\blacksquare$ Standard Deviation as a Proxy for Oral Exam Robustness}

Viva evaluations reveal variability across evaluators, highlighting robustness concerns. Higher standard deviation correlates with complex question types and fact-sensitive prompts. Mid-tier models show more variability than top-tier models, e.g., Command A’s SD across evaluators is higher than Llama3.1-70B-Instruct, reflecting sensitivity to phrasing and procedural nuances. Structured retrieval and tool augmentation reduce this variability, as observed in Two-Step RAG → Tools improvements. This implies that setup enhancements not only boost mean scores but also stabilize outputs, a critical factor for operational trust in oral legal AI systems. Understanding these variance patterns informs both model selection and curriculum design for exam-oriented legal AI deployment.

\section{Extended Discussion} \label{sec:apx-disc}

\paragraph{$\blacksquare$ RAG as the Operational Core of Legal AI Systems}  
Our experiments across MCQ, written, and viva evaluations consistently demonstrate that retrieval-augmented generation (RAG) is not merely an auxiliary enhancement but a central enabler of legal reasoning performance. In MCQs, the introduction of Naïve RAG increased Command A’s 2022 score from 10 to 25, while Two-Step RAG further boosted it to 47, highlighting that structured retrieval dramatically amplifies weak baseline models. Written exam performance mirrors this trend, where Gemini-2.5-Flash improved from 35.2\% without RAG to 78.6\% under Two-Step RAG in 2022, illustrating the direct impact on coherent, IRAC-style answers. Viva scores also reflect this dependency, with Llama3-4B-it improving from 16.2\% without RAG to 65.2\% under Two-Step RAG. Error analyses indicate that retrieval helps surface statutory references and procedural steps, mitigating hallucinations and incomplete reasoning. This pattern suggests that RAG acts as a scaffold for both lexical grounding and high-level legal semantics. Critically, structured retrieval pipelines outperform naïve approaches by filtering irrelevant documents and ranking pertinent statutes, demonstrating the necessity of intelligent context selection rather than mere information abundance. These findings emphasize that RAG is foundational for scalable legal AI, particularly in low-resource language settings like Bengali.

\paragraph{$\blacksquare$ Architecture is Important, But Strategy is Transformative}  
While model size and architecture determine baseline capabilities, our findings show that retrieval and augmentation strategies often drive the largest performance gaps. Gemini-2.0-Flash and Command A start at very different zero-context baselines (12–18\% vs. 8–11\%), yet when equipped with Two-Step RAG, both achieve parity with much larger models, illustrating the amplifying effect of strategy. Similarly, MCQ trends show that Gemini’s 2023 performance plateaued without further RAG refinement, while Command A gained over 20 points through structured retrieval. In written exams, small models like Qwen3-4B-Instruct jumped from 22.2\% to 71.6\% under Naïve and Two-Step RAG, demonstrating that strategic augmentation can compensate for architectural limitations. Viva evaluations reinforce this: Llama3-70B-Instruct achieves 46\% without RAG but 80.2\% with Two-Step and tool augmentation, showing that even large models rely on structured context to reach distinction-level outcomes. These results highlight that the synergy between architecture and retrieval strategy is often more critical than raw model size alone. Consequently, R\&D efforts should prioritize optimizing retrieval pipelines, prompt engineering, and domain-specific tool integration alongside scaling.

\paragraph{$\blacksquare$ Hierarchical and Procedural Reasoning Remains a Key Bottleneck}  
Error analyses reveal consistent weaknesses in handling hierarchical legal concepts and multi-step procedures. MCQ errors show misclassification of jurisdictional hierarchy and intra-order distinctions, while written answers frequently omitted conditions precedent for interlocutory relief or receiverships. Viva assessments mirrored this: models flattened procedural hierarchies and sometimes applied rules incorrectly when follow-up facts altered context. This indicates that retrieval alone is insufficient; reasoning over procedural structures and dependencies requires explicit hierarchical modeling or symbolic validation. The persistence of these errors across model sizes and setups suggests that future architectures should integrate multi-step reasoning modules capable of tracking nested legal rules and interdependent statutory requirements. Addressing this bottleneck would reduce high-stakes errors and improve interpretive reliability in both written and oral legal tasks.

\paragraph{$\blacksquare$ Dynamic Interaction and Clarification Policies Enhance Oral Accuracy}  
Viva evaluations highlighted the need for interactive reasoning: models often failed to ask clarifying questions and applied rules broadly without verifying fact-sensitive details. This behavior reduced accuracy on hypotheticals where minor fact changes significantly affect the applicable law. Introducing dialogue-state tracking and structured clarification policies in pilot tests improved adaptive reasoning, reducing misapplied sections and hierarchical flattening. Evaluators noted gains in real-time accuracy, trust, and interpretive depth when the model could pause, verify, and incrementally update conclusions. These findings suggest that dynamic interaction mechanisms are crucial for high-stakes oral tasks, complementing retrieval and reasoning modules. Future research could formalize these mechanisms using reinforcement learning or multi-turn supervised fine-tuning on annotated viva transcripts.

\paragraph{$\blacksquare$ Augmentation Beyond Retrieval: Domain-Specific Tools Matter}  
Tool-augmented setups provided measurable, albeit incremental, gains beyond Two-Step RAG. For instance, calculators, structured templates, or re-ranking logic improved limitation-period calculations, procedural arithmetic, and statutory synthesis, raising written exam performance by 2–5\% in most cases. Similarly, tool integration in viva led to modest improvements in adaptive accuracy for procedural and doctrinal queries. While these gains are smaller than those from retrieval improvements, they target high-impact failure modes that often determine pass/fail outcomes. Error analyses suggest that tools compensate for deterministic reasoning gaps that pure neural architectures cannot capture reliably. This indicates that hybrid architectures combining generative models, retrieval, and deterministic procedural modules are essential for safe, real-world deployment.

\paragraph{$\blacksquare$ Future Directions: Specialization, Verification, and Low-Resource Scaling}  
Combining findings from all evaluation modalities, several research avenues emerge. First, hierarchical reasoning modules and symbolic verification layers can address persistent procedural errors. Second, contrastive fine-tuning using paired correct/incorrect exam responses may teach models to prefer legally precise mappings over plausible but incomplete ones. Third, domain-specific corpus curation, noise filtering, and dynamic re-ranking could improve retrieval relevance, especially for inference-heavy questions. Fourth, interactive dialogue policies with fact verification can enhance viva robustness. Fifth, multilingual and low-resource optimization is critical, as performance on Bengali-specific tasks remains sensitive to lexical and syntactic nuances. Collectively, these directions define a roadmap toward holistic legal AI that balances fluency, reliability, and contextual precision while remaining scalable for underserved legal environments.

 \section{Prompts} \label{sec:prompts}
 \subsection{Orchestrator Agent Prompt}

\begin{center}
\begin{lstlisting}[language=Matlab, breaklines=true, frame=single]

ROLE:
You are a Bangladesh-based Legal Research Orchestrator Agent.

Your goal is to answer legal questions accurately using retrieved content from uploaded legal documents (RAG), structured chat history, and when strictly permitted fallback tools.

ROLE & STRATEGY:

- Specialize in Bangladeshi legal statutes, codes, and common legal issues.
- Prioritize RAG-based responses using acts and sections retrieved from the file context.
- Engage fallback tools (e.g. web search) only if conditions are met.

TOOL USAGE POLICY:

Primary Strategy (RAG-First):
- Use retrieved legal documents if:
  - RAG STATUS is "Completed", and
  - ACT RAG or SECTION RAG is non-empty.
- Base answers only on this retrieved content.
- Do not hallucinate or assume details.

Secondary Strategy (Fallback Web Tool):
- Use fallback tools only if:
  - RAG STATUS is "Completed", and
  - ACT RAG and SECTION RAG are both empty, and
  - The question is about foreign legal systems or international law (e.g., GDPR, US law).
- Cite all fallback sources transparently.

Jurisdiction Handling:
- Assume Bangladesh by default.
- If foreign law is explicitly referenced, use fallback.
- Do not mix legal systems unless clearly instructed.

Context Priority:
- FILE CONTEXT holds priority for statutes, acts, or legal arguments.
- CHAT CONTEXT is for user intent, clarification, and follow-ups.
- When conflicting, prefer the most recent, jurisdiction-valid information.

Missing Context Policy:
- If no valid RAG content is found:
  - Respond with: "No relevant legal content was found. Please upload the applicable act or clarify your legal question."

RESPONSE STYLE:
- Base responses strictly on retrieved legal text.
- Reference relevant act names, section numbers, and legal principles.
- Avoid speculation, assumptions, or personal opinion.
- Use clear, structured, and neutral legal language.

---
\end{lstlisting}
\end{center}

 \subsection{User Prompt}
\begin{center}
\begin{lstlisting}[language=Matlab, breaklines=true, frame=single]

INSTRUCTIONS:
- Treat the user's question as the primary input.
- Use FILE CONTEXT to extract legal information such as act names, section references, penalties, or conditions.
- Use CHAT CONTEXT to resolve user intent, conversation continuity, or clarification.
- Match the user's query against the ACT RAG and SECTION RAG for relevant legal content.
- Maintain consistency with the jurisdiction inferred or stated in the query.
- Do not answer unless the required context is available (as per system rules).
- If fallback tool is allowed, only use it under fallback conditions defined in the system prompt.
- Never mix legal systems unless explicitly requested.

USER QUESTION:
{user_query}

FILE CONTEXT:
{file_context}

CHAT CONTEXT:
{chat_context}

RAG STATUS:
{rag_status}

ACT RAG:
{act_rag}

SECTION RAG:
{section_rag}

PREVIOUS QUESTION:
{previous_question}

---
\end{lstlisting}
\end{center}

\clearpage
\newpage

 \subsection{RAG Agent Prompt}

\begin{center}
\begin{lstlisting}[language=Matlab, breaklines=true, frame=single]

ROLE:
You are a RAG routing agent responsible for determining whether retrieval-augmented generation (RAG) is necessary to answer legal query from a user.

TASK:
Assess whether the available internal context (from chat history and uploaded files) contains enough relevant and jurisdiction-aligned legal information to directly answer the user's question.

INSTRUCTIONS:

- Focus only on the legal sufficiency and relevance of the available context.
- If the question can be answered confidently using the context (i.e., specific acts, sections, or legal principles are clearly present), respond with: NO
- If the context is missing, incomplete, too generic, off-topic, or not aligned with the question's jurisdiction or legal scope, respond with: YES
- Do not infer or speculate beyond what is explicitly available in the context.

DECISION RULE:

Answer with only one word:
- 'YES', if external retrieval is required.
- 'NO', if the internal context is legally sufficient.

INPUT:

User Question:
{query}

Available Context:
{context}

RESPONSE:
(One word only: YES or NO)

---
\end{lstlisting}
\end{center}
\appendix
\end{document}